\documentclass[journal,onecolumn]{IEEEtran}

\ifCLASSINFOpdf
\else
   \usepackage[dvips]{graphicx}
\fi
\usepackage{url}

\usepackage{graphicx}
\usepackage{amsmath}
\usepackage{amssymb}
\usepackage{xcolor}
\usepackage{cite}
\usepackage{xcolor}
\usepackage{subcaption}
\usepackage{float}

\usepackage[linesnumbered,ruled,vlined]{algorithm2e}

\SetCommentSty{mycommfont}
\SetKwInput{KwInput}{Input}                % Set the Input
\SetKwInput{KwOutput}{Output}              % set the Output

\begin{document}
	
	\begin{center}
		{\LARGE \textbf{NARS vs. Reinforcement learning}}\\
             \vspace{1em}
		{\Large ONA vs. $Q$-Learning}\\
		\vspace{1em}
		{\large Project of AGI course}\\
		\vspace{1em}
		{\large Ali Beikmohammadi}\\
		\vspace{1em}
		\textit{Department of Computer and Systems Sciences\\
  Stockholm University\\
  SE-164 07 Kista, Sweden \\
  \texttt{beikmohammadi@dsv.su.se} \\}
	\end{center}

	\begin{center}
		\rule{\textwidth}{0.2mm}
		\vspace{1mm}
	\end{center}		

	\begin{abstract}
One of the realistic scenarios is taking a sequence of optimal actions to do a task. Reinforcement learning is the most well-known approach to deal with this kind of task in the machine learning community. Finding a suitable alternative could always be an interesting and out-of-the-box matter. Therefore, in this project, we are looking to investigate the capability of NARS and answer the question of whether NARS has the potential to be a substitute for RL or not. Particularly, we are making a comparison between $Q$-Learning and ONA on some environments developed by an Open AI gym. The source code for the experiments is publicly available in the following link: \url{https://github.com/AliBeikmohammadi/OpenNARS-for-Applications/tree/master/misc/Python}.

	%\textbf{Collaborators}: list the names and affiliations of expected collaborators on the project here
	\end{abstract}

	\begin{center}
		\rule{\textwidth}{0.2mm}
	\end{center}		

	\vspace{5mm}
	
%\begin{multicols*}{2}

\section{Introduction and Background}
Model-free Reinforcement Learning (RL) algorithms by combining RL and high-capacity function approximations gives hope of automation of a broad range of decision making and control tasks \cite{wang2022sample,deisenroth2011pilco, sutton2018}.
Researchers have solved challenging problems, e.g., in game playing \cite{mnih2015,silver2016, silver2017}, financial markets \cite{meng2019, fischer2018, 10.1145/3383455.3422540}, robotic control \cite{haarnoja2018soft, kober2013, lillicrap2015, schulman2015trust}, optimal control \cite{126844, PERRUSQUIA2021145}, healthcare \cite{CORONATO2020101964, 10.1145/3477600}, autonomous driving \cite{shalev2016, ref4}, and recommendation systems~\cite{rec1,rec2, ref3}. While RL works well for environments where
infinite data can be generated, it does not feature compositional representations that would allow for data-efficient learning \cite{hammer2021autonomy}.

Recently, Non-Axiomatic Reasoning System (NARS) is proposed as a general-purpose reasoner that adapts under the Assumption of Insufficient Knowledge and Resources (AIKR) \cite{hammer2020reasoning, wang2009insufficient, wang2013non}. There have been implementations based on this non-axiomatic logic, including OpenNARS \cite{hammer2016opennars} and ONA (OpenNARS for Applications) \cite{hammer2020opennars}. ONA is more capable than OpenNARS in terms of reasoning performance, and recently has also been experimentally compared with RL \cite{eberding2020sage, hammer2021autonomy}. Specifically, in \cite{hammer2021autonomy}, a comparison has been made between ONA and $Q$-Learning \cite{watkins1992q}. Three simple environments, Space invaders, Pong, and grid robot, have been used for comparison. The results have shown that ONA has provided more stable results while maintaining almost the same performance in terms of success ratio.

In this software project, the aim is to compare $Q$-Learning as a basic algorithm in RL with ONA in several more challenging tasks. There are several challenges when comparing $Q$-Learning with ONA, which can be summarized as:
\begin{itemize}
    \item Statements instead of states,
    \item Unobservable information,
    \item One action in each step,
    \item Multiple objectives,
    \item Hierarchical abstraction,
    \item Changing objectives,
    \item Goal achievement as reward.
\end{itemize}
Their details can be found in \cite{hammer2021autonomy}. Since ONA does not assume that in every step an action has to be chosen, in \cite{hammer2021autonomy}, to make the techniques comparable, they have added an additional \textit{nothing} action for the $Q$-Learner in each example. However, in this work, we will choose a random action when ONA does not recommend any action to be taken. This has been decided to keep the originality of the tasks/environments as much as possible. Also, taking random actions could be beneficial for the agent in terms of exploration of the environment. In this work, we will basically use the assumptions mentioned in \cite{hammer2021autonomy} when comparing the two algorithms, otherwise we will mention it. 
The rest of this work is organized as follows. In Section \ref{section2}, we describe the environments as well as the hyperparameters that have been used for both algorithms. Experimental results and analyses are reported in Sections \ref{section3}. Also, additional figures are included in Appendix \ref{section4}.

\section{Hyperparameters and Environments} \label{section2}
We compare ONA with a standard table-based Q-Learner implementation \cite{watkins1992q} with exponentially decaying $\epsilon$ value. Specifically, we have used a series of fixed hyperparameters in all tasks, which are as follows:
\begin{itemize}
    \item $\alpha = 0.7$,
    \item $\gamma = 0.618$,
    \item $\epsilon_{max} = 1$,
    \item $\epsilon_{min} = 0.01$,
    \item decay $= 0.01$,
\end{itemize}
where $\alpha$ is the learning rate, $\gamma$, known as discount factor, controls how much to favour future rewards over short-term reward. Besides $\epsilon$ is the exploration rate. It encodes the chance to select a random action with probability of $\epsilon$ as $t+1$ instead of the one with the highest expected reward. As for $\epsilon$ we have employed an exponentially decaying version, where $\epsilon = \epsilon_{min} + (\epsilon_{max} - \epsilon_{min})*\exp{(-decay * episode counter)}$.

On the other hand, regarding ONA hyperparameters, specifically \textit{motorbabling}, we have used default config in ONA v0.9.1 \cite{hammer2020opennars}. However, \textit{babblingops} have been changed due to the variety of number of available actions in each of the environments. Also, we have used \textit{setopname} to set allowed actions in ONA.

\begin{figure}[hp]%[t] %[h]
     \centering
     \begin{subfigure}{0.32\columnwidth}
         \centering
         \includegraphics[width=\columnwidth]{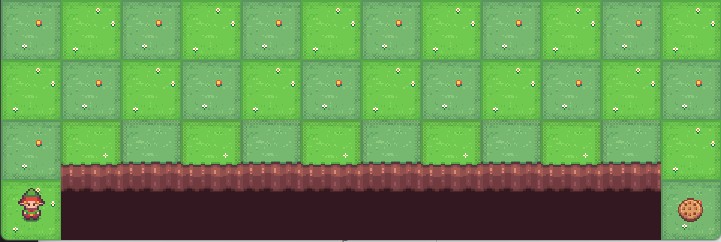}
         \caption{CliffWalking-v0}
         \label{CliffWalking-v0}
     \end{subfigure}
     \begin{subfigure}{0.32\columnwidth}
         \centering
         \includegraphics[width=\columnwidth]{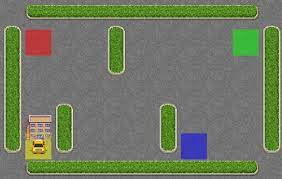}
         \caption{Taxi-v3}
         \label{Taxi-v3}
     \end{subfigure}
     \begin{subfigure}{0.21\columnwidth}
         \centering
         \includegraphics[width=\columnwidth]{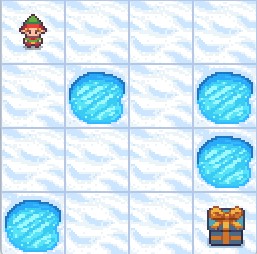}
         \caption{FrozenLake-v1 4x4}
         \label{FrozenLake-v1 4x4}
     \end{subfigure}
    \begin{subfigure}{0.32\columnwidth}
         \centering
         \includegraphics[width=\columnwidth]{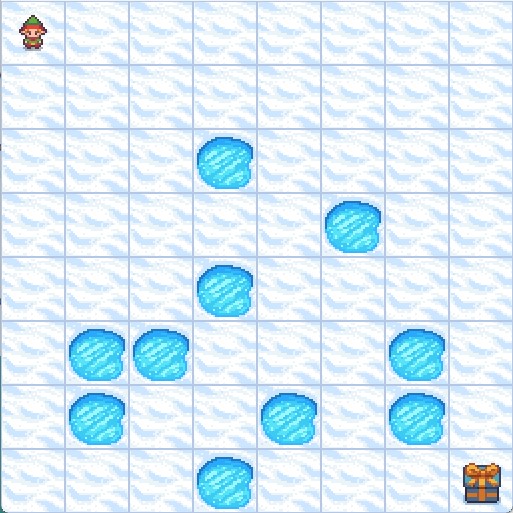}
         \caption{FrozenLake-v1 8x8}
         \label{FrozenLake-v1 8x8}
     \end{subfigure}
     \begin{subfigure}{0.18\columnwidth}
         \centering
         \includegraphics[width=\columnwidth]{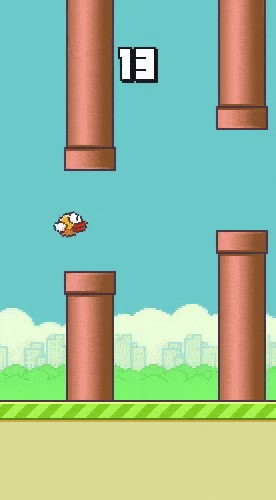}
         \caption{FlappyBird-v0}
         \label{FlappyBird-v0}
     \end{subfigure}
        \caption{OpenAI gym environments used as experiment tasks}
        \label{environments}
\end{figure}

We compare ONA to $Q$-Learning on a variety of challenging control tasks from the OpenAI gym benchmark suite \cite{brockman2016openai} (Figure \ref{environments}). It should be noted that since both  ONA to $Q$-Learning algorithms were developed specifically for discrete tasks, we have to map the FlappyBird-v0 observation space, but we did it in different ways for each algorithm, which we describe in detail in the following. Except for FlappyBird-v0, we use the original environments from \cite{brockman2016openai} without any modifications to the environment and reward. 

To make the practical comparison possible, as for ONA, the events will hold the same information as the corresponding states the Q-Learner will receive in the simulated experiments, except for FlappyBird-v0. To be more specific, as it is mentioned in \cite{hammer2021autonomy}, when $s$ is observation, it is interpreted by ONA
as event (s. $:|:$), and by the $Q$-Learner simply as current state. Then both algorithms suggest an operation/action by exploitation or sometimes randomly. After feeding the action to the environment, we will receive new observation, reward, and some info about reaching to the goal. The reward for the $Q$-Learner will be used without any change, while ONA receives event (G. $:|:$) when the task completely done. So there is no event if rewards are related anything except finishing the task. This of course assumes that the goal does not change, as else the $Q$-table entries would have to be re-learned, meaning the learned behavior would often not apply anymore. For the purposes of this work, and for a
fair comparison with Reinforcement Learning, the examples include a fixed objective.

\textbf{CliffWalking-v0:}
This environment is part of the Toy Text environments. This is a simple implementation of the Gridworld Cliff reinforcement learning task. Adapted from Example 6.6 (page 106) from Reinforcement Learning: An Introduction by Sutton and Barto \cite{sutton2018}. As shown in Figure \ref{CliffWalking-v0}, the board is a 4x12 matrix, with (using NumPy matrix indexing): [3, 0] as the start at bottom-left, [3, 11] as the goal at bottom-right, and [3, 1..10] as the cliff at bottom-center. If the agent steps on the cliff, it returns to the start. An episode terminates when the agent reaches the goal. There are 4 discrete deterministic actions: 0: move up, 1: move right, 2: move down, 3: move left.
As for observations, there are 3x12 + 1 possible states. In fact, the agent cannot be at the cliff, nor at the goal (as this results in the end of the episode). It remains all the positions of the first 3 rows plus the bottom-left cell. The observation is simply the current position encoded as flattened index. About reward, each time step incurs -1 reward, and stepping into the cliff incurs -100 reward. 

\textbf{Taxi-v3:} This environment is also part of the Toy Text environments form the work done by Tom Dietterich \cite{dietterich2000hierarchical}. In this environment, as depicted in Figure \ref{Taxi-v3}, there are four designated locations in the grid world indicated by R(ed), G(reen), Y(ellow), and B(lue). When the episode starts, the taxi starts off at a random square and the passenger is at a random location. The taxi drives to the passenger’s location, picks up the passenger, drives to the passenger’s destination (another one of the four specified locations), and then drops off the passenger. Once the passenger is dropped off, the episode ends.

There are 6 discrete deterministic actions: 0: move south, 1: move north, 2: move east, 3: move west, 4: pickup passenger, 5: drop off passenger. As for observations, there are 500 discrete states since there are 25 taxi positions, 5 possible locations of the passenger (including the case when the passenger is in the taxi), and 4 destination locations. Note that there are 400 states that can actually be reached during an episode. The missing states correspond to situations in which the passenger is at the same location as their destination, as this typically signals the end of an episode. Four additional states can be observed right after a successful episodes, when both the passenger and the taxi are at the destination. This gives a total of 404 reachable discrete states. Each state space is represented by the tuple: (taxi\_row, taxi\_col, passenger\_location, destination). An observation is an integer that encodes the corresponding state. Passenger locations are 0: R(ed), 1: G(green), 2: Y(fellow), 3: B(lue), and 4: in taxi. Destinations also are 0: R(ed), 1: G(green), 2: Y(fellow), 3: B(lue). Agent will be rewarded -1 per step unless other reward is triggered, +20 for delivering passenger, and -10 in the case of executing “pickup” and “drop-off” actions illegally.

\textbf{FrozenLake-v1:}
As depicted in Figures \ref{FrozenLake-v1 4x4} and \ref{FrozenLake-v1 8x8}, Frozen lake involves crossing a frozen lake from Start (S) to Goal (G) without falling into any Holes (H) by walking over the Frozen (F) lake. The agent may not always move in the intended direction due to the slippery nature of the frozen lake. The agent takes a 1-element vector for actions. The action space is (dir), where dir decides direction to move in which can be: 0: LEFT, 1: DOWN, 2: RIGHT, 3: UP. In addition, the observation is a value representing the agent’s current position as current\_row * nrows + current\_col (where both the row and col start at 0). For example, the goal position in the 4x4 map can be calculated as follows: 3 x 4 + 3 = 15. The number of possible observations is dependent on the size of the map. For example, the 4x4 map has 16 possible observations, while in the case of 8x8 is 64. Reward schedule is as follow:
\begin{itemize}
    \item Reach goal(G): +1,
    \item Reach hole(H): 0,
    \item Reach frozen(F): 0,
\end{itemize}

While one could specify custom map for frozen lake by "desc" argument, for example, desc=["SFFF", "FHFH", "FFFH", "HFFG"], we have used preloaded maps which are shown in  \ref{FrozenLake-v1 4x4} and \ref{FrozenLake-v1 8x8}, known with 4x4 and 8x8 IDs, respectively. Thanks to "is\_slippery" argument, we could even have a non-deterministic environment, which is very interesting to see how algorithms behave in such a problem. Particularly, if "is\_slippery" is True, the agent moves in intended direction with probability of 1/3 else moves in either perpendicular direction with equal probability of 1/3 in both directions. For example, if action is left and "is\_slippery" is True, then: P(move left)= 1/3, P(move up) = 1/3, P(move down) = 1/3.

\textbf{FlappyBird-v0:} As shown in \ref{FlappyBird-v0}, the last environment is the Flappy Bird game. The implementation of the game's logic and graphics is based on the FlapPyBird project\footnote{https://github.com/sourabhv/FlapPyBird}. This environment yields simple numerical information about the game's state as observations. Specifically, the yielded attributes are: ($O_1$) the horizontal distance to the next pipe, and ($O_2$) the difference between the player's $y$ position and the next hole's $y$ position. The reward received by the agent in each step is one, if the bird is  still alive. Also a score point is obtained every time the bird passes a pipe. The action taken by the agent could be 0: "do nothing" and 1: "flap". We have access to a status report also which will be true if the game is over, and otherwise is false. Since observation space is continues, to be able to use ONA and $Q$-Learning, we have mapped it to a discrete space. Specifically, as for ONA, the event is "round(100x$O_1$)\_round(1000x$O_2$). $:|:$", which could be for instance "138\_-4. $:|:$". However, since for defining $Q$-table, the states should correspond to the specific row, we have to subtly change the mapping to "$|$round(100x$O_1$)$|$+$|$round(1000x$O_2$)$|$", which results "142", for our instance. However, one could find a better way to do this mapping.

In the next section, we examine both algorithms performances in detail on all these seven tasks.

\section{Results and Discussion} \label{section3}
\begin{figure}[H]%[t] %[h]
     \centering
     \begin{subfigure}{0.32\columnwidth}
         \centering
         \includegraphics[width=\columnwidth]{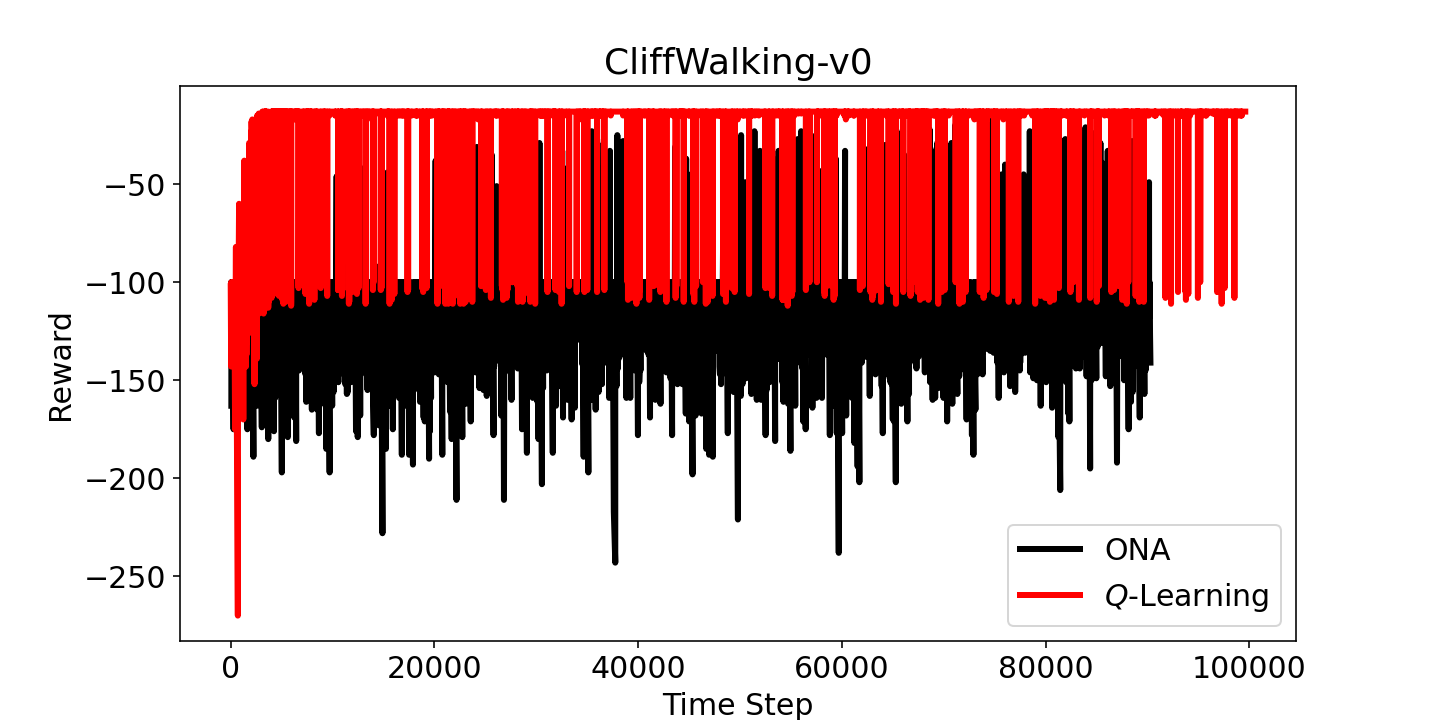}
         \caption{CliffWalking-v0}
         \label{Reward_vs_Time_Step_CliffWalking-v0}
     \end{subfigure}
     \begin{subfigure}{0.32\columnwidth}
         \centering
         \includegraphics[width=\columnwidth]{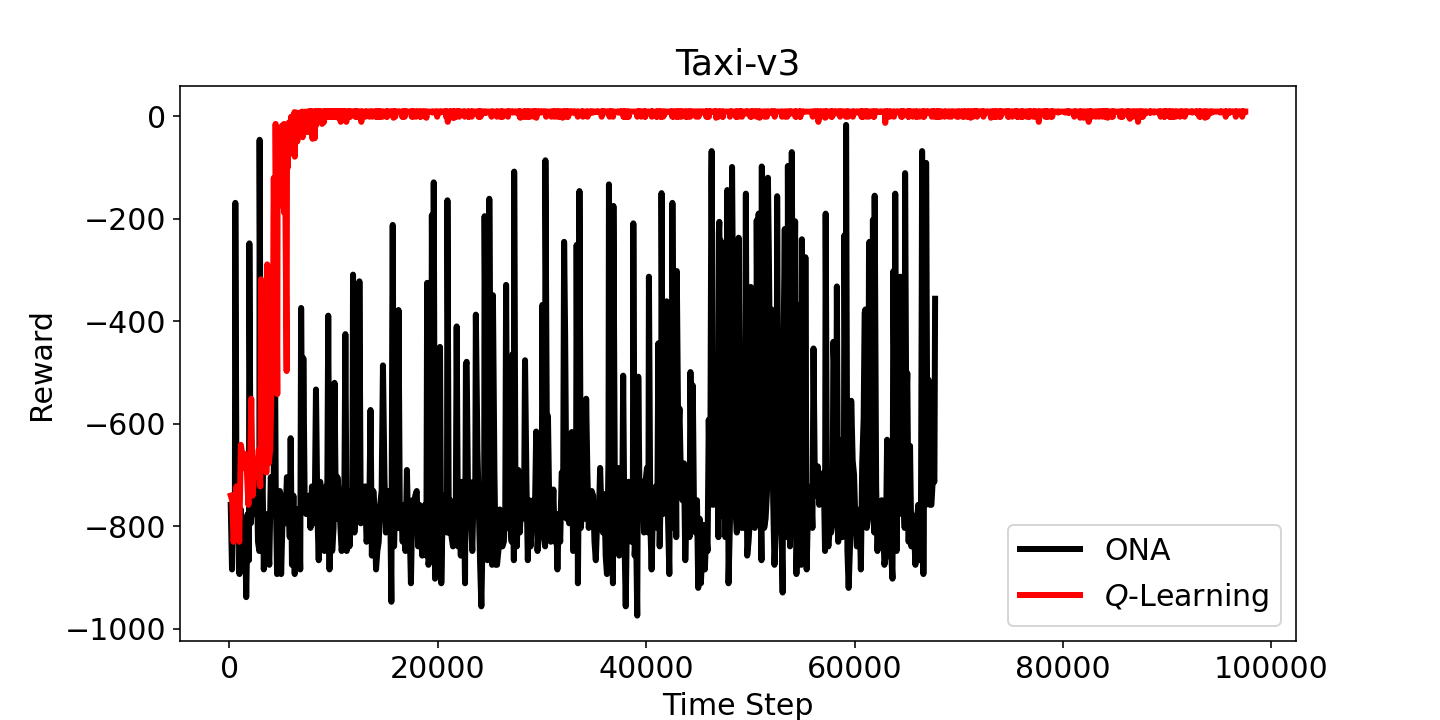}
         \caption{Taxi-v3}
         \label{Reward_vs_Time_Step_Taxi-v3}
     \end{subfigure}
     \begin{subfigure}{0.32\columnwidth}
         \centering
         \includegraphics[width=\columnwidth]{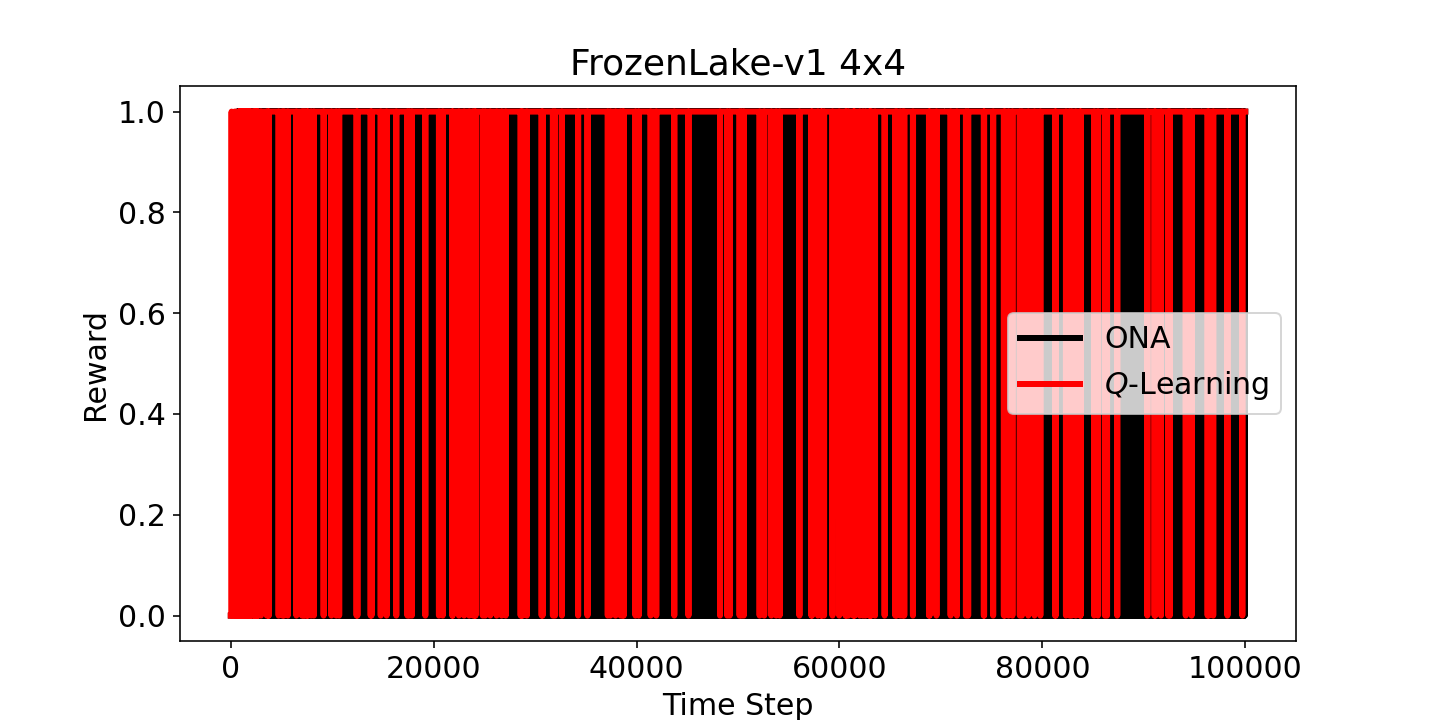}
         \caption{FrozenLake-v1 4x4}
         \label{Reward_vs_Time_Step_FrozenLake-v1 4x4}
     \end{subfigure}
     \begin{subfigure}{0.32\columnwidth}
         \centering
         \includegraphics[width=\columnwidth]{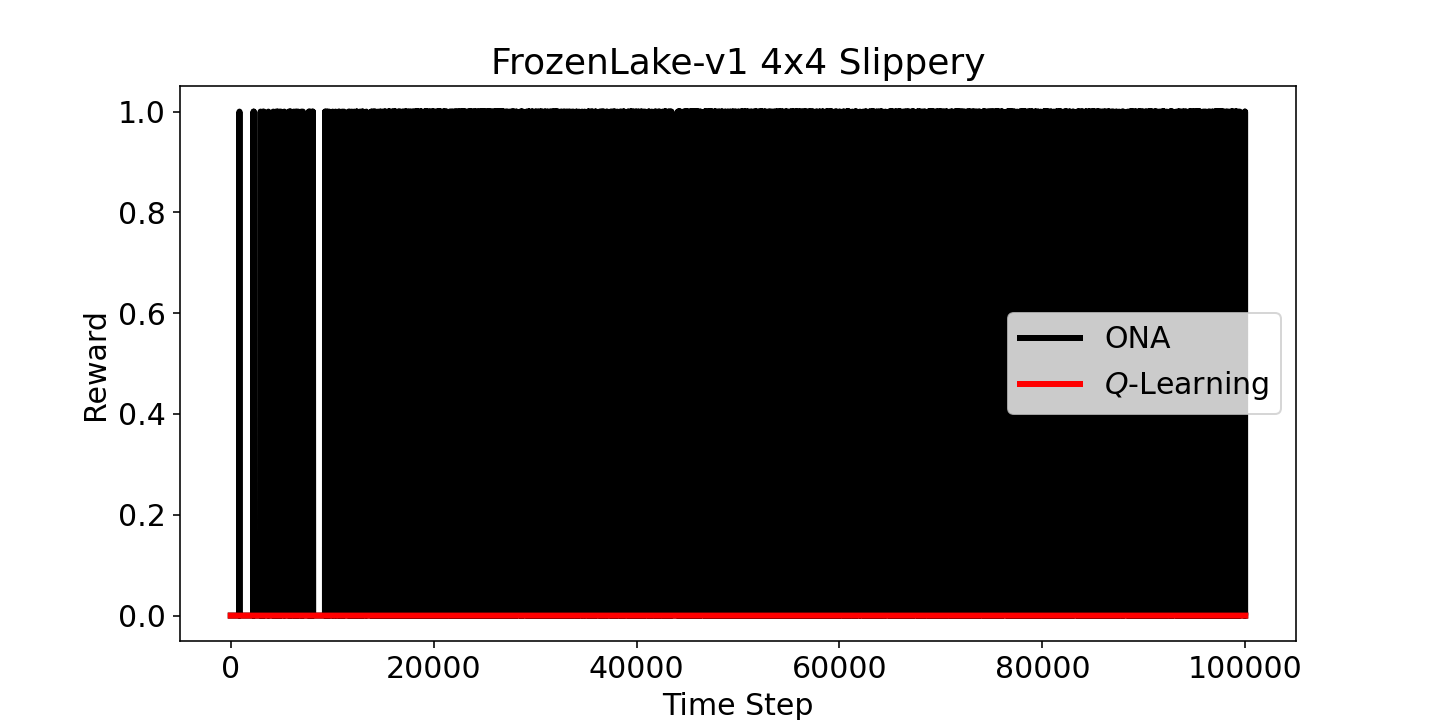}
         \caption{FrozenLake-v1 4x4 Slippery}
         \label{Reward_vs_Time_Step_FrozenLake-v1_4x4_Slippery}
     \end{subfigure}
    \begin{subfigure}{0.32\columnwidth}
         \centering
         \includegraphics[width=\columnwidth]{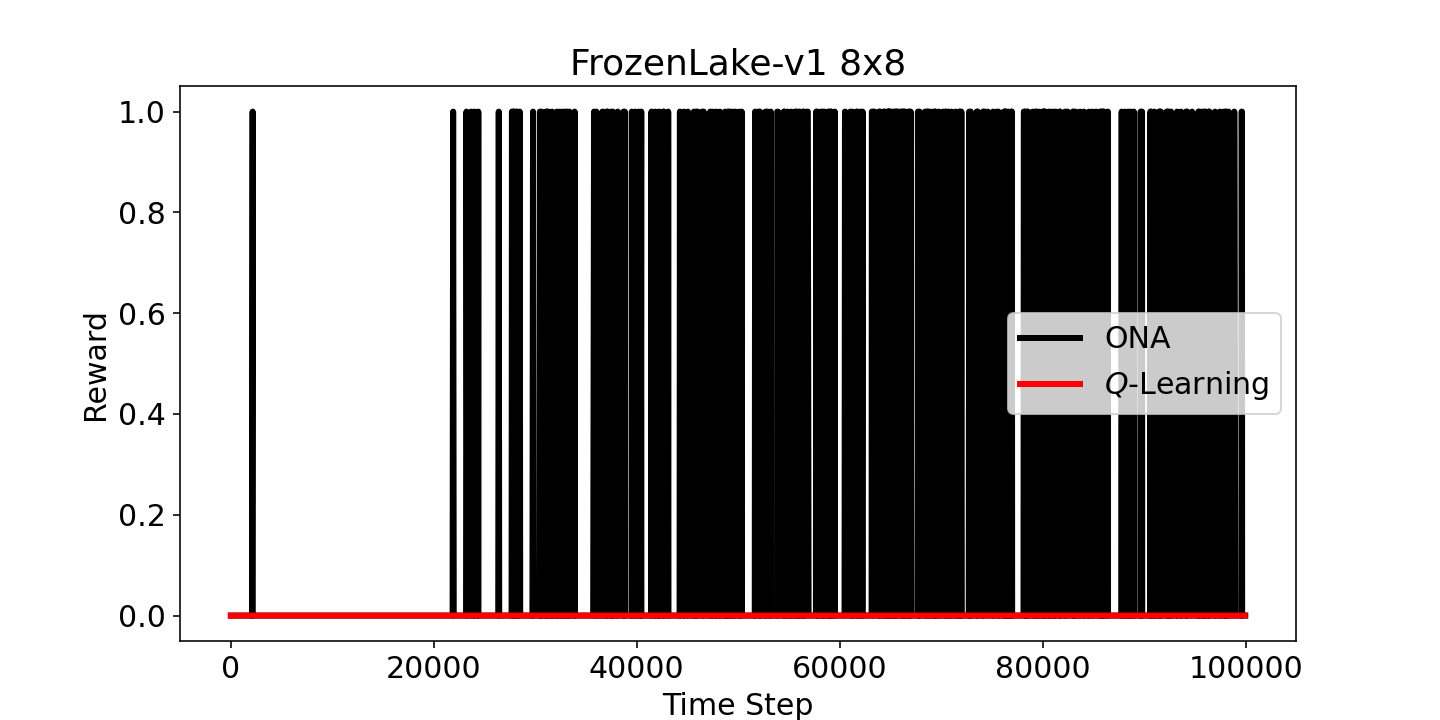}
         \caption{FrozenLake-v1 8x8}
         \label{Reward_vs_Time_Step_FrozenLake-v1 8x8}
     \end{subfigure}
    \begin{subfigure}{0.32\columnwidth}
         \centering
         \includegraphics[width=\columnwidth]{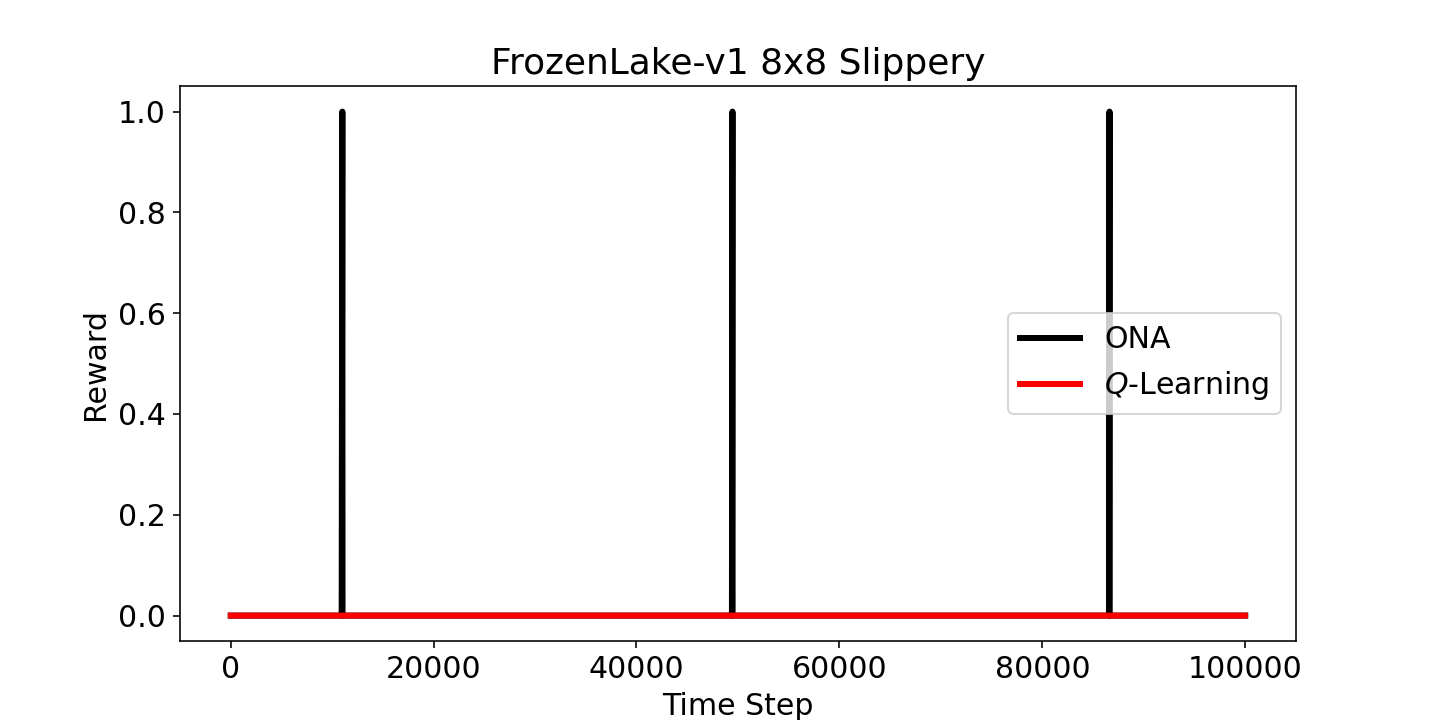}
         \caption{FrozenLake-v1 8x8 Slippery}
         \label{Reward_vs_Time_Step_FrozenLake-v1_8x8_Slippery}
     \end{subfigure}
     \begin{subfigure}{0.32\columnwidth}
         \centering
         \includegraphics[width=\columnwidth]{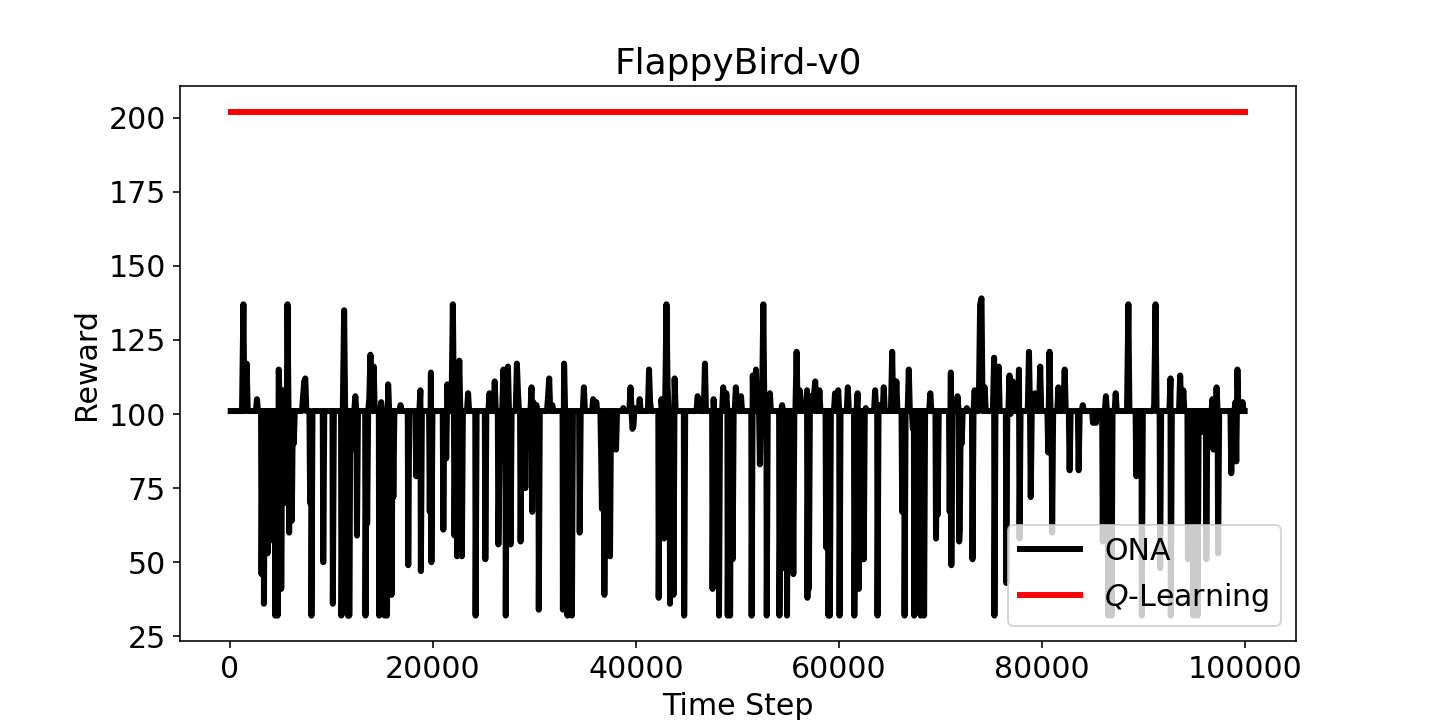}
         \caption{FlappyBird-v0}
         \label{Reward_vs_Time_Step_FlappyBird-v0}
     \end{subfigure}
        \caption{Reward vs. Time steps. The reward is measured at time steps where the episode ends (by reaching the goal, truncating the episode length, falling into the hole, falling from the cliff, hitting the pipe.)}
        \label{Reward_vs_Time Step}
\end{figure}

\begin{figure}[H]%[t] %[h]
     \centering
     \begin{subfigure}{0.32\columnwidth}
         \centering
         \includegraphics[width=\columnwidth]{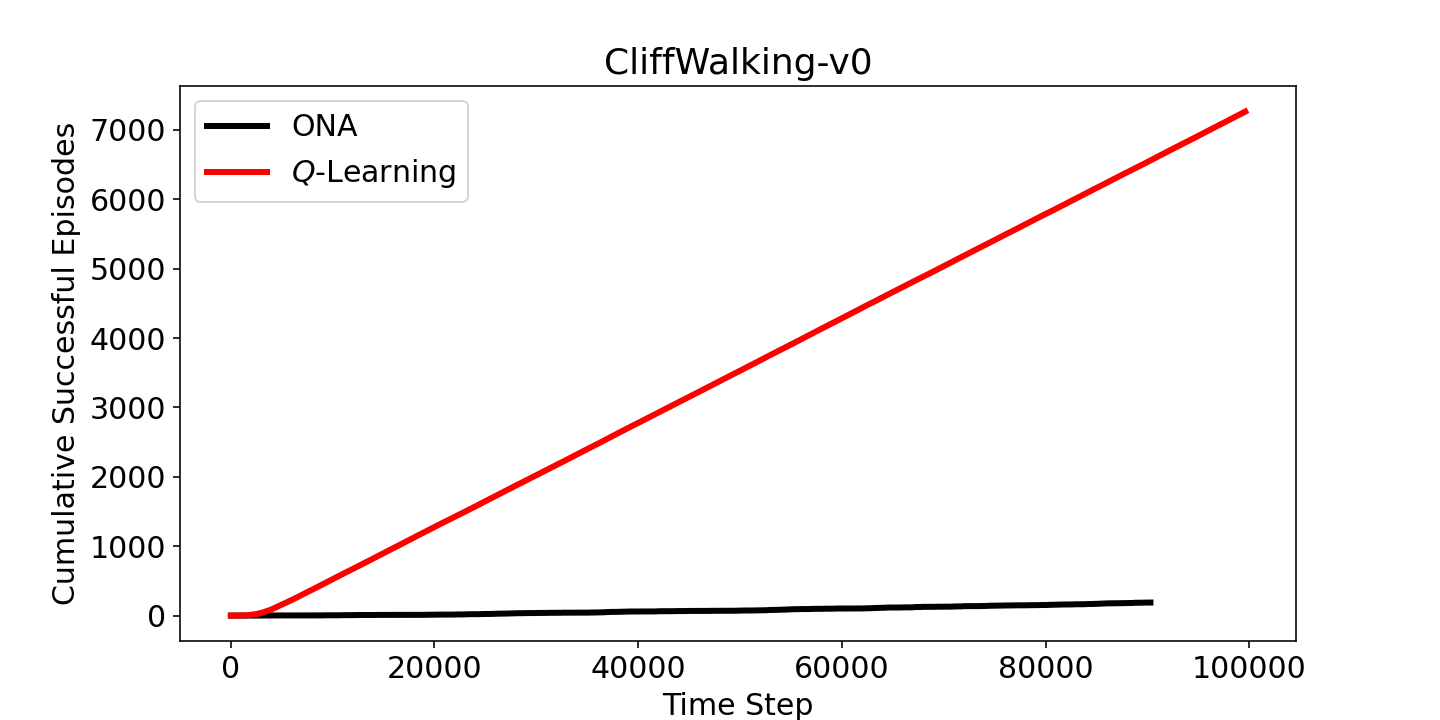}
         \caption{CliffWalking-v0}
         \label{Cumulative_Successful_Episodes_vs_Time_Step_CliffWalking-v0}
     \end{subfigure}
     \begin{subfigure}{0.32\columnwidth}
         \centering
         \includegraphics[width=\columnwidth]{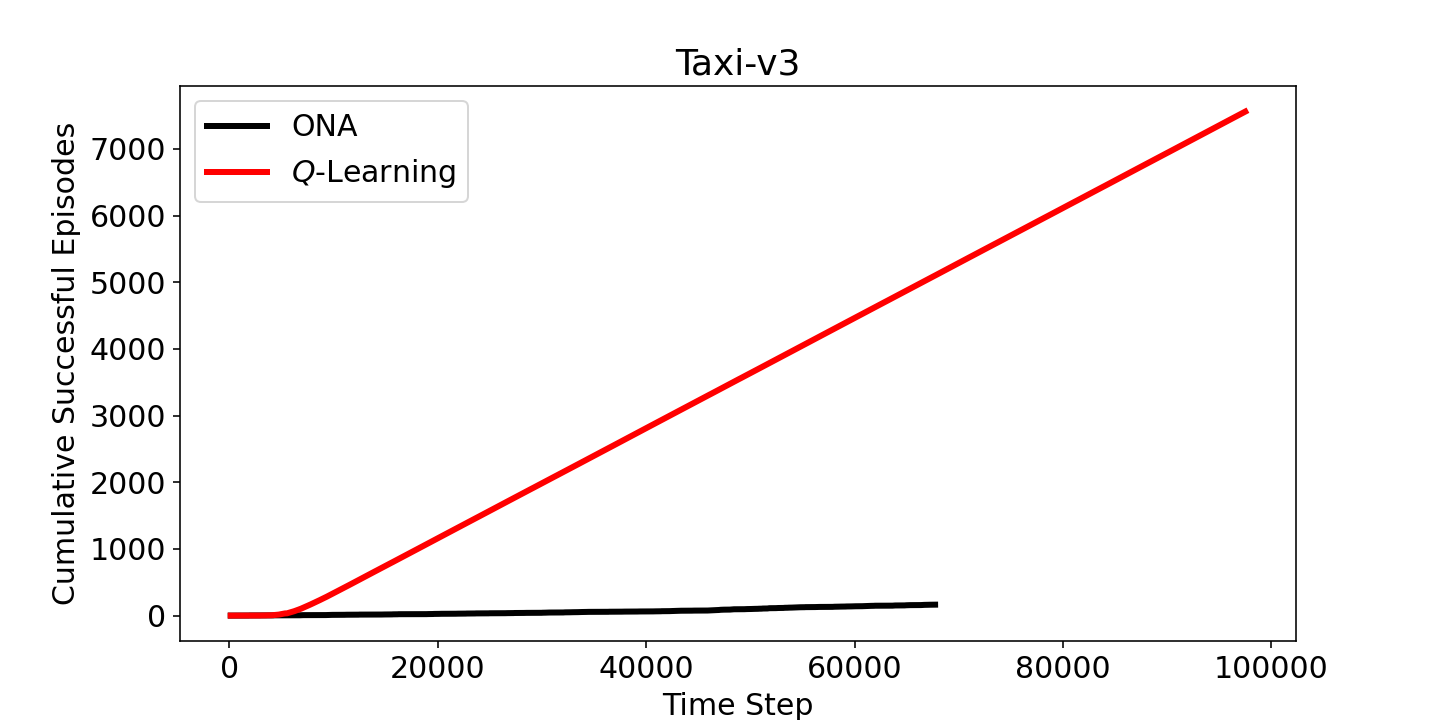}
         \caption{Taxi-v3}
         \label{Cumulative_Successful_Episodes_vs_Time_Step_Taxi-v3}
     \end{subfigure}
     \begin{subfigure}{0.32\columnwidth}
         \centering
         \includegraphics[width=\columnwidth]{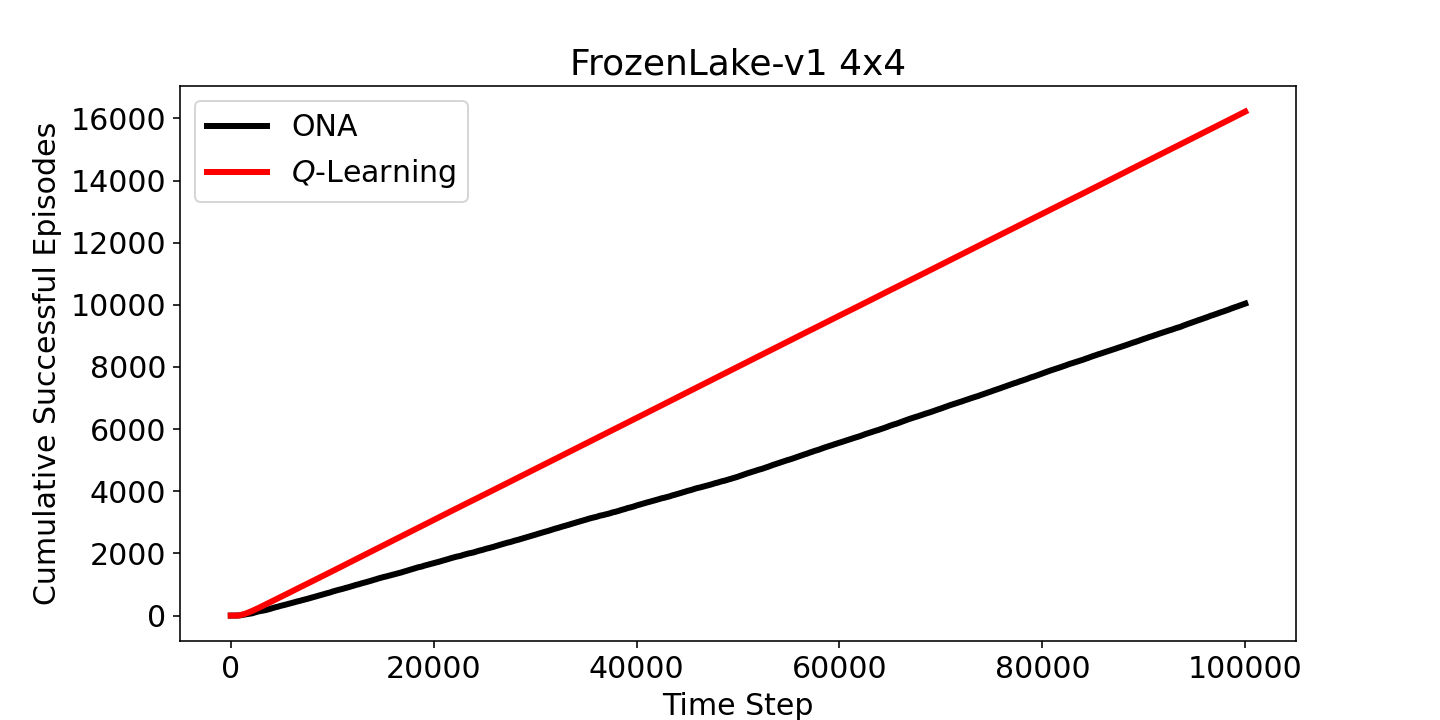}
         \caption{FrozenLake-v1 4x4}
         \label{Cumulative_Successful_Episodes_vs_Time_Step_FrozenLake-v1 4x4}
     \end{subfigure}
     \begin{subfigure}{0.32\columnwidth}
         \centering
         \includegraphics[width=\columnwidth]{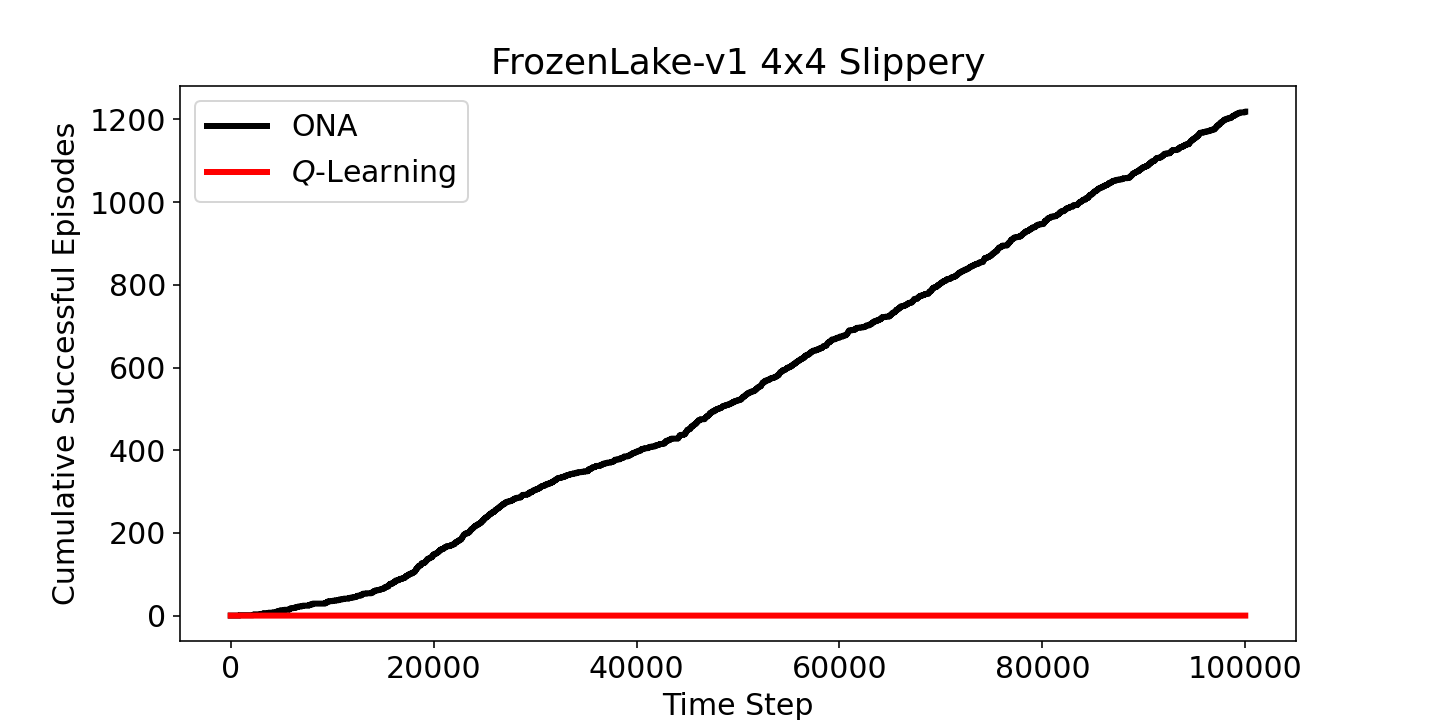}
         \caption{FrozenLake-v1 4x4 Slippery}
         \label{Cumulative_Successful_Episodes_vs_Time_Step_FrozenLake-v1_4x4_Slippery}
     \end{subfigure}
    \begin{subfigure}{0.32\columnwidth}
         \centering
         \includegraphics[width=\columnwidth]{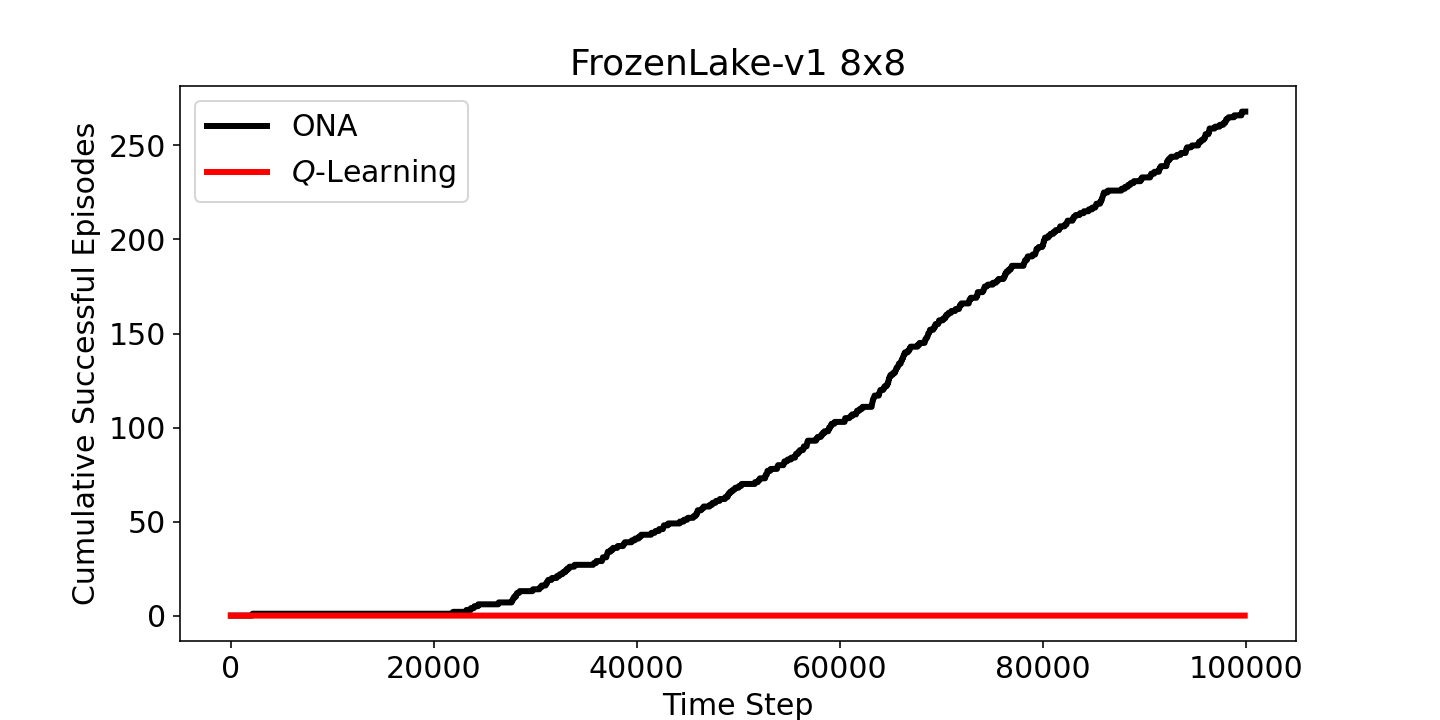}
         \caption{FrozenLake-v1 8x8}
         \label{Cumulative_Successful_Episodes_vs_Time_Step_FrozenLake-v1 8x8}
     \end{subfigure}
    \begin{subfigure}{0.32\columnwidth}
         \centering
         \includegraphics[width=\columnwidth]{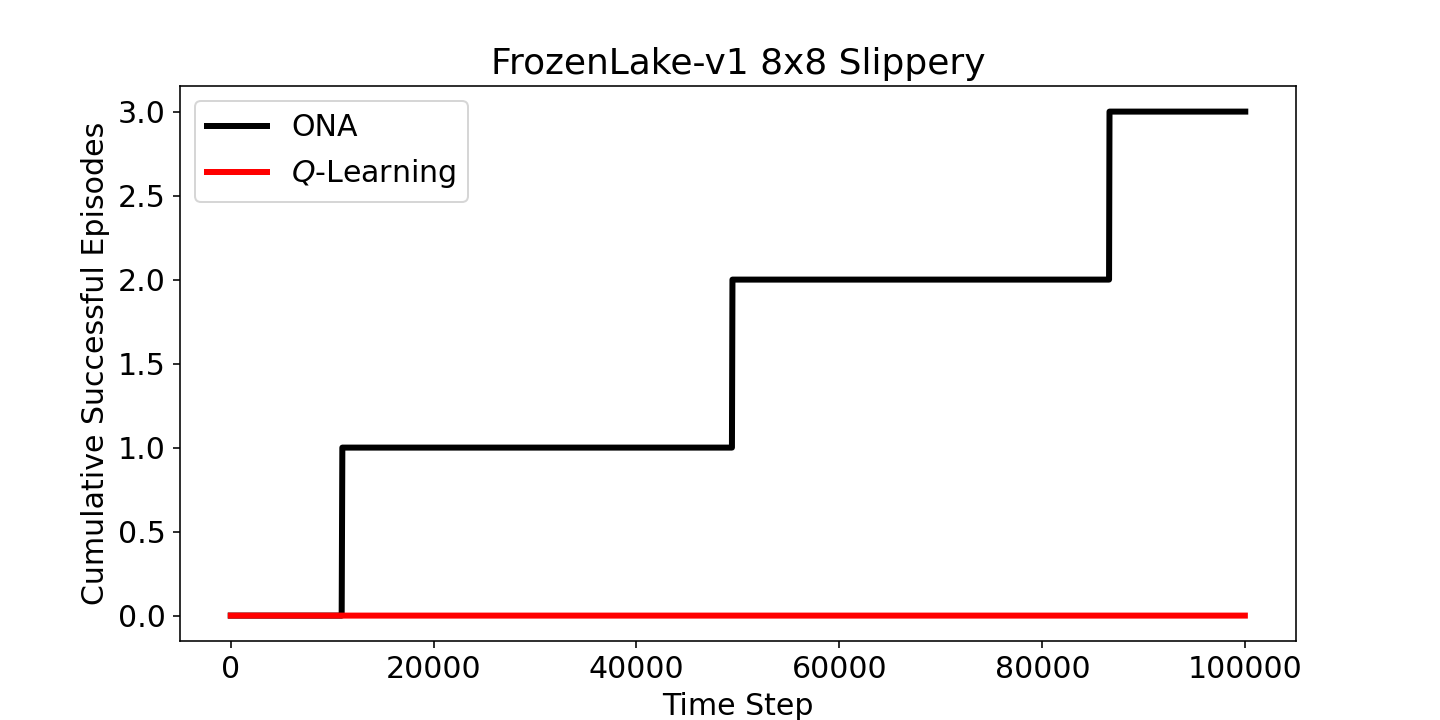}
         \caption{FrozenLake-v1 8x8 Slippery}
         \label{Cumulative_Successful_Episodes_vs_Time_Step_FrozenLake-v1_8x8_Slippery}
     \end{subfigure}
        \caption{Cumulative Successful Episodes vs. Time steps.}
        \label{Cumulative_Successful_Episodes_vs_Time Step}
\end{figure}

\begin{figure}[H]%[t] %[h]
     \centering
     \begin{subfigure}{0.32\columnwidth}
         \centering
         \includegraphics[width=\columnwidth]{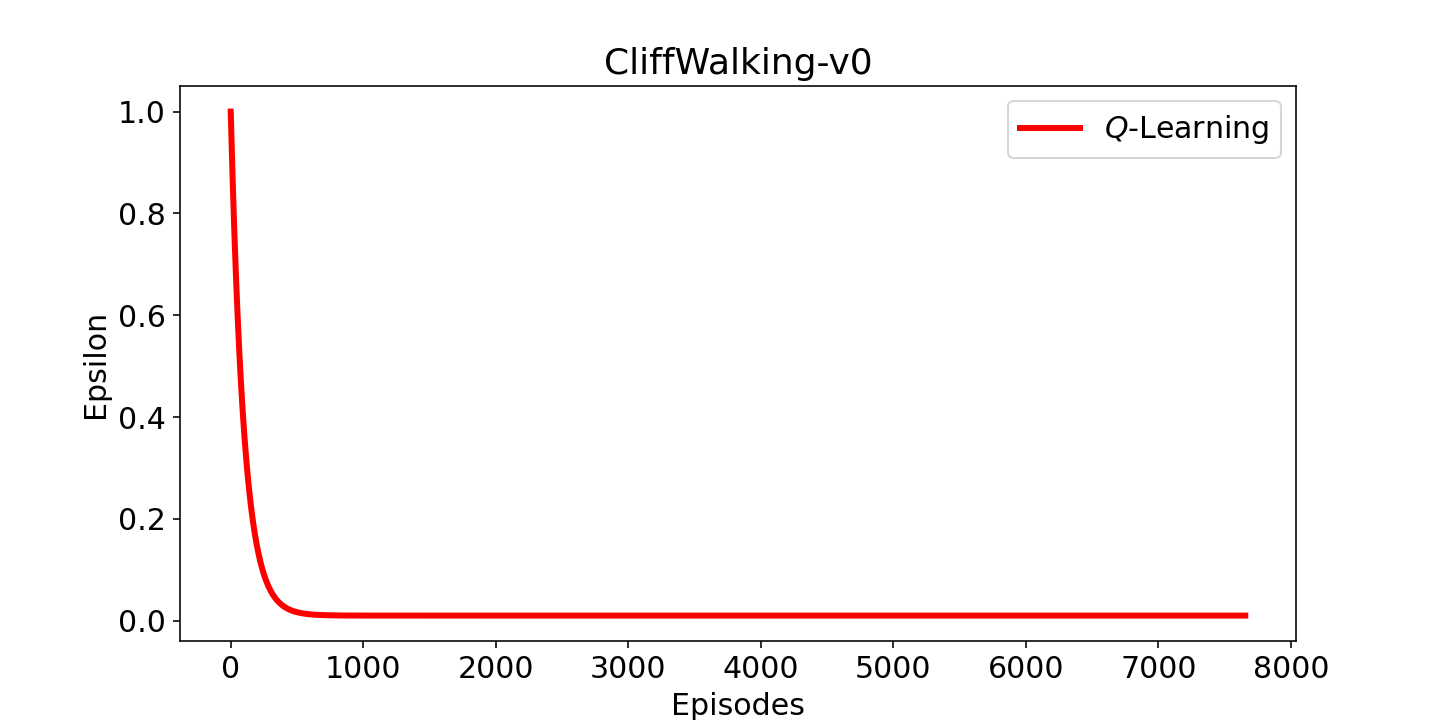}
         \caption{CliffWalking-v0}
         \label{Epsilon_vs_Episodes_CliffWalking-v0}
     \end{subfigure}
     \begin{subfigure}{0.32\columnwidth}
         \centering
         \includegraphics[width=\columnwidth]{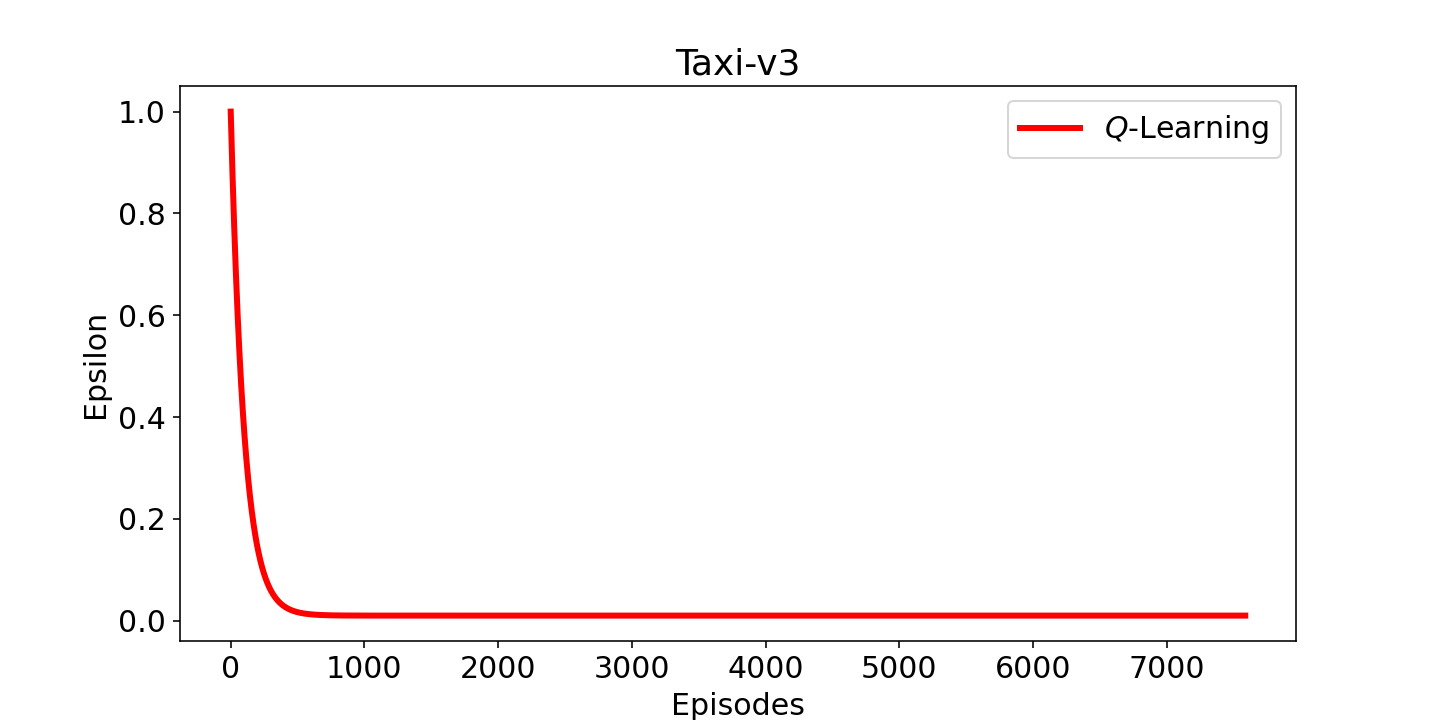}
         \caption{Taxi-v3}
         \label{Epsilon_vs_Episodes_Taxi-v3}
     \end{subfigure}
     \begin{subfigure}{0.32\columnwidth}
         \centering
         \includegraphics[width=\columnwidth]{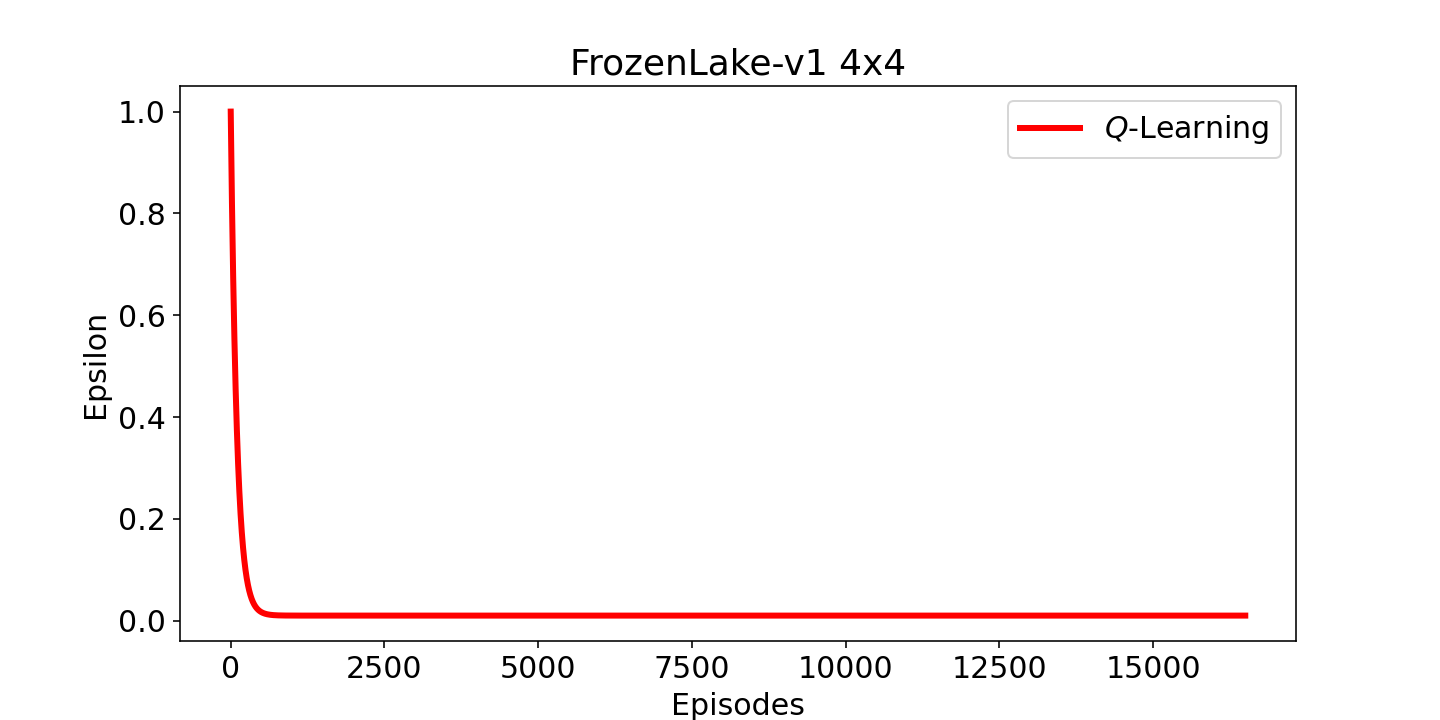}
         \caption{FrozenLake-v1 4x4}
         \label{Epsilon_vs_Episodes_FrozenLake-v1 4x4}
     \end{subfigure}
     \begin{subfigure}{0.32\columnwidth}
         \centering
         \includegraphics[width=\columnwidth]{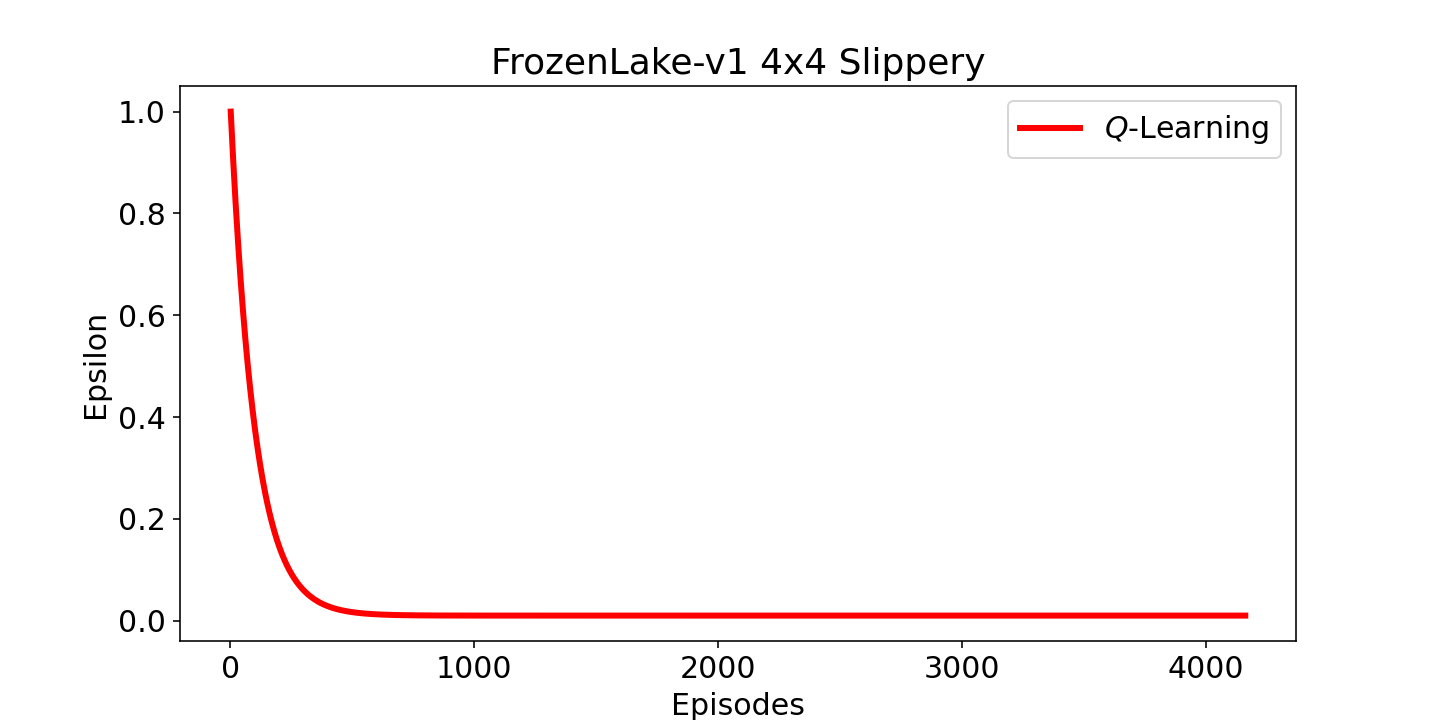}
         \caption{FrozenLake-v1 4x4 Slippery}
         \label{Epsilon_vs_Episodes_FrozenLake-v1_4x4_Slippery}
     \end{subfigure}
    \begin{subfigure}{0.32\columnwidth}
         \centering
         \includegraphics[width=\columnwidth]{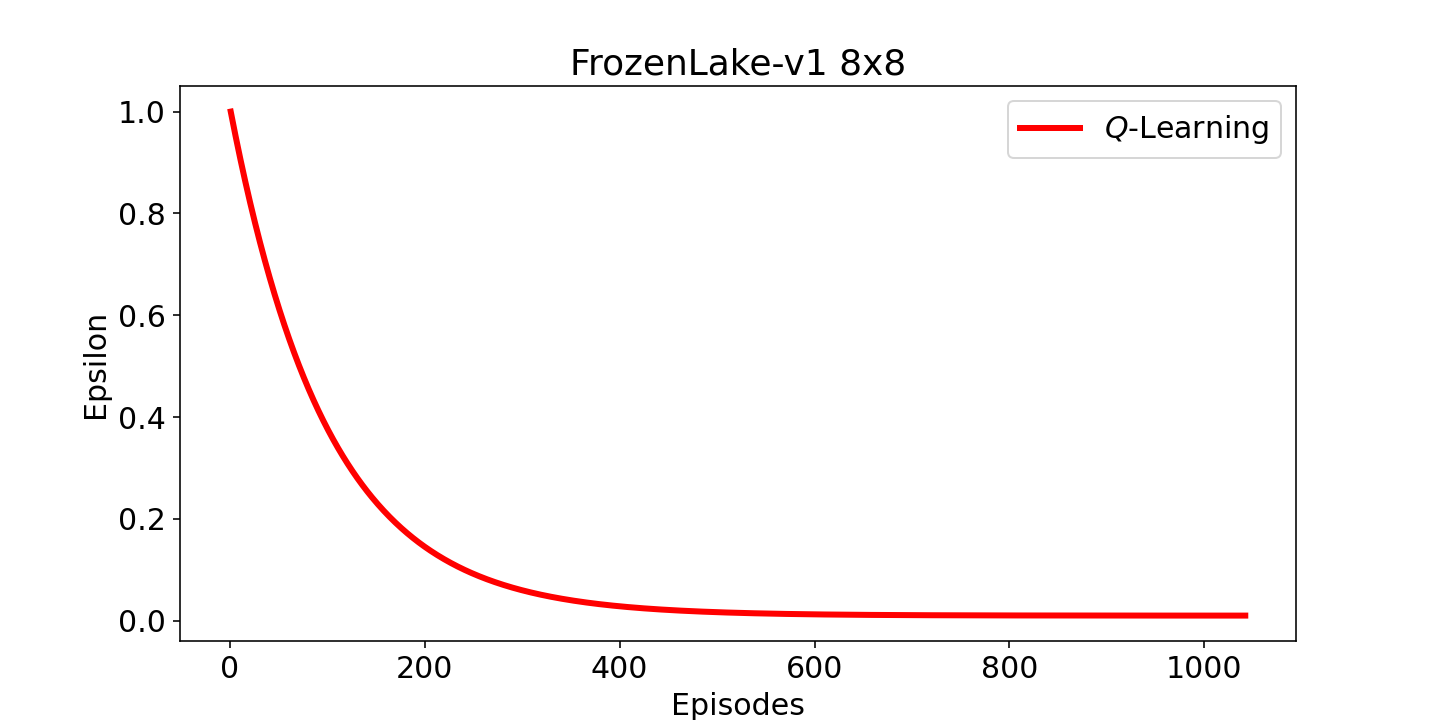}
         \caption{FrozenLake-v1 8x8}
         \label{Epsilon_vs_Episodes_FrozenLake-v1 8x8}
     \end{subfigure}
    \begin{subfigure}{0.32\columnwidth}
         \centering
         \includegraphics[width=\columnwidth]{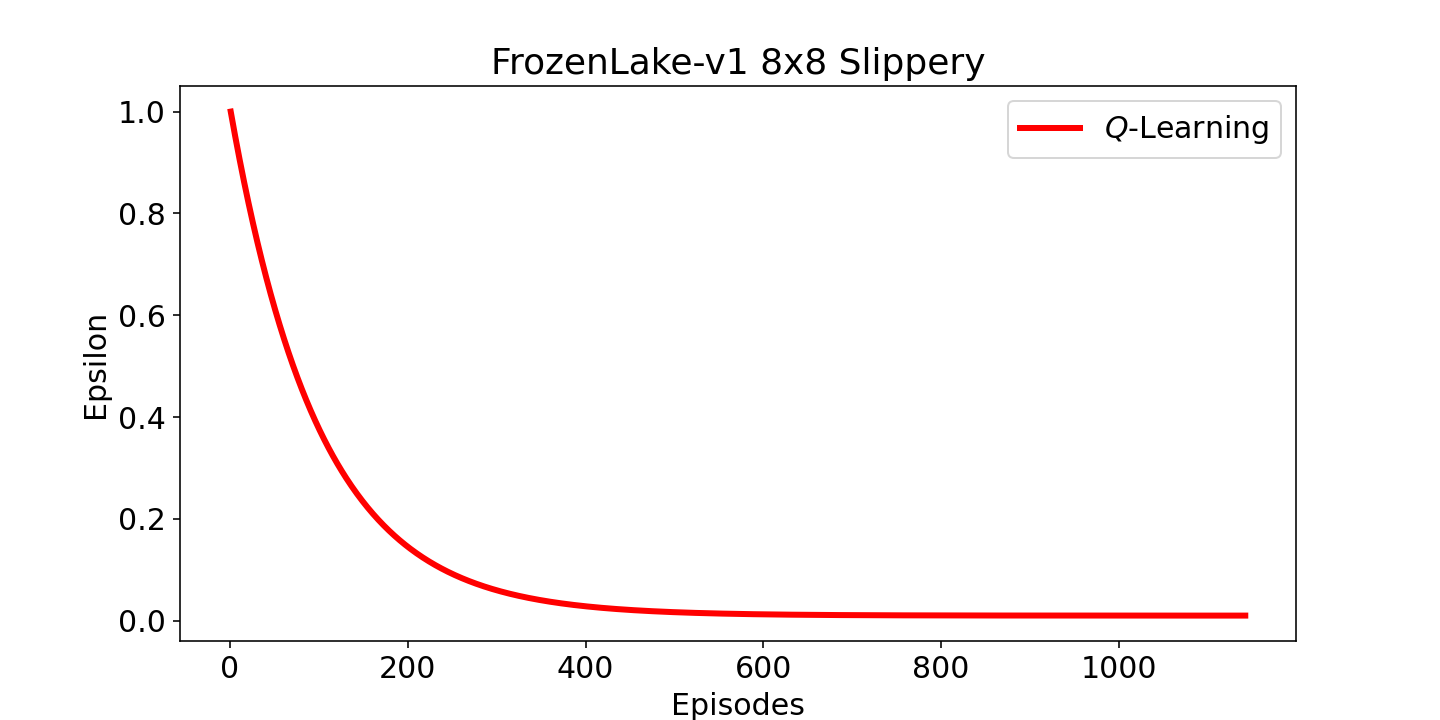}
         \caption{FrozenLake-v1 8x8 Slippery}
         \label{Epsilon_vs_Episodes_FrozenLake-v1_8x8_Slippery}
     \end{subfigure}
     \begin{subfigure}{0.32\columnwidth}
         \centering
         \includegraphics[width=\columnwidth]{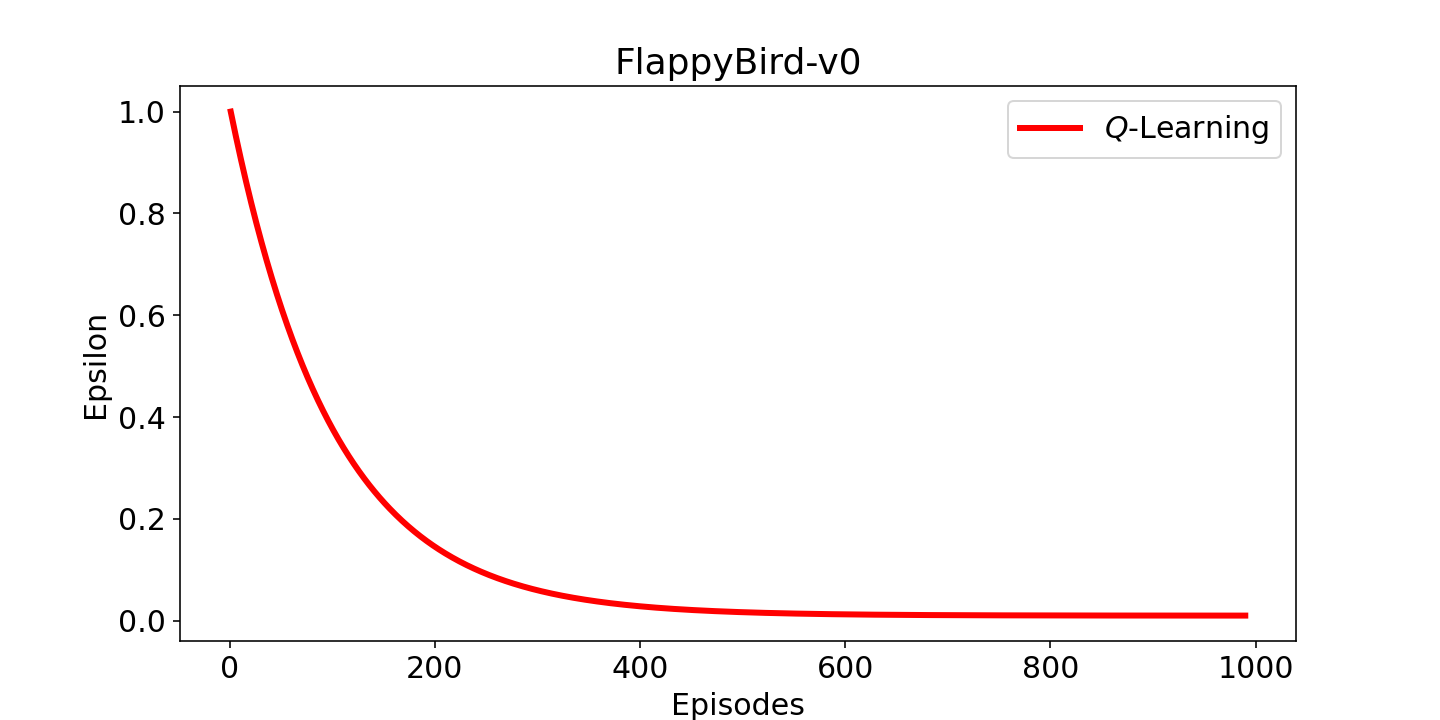}
         \caption{FlappyBird-v0}
         \label{Epsilon_vs_Episodes_FlappyBird-v0}
     \end{subfigure}
        \caption{Decaying $\epsilon$ used for $Q$-Learning vs. Episodes. The $\epsilon$ value has been updated just before stating a new episode. }
        \label{Epsilon_vs_Episodes}
\end{figure}

\begin{figure}[H]%[t] %[h]
     \centering
     \begin{subfigure}{0.32\columnwidth}
         \centering
         \includegraphics[width=\columnwidth]{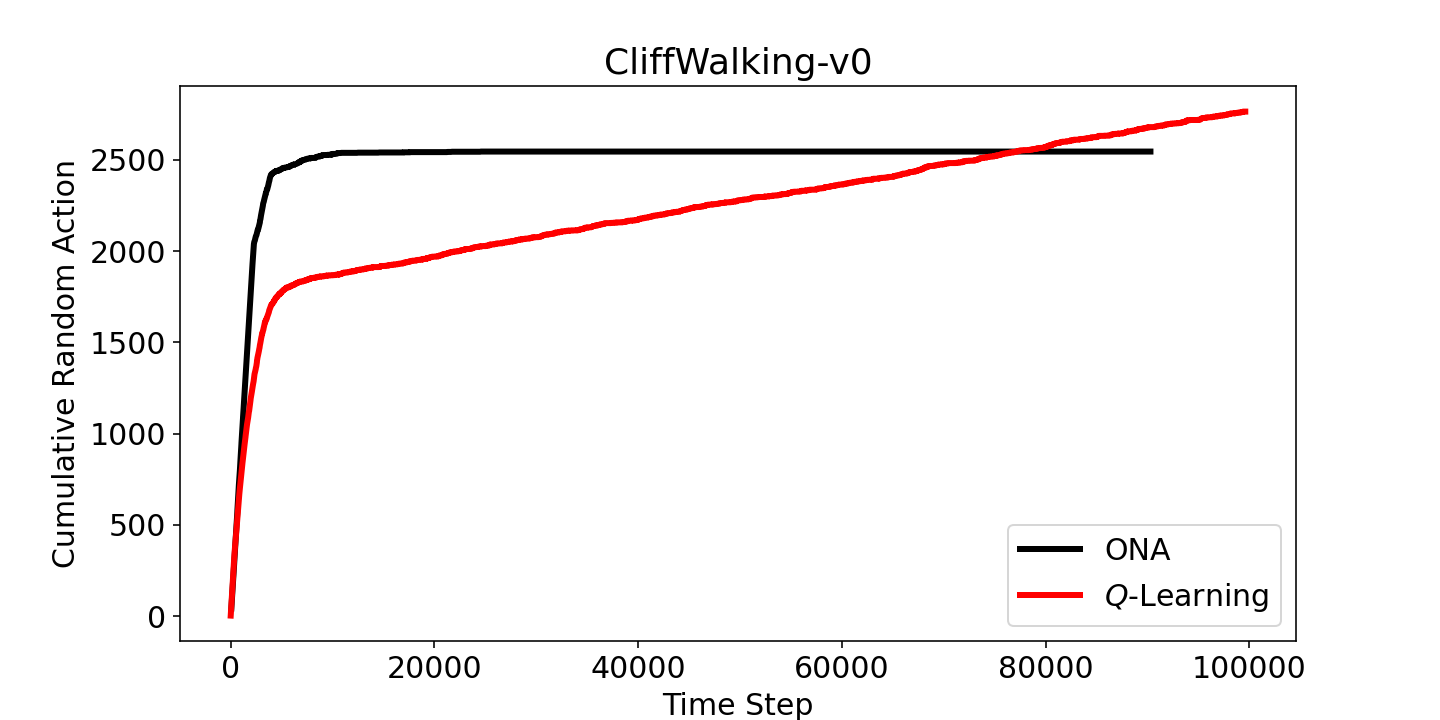}
         \caption{CliffWalking-v0}
         \label{Cumulative_Random_Action_vs_Time_Step_CliffWalking-v0}
     \end{subfigure}
     \begin{subfigure}{0.32\columnwidth}
         \centering
         \includegraphics[width=\columnwidth]{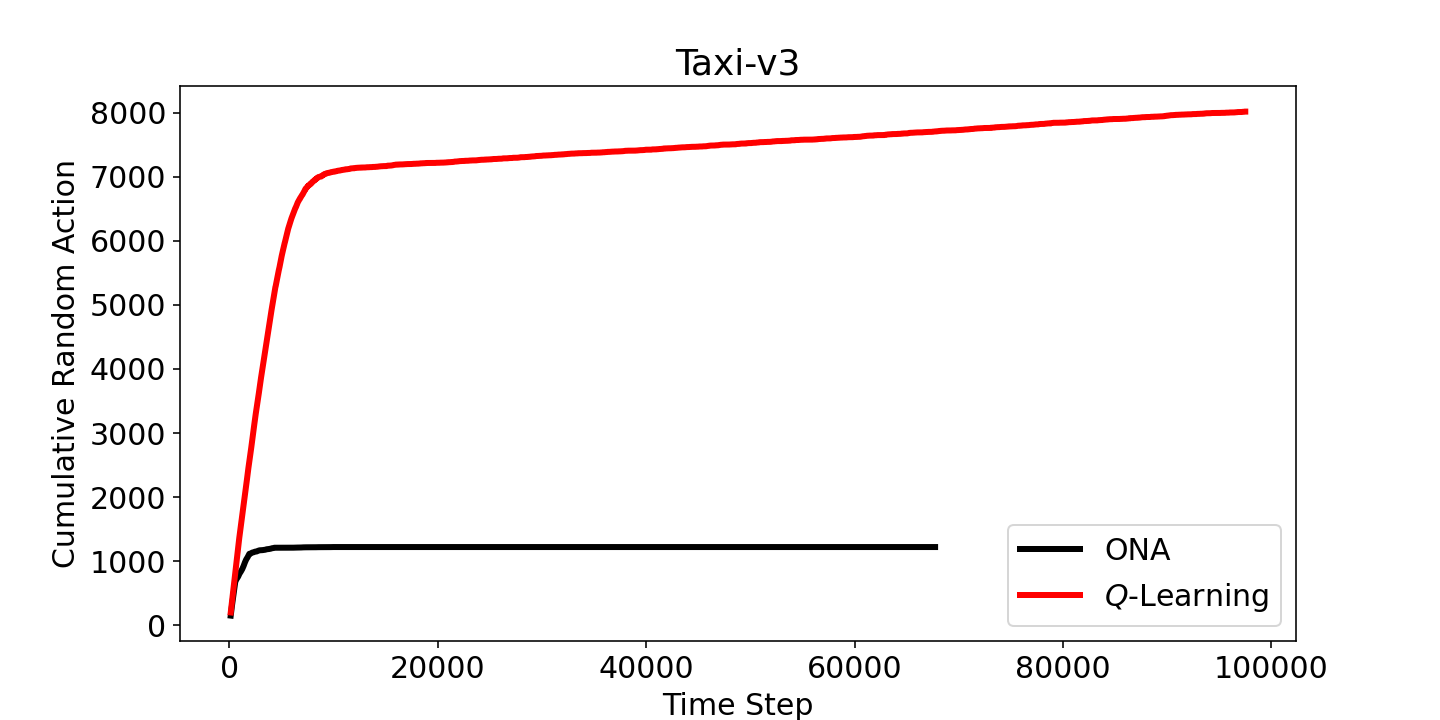}
         \caption{Taxi-v3}
         \label{Cumulative_Random_Action_vs_Time_Step_Taxi-v3}
     \end{subfigure}
     \begin{subfigure}{0.32\columnwidth}
         \centering
         \includegraphics[width=\columnwidth]{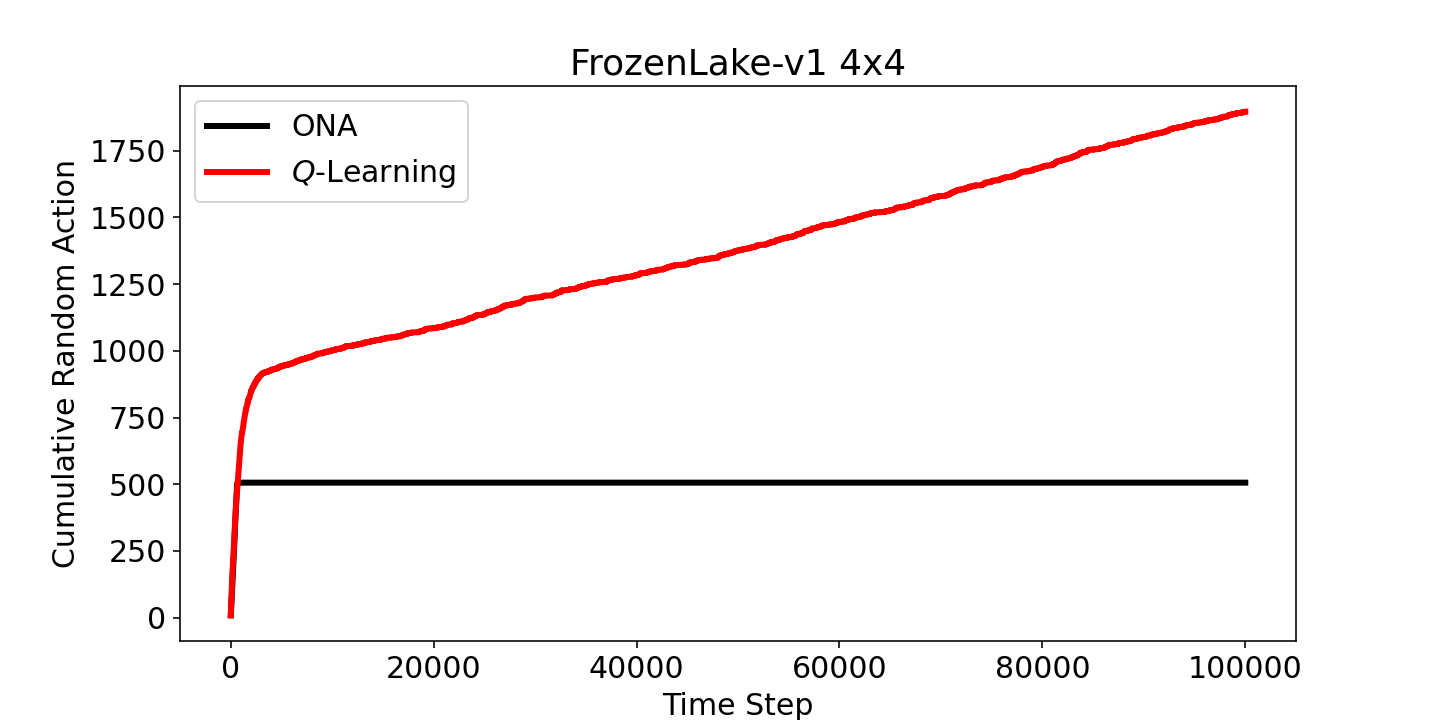}
         \caption{FrozenLake-v1 4x4}
         \label{Cumulative_Random_Action_vs_Time_Step_FrozenLake-v1 4x4}
     \end{subfigure}
     \begin{subfigure}{0.32\columnwidth}
         \centering
         \includegraphics[width=\columnwidth]{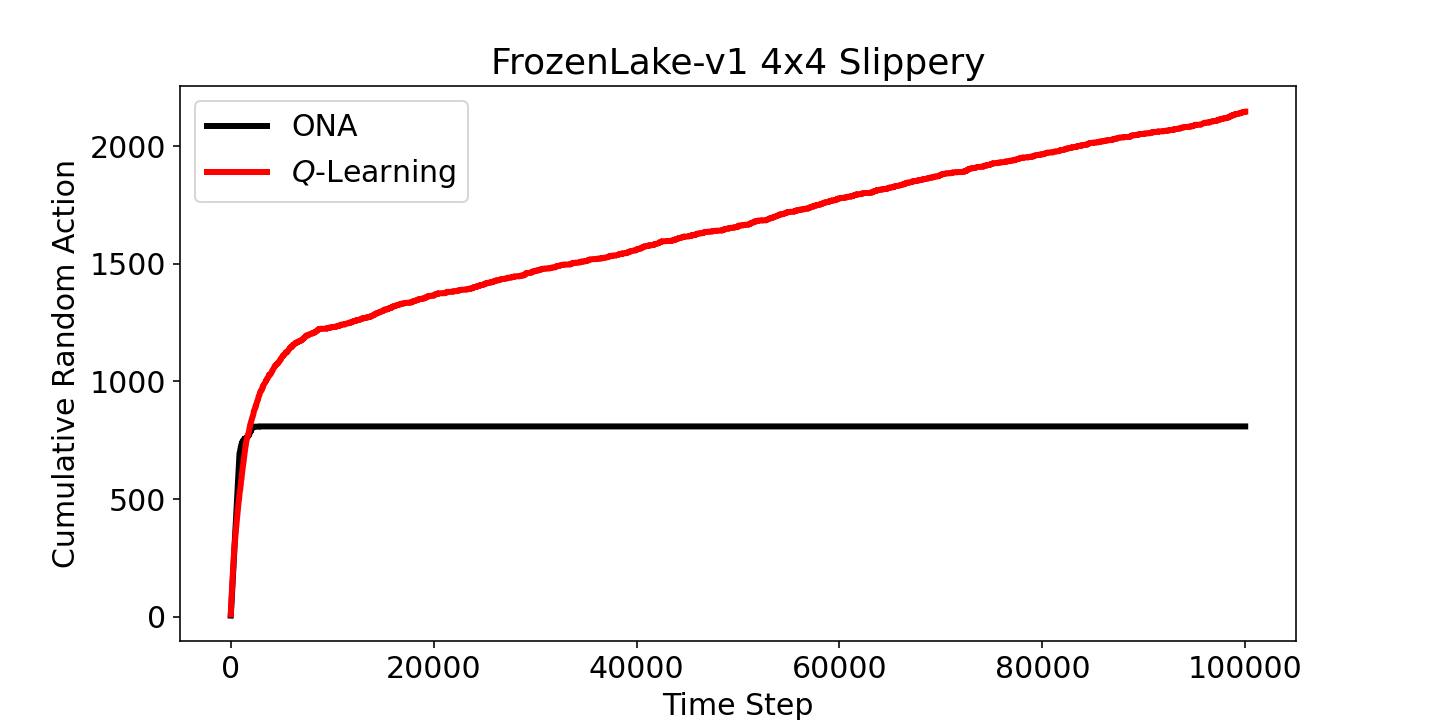}
         \caption{FrozenLake-v1 4x4 Slippery}
         \label{Cumulative_Random_Action_vs_Time_Step_FrozenLake-v1_4x4_Slippery}
     \end{subfigure}
    \begin{subfigure}{0.32\columnwidth}
         \centering
         \includegraphics[width=\columnwidth]{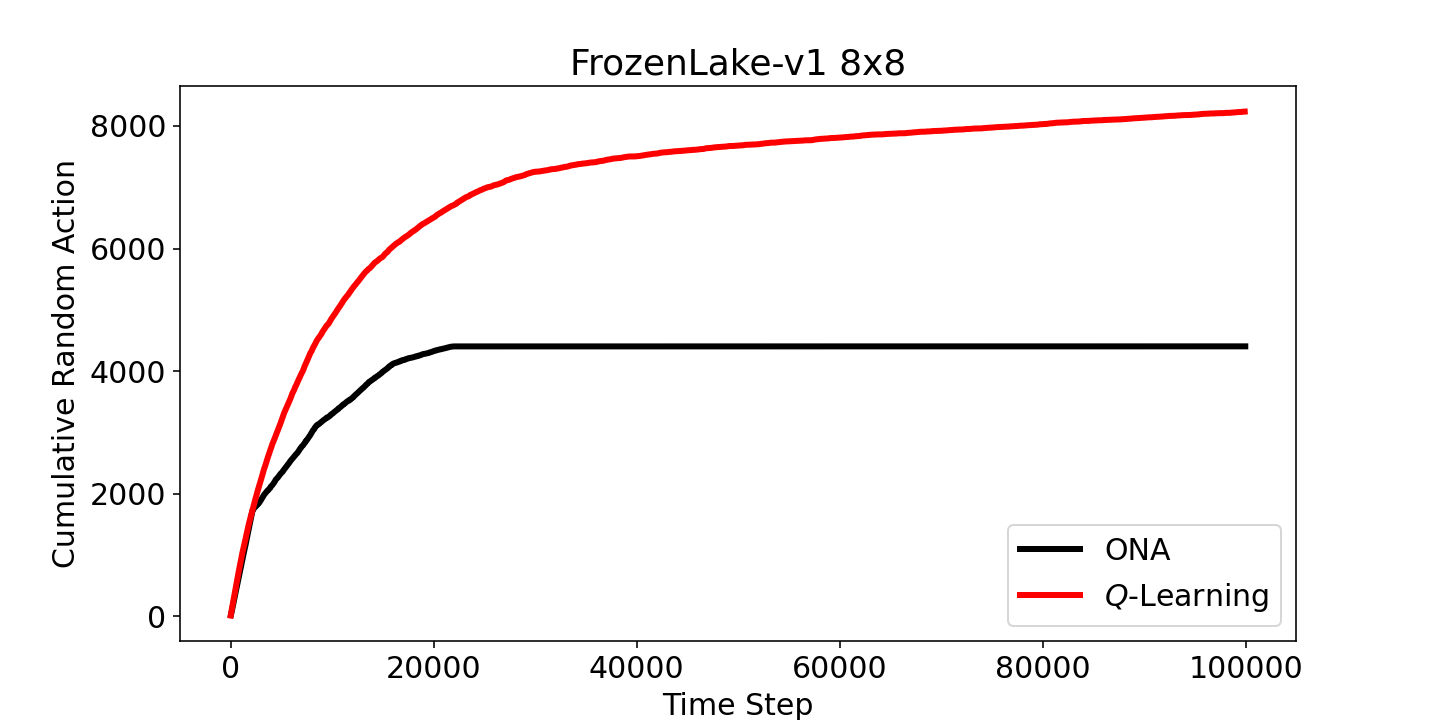}
         \caption{FrozenLake-v1 8x8}
         \label{Cumulative_Random_Action_vs_Time_Step_FrozenLake-v1 8x8}
     \end{subfigure}
    \begin{subfigure}{0.32\columnwidth}
         \centering
         \includegraphics[width=\columnwidth]{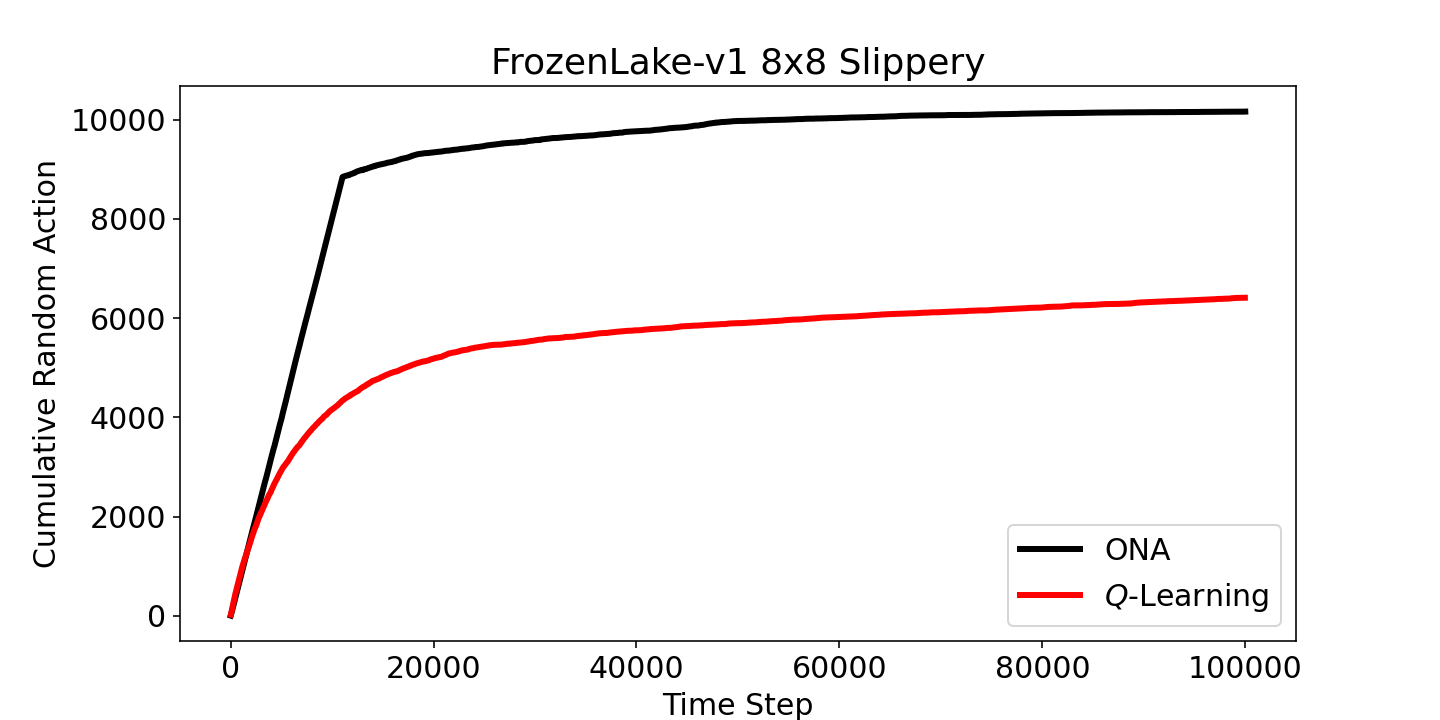}
         \caption{FrozenLake-v1 8x8 Slippery}
         \label{Cumulative_Random_Action_vs_Time_Step_FrozenLake-v1_8x8_Slippery}
     \end{subfigure}
     \begin{subfigure}{0.32\columnwidth}
         \centering
         \includegraphics[width=\columnwidth]{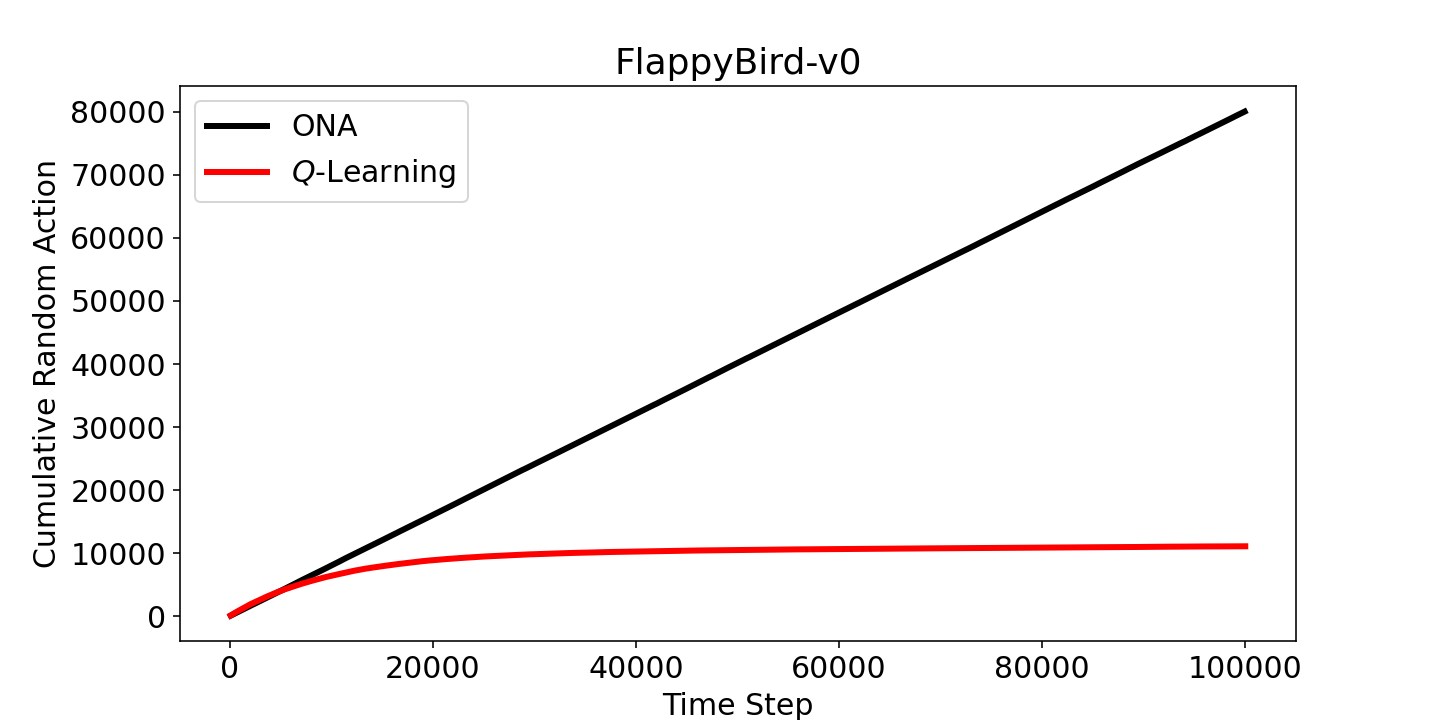}
         \caption{FlappyBird-v0}
         \label{Cumulative_Random_Action_vs_Time_Step_FlappyBird-v0}
     \end{subfigure}
        \caption{Cumulative Random Action vs. Time steps.}
        \label{Cumulative_Random_Action_vs_Time Step}
\end{figure}

\begin{figure}[H]%[t] %[h]
     \centering
     \begin{subfigure}{0.32\columnwidth}
         \centering
         \includegraphics[width=\columnwidth]{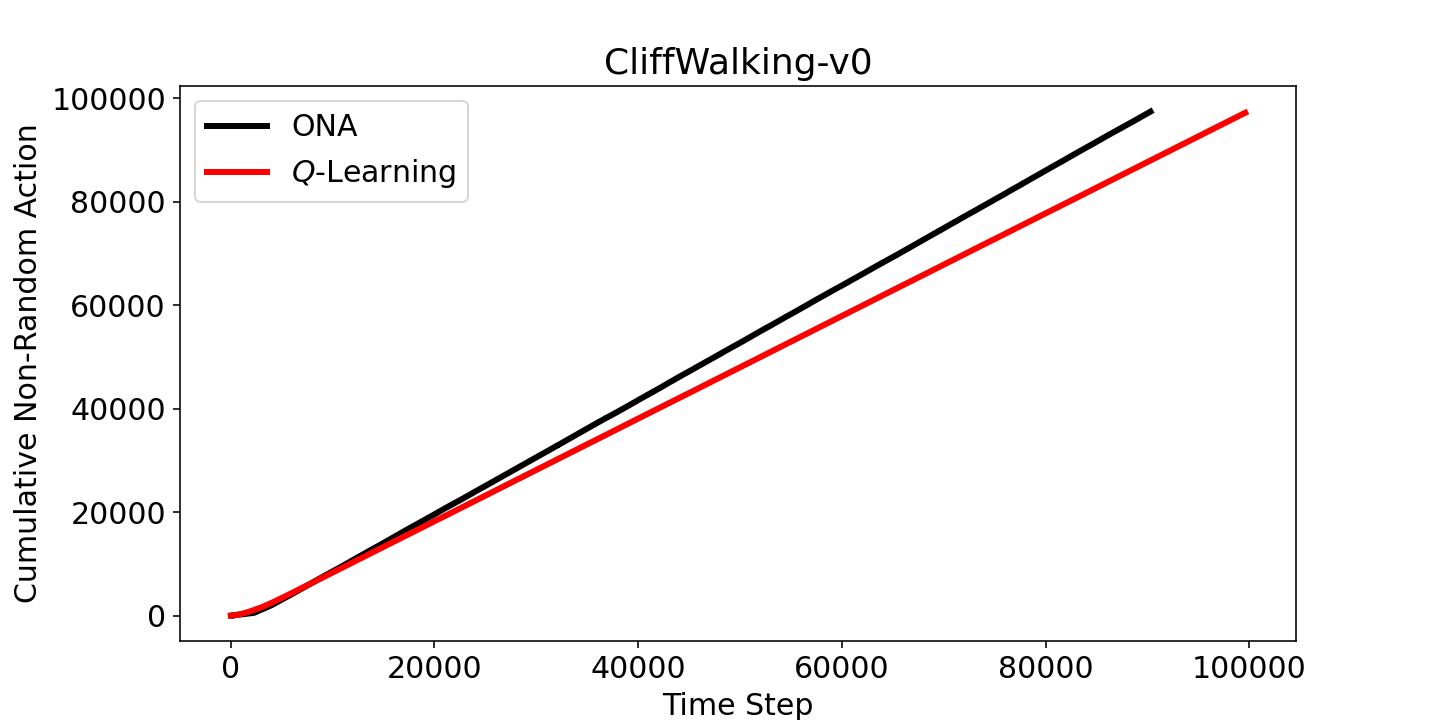}
         \caption{CliffWalking-v0}
         \label{Cumulative_Non-Random_Action_vs_Time_Step_CliffWalking-v0}
     \end{subfigure}
     \begin{subfigure}{0.32\columnwidth}
         \centering
         \includegraphics[width=\columnwidth]{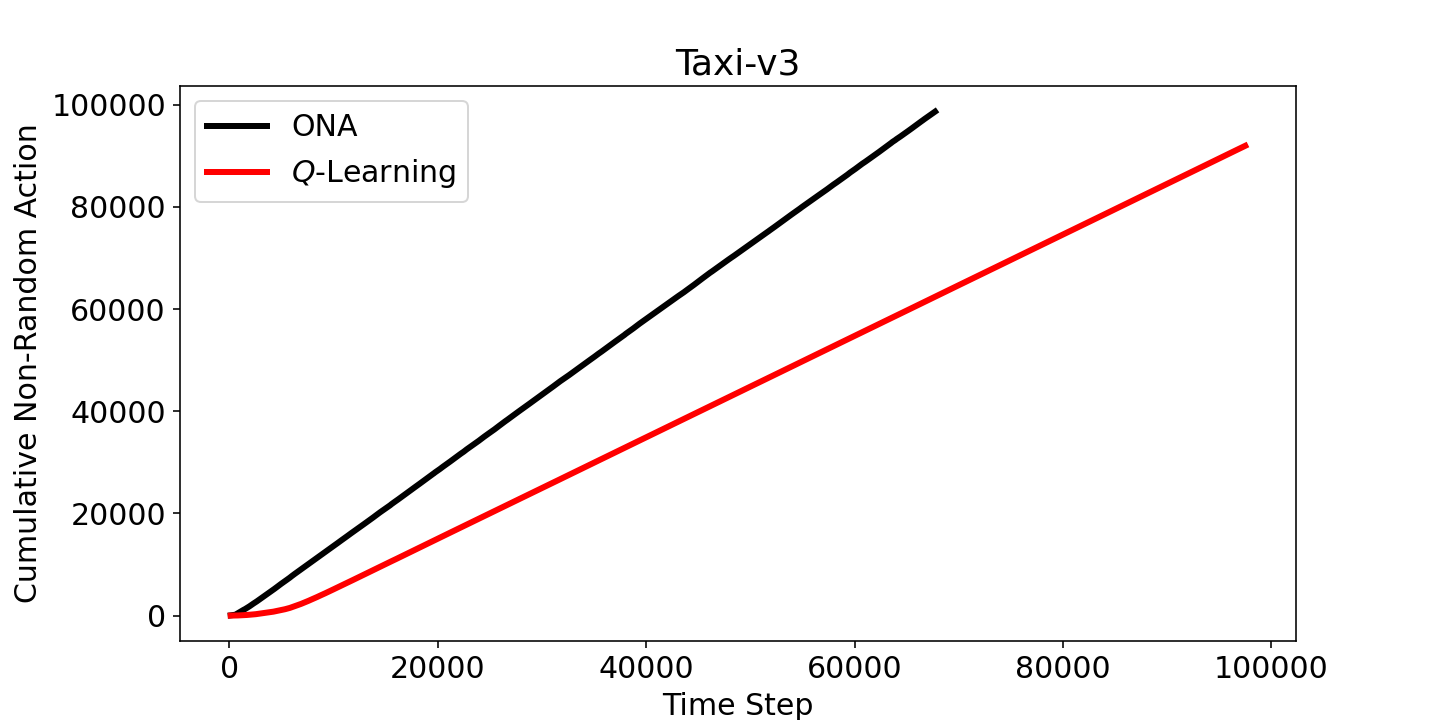}
         \caption{Taxi-v3}
         \label{Cumulative_Non-Random_Action_vs_Time_Step_Taxi-v3}
     \end{subfigure}
     \begin{subfigure}{0.32\columnwidth}
         \centering
         \includegraphics[width=\columnwidth]{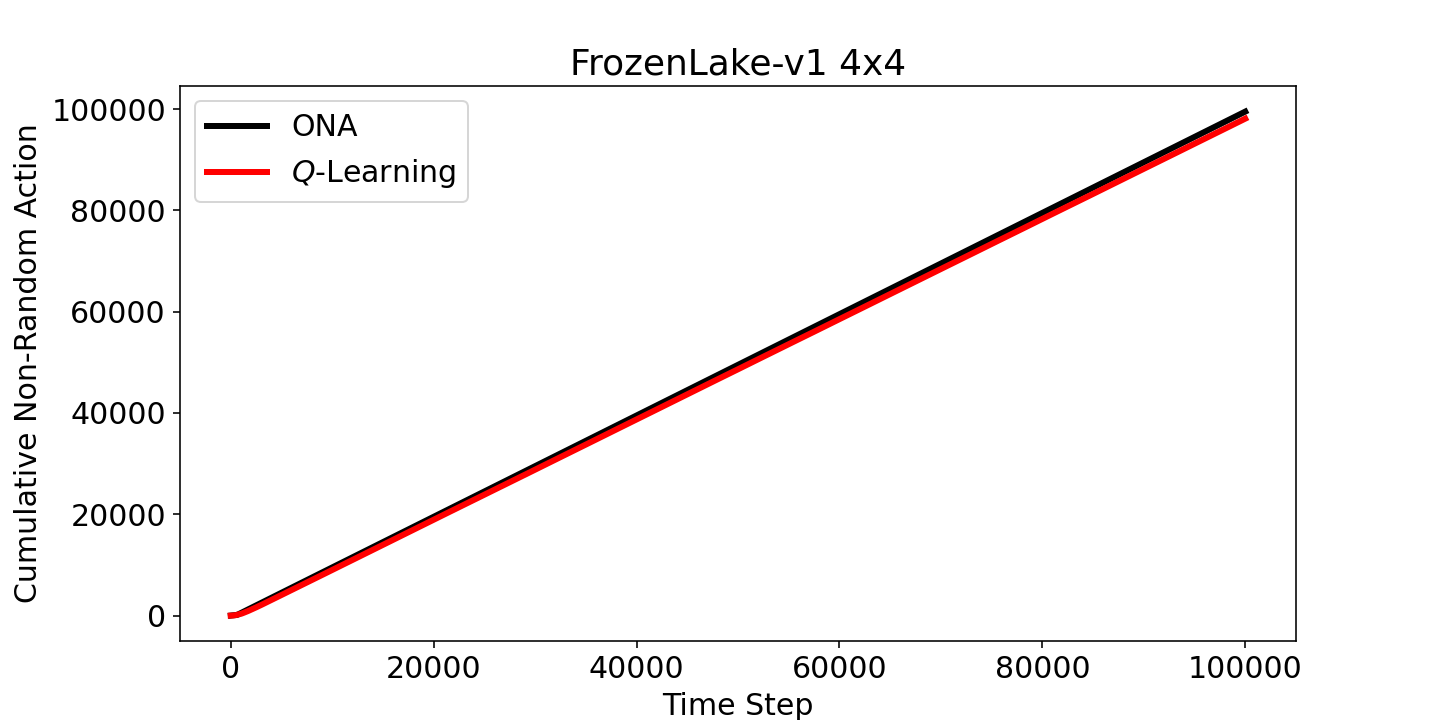}
         \caption{FrozenLake-v1 4x4}
         \label{Cumulative_Non-Random_Action_vs_Time_Step_FrozenLake-v1 4x4}
     \end{subfigure}
     \begin{subfigure}{0.32\columnwidth}
         \centering
         \includegraphics[width=\columnwidth]{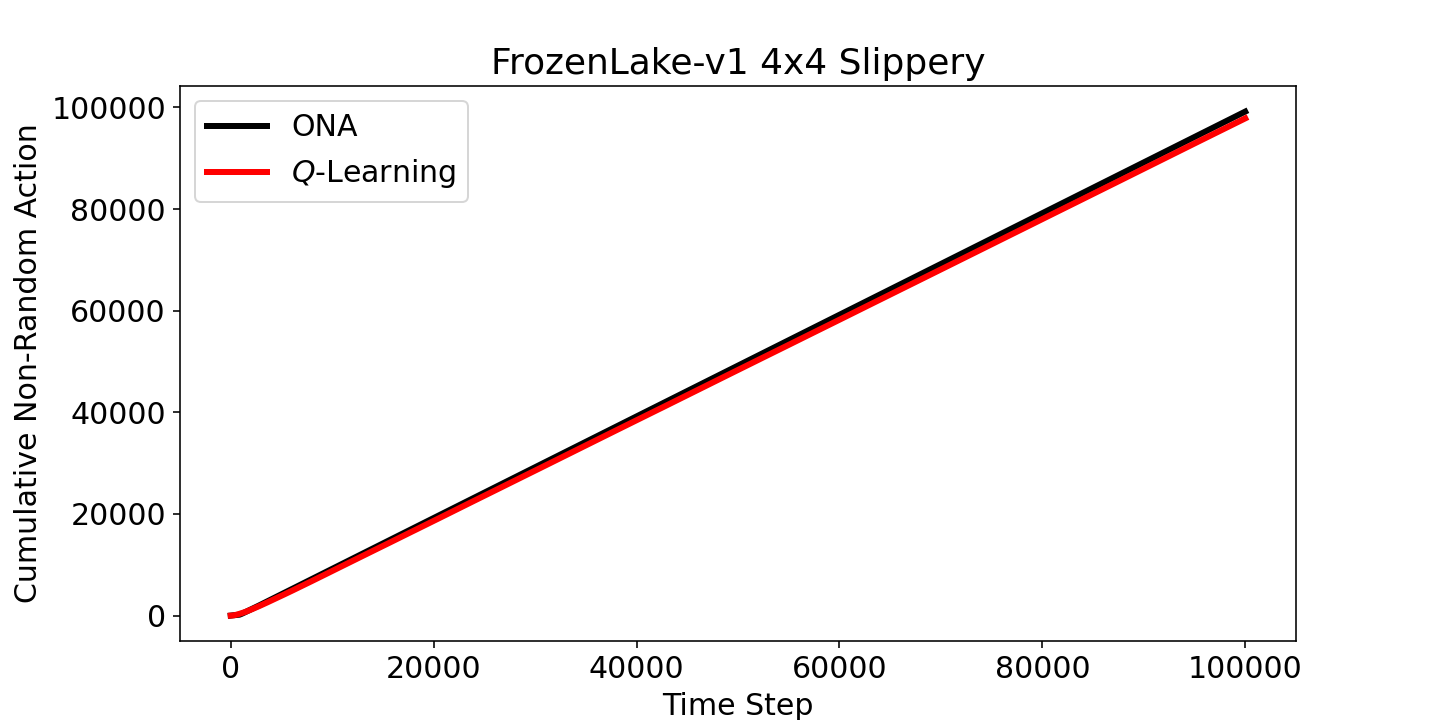}
         \caption{FrozenLake-v1 4x4 Slippery}
         \label{Cumulative_Non-Random_Action_vs_Time_Step_FrozenLake-v1_4x4_Slippery}
     \end{subfigure}
    \begin{subfigure}{0.32\columnwidth}
         \centering
         \includegraphics[width=\columnwidth]{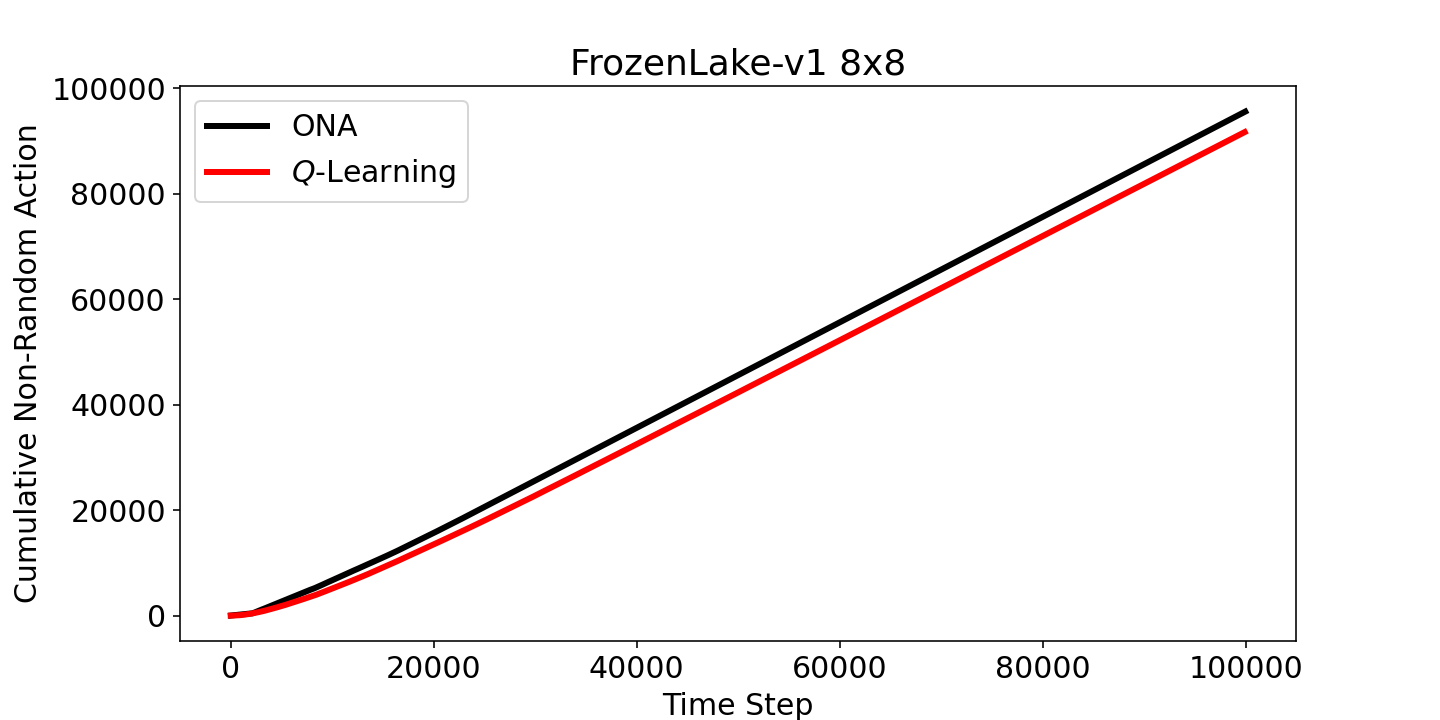}
         \caption{FrozenLake-v1 8x8}
         \label{Cumulative_Non-Random_Action_vs_Time_Step_FrozenLake-v1 8x8}
     \end{subfigure}
    \begin{subfigure}{0.32\columnwidth}
         \centering
         \includegraphics[width=\columnwidth]{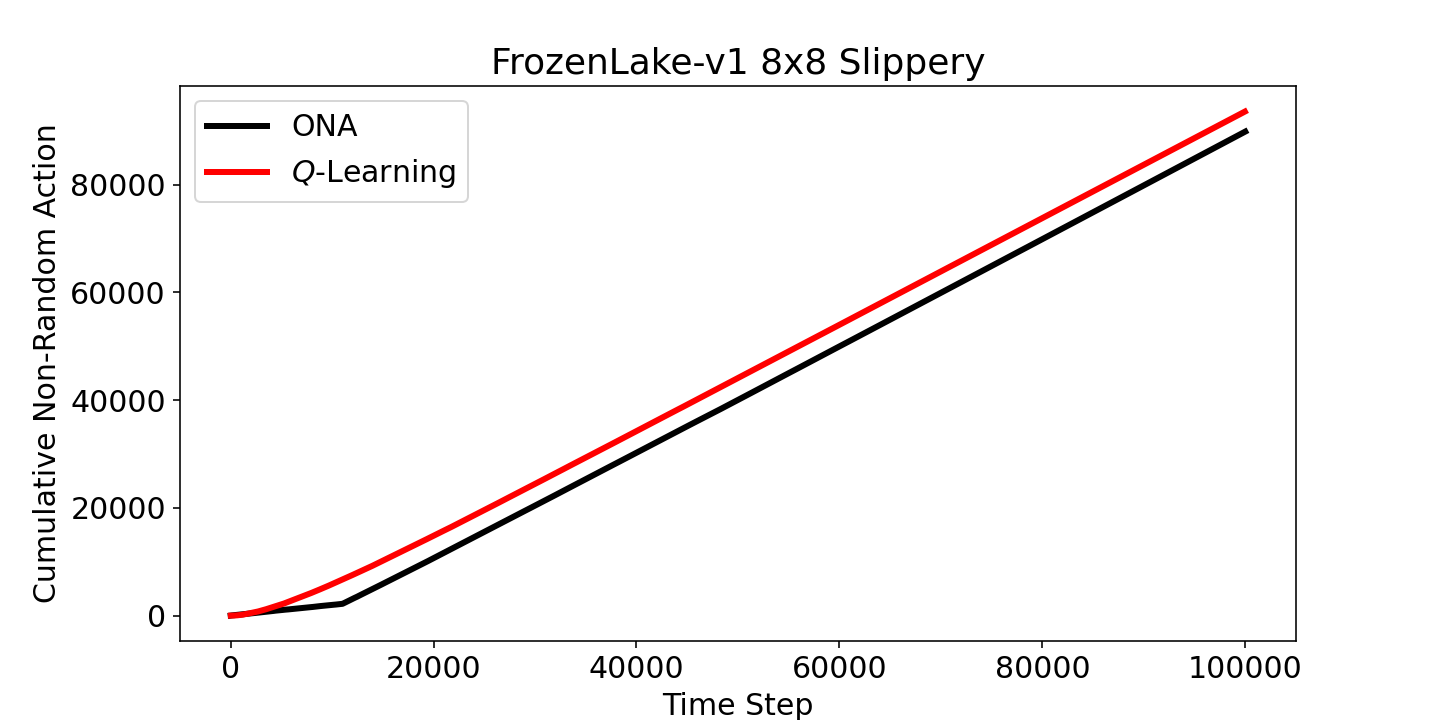}
         \caption{FrozenLake-v1 8x8 Slippery}
         \label{Cumulative_Non-Random_Action_vs_Time_Step_FrozenLake-v1_8x8_Slippery}
     \end{subfigure}
     \begin{subfigure}{0.32\columnwidth}
         \centering
         \includegraphics[width=\columnwidth]{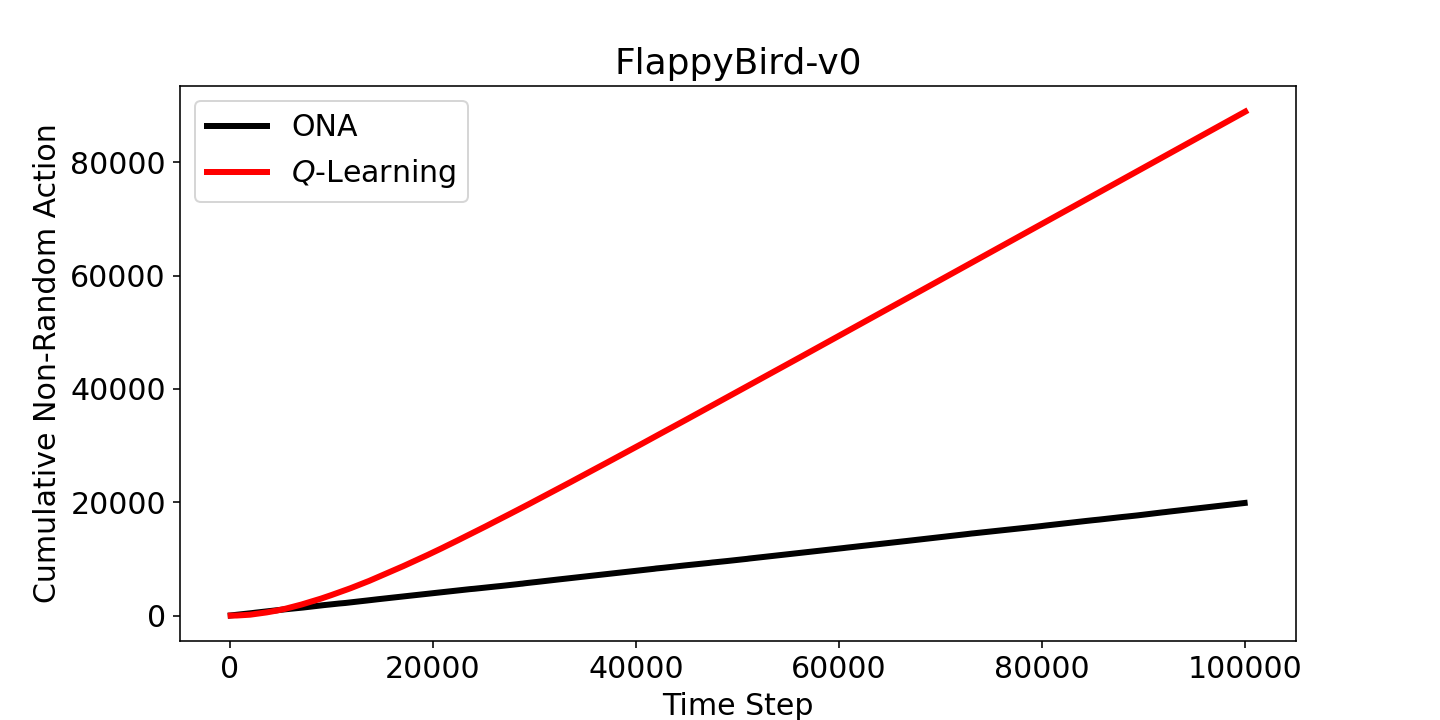}
         \caption{FlappyBird-v0}
         \label{Cumulative_Non-Random_Action_vs_Time_Step_FlappyBird-v0}
     \end{subfigure}
        \caption{Cumulative Non-Random Action vs. Time steps.}
        \label{Cumulative_Non-Random_Action_vs_Time Step}
\end{figure}

Both techniques are run one time due to time limit in each experiment, and the behavior of many parameters, which are shown in Figures \ref{Reward_vs_Time Step}, \ref{Cumulative_Successful_Episodes_vs_Time Step}, \ref{Epsilon_vs_Episodes}, \ref{Cumulative_Random_Action_vs_Time Step}, and \ref{Cumulative_Non-Random_Action_vs_Time Step}, kept track of for each time step across 100000 iterations. Each technique is run one time with random seed = 1. Details can be found in the source code in the following link: \url{https://github.com/AliBeikmohammadi/OpenNARS-for-Applications/tree/master/misc/Python}.

As can be seen from Figures \ref{Reward_vs_Time Step} and \ref{Cumulative_Successful_Episodes_vs_Time Step}, the results of two algorithms are very dependent on the task and one cannot be considered superior for all environments. Specifically, the $Q$-Learning algorithm has performed better on CliffWalking-v0, Taxi-v3, and FlappyBird-v0 environments. But ONA is more promising on environments based on FrozenLake-v1. 

A very interesting point is the ability of ONA to solve non-deterministic problems, where it is able to solve the slippery-enable problems as shown in Figures \ref{Reward_vs_Time_Step_FrozenLake-v1_4x4_Slippery},  \ref{Reward_vs_Time_Step_FrozenLake-v1_8x8_Slippery}, \ref{Cumulative_Successful_Episodes_vs_Time_Step_FrozenLake-v1_4x4_Slippery}, and \ref{Cumulative_Successful_Episodes_vs_Time_Step_FrozenLake-v1_8x8_Slippery}, while $Q$-Learning has not been successful. However, by changing the hyperparameters of $Q$-Learning, it might be possible to finally draw conclusions from it. However, note that, all time dependencies of hyperparameters are implicitly example-specific, and have hence to be avoided when generality is evaluated. With the passing of time, a reduction of the learning rate makes the $Q$-Learner take longer to change its policy when new circumstances demand it. 

On the other side, in general, it can be seen that ONA guarantees more reliability thanks to having fewer hyperparameters. To be more specific, ONA does not need a specific reduction of learning rate and exploration rate to work well for a particular example, hence needs less parameter tuning. As an example, ONA does not rely on learning rate decay. In fact, how much new evidences changes an existing belief is only dependent on the amount of evidence which already supports it, making high-confident beliefs automatically more stable. So, that is why the learning behavior of ONA is more consistent.

In Figure \ref{Epsilon_vs_Episodes}, we have drawn how to reduce $\epsilon$, which directly determines the probability of choosing a random action for $Q$-Learning.
Then, in Figures \ref{Cumulative_Random_Action_vs_Time Step} and \ref{Cumulative_Non-Random_Action_vs_Time Step}, the behavior of the two algorithms is drawn in terms of referring to a random or non-random action. Regarding $Q$-Learning, since the selection of the random action is controlled only by $\epsilon$, it can be clearly seen that after a short period of time, the probability of performing a random action is greatly reduced. This means that if a good policy has not yet been found, or if the environment changes, the agent will not be able to solve the task. So, reduction of $\epsilon$ over time will make it increasingly unlikely to attempt an alternative solutions. 

But on the other hand, in ONA, we know that apart from the fact that sometimes the system does not suggest any action to us, which subsequently we choose a random action in this situation, which is shown in Figure \ref{Cumulative_Random_Action_vs_Time Step}, thanks to the \textit{motorbabbling} parameter, it is possible that the system itself suggests a random action to explore the environment. ONA reduces \textit{motorbabbling} by itself once the hypotheses it bases its decisions on are stable and predict successfully, and hence does not depend on a time-dependent reduction of the exploration rate either.
Specifically, however, in this case, some of the actions drawn in Figure \ref{Cumulative_Non-Random_Action_vs_Time Step}, despite being named non-randomly (since they were chosen by the system and not by us), are actually random and exploratory in nature. This could be a reason for ONA's success, despite not receiving frequent reward like $Q$-Learning. So, ONA has the capability to deal with multiple and changing objectives, while it also demands less implicitly example-dependent parameter tuning than $Q$-Learning.

As future work, some of our examples can easily be extended to multi-objective scenarios and scenarios with changing objectives to show merits in these areas. Also, this technique can be used to improve the sample efficiency challenge of the new generation of RL algorithms which are established on using deep learning methods, which itself has achieved outstanding performance on complicated tasks like plant identification \cite{beikmohammadi2020swp, BEIKMOHAMMADI2022117470}, handwritten digit recognition \cite{beikmohammadi2021hierarchical}, and human action detection \cite{beikmohammadi2019mixture}, as the function approximators. In addition, one could also think  about any aggregation of both approaches to solve a problem. This aggregation could be done in different scenarios, including voting, hierarchically, teacher-student learning. Overall, both approaches were comparative in performance on average. Therefore the reasoning-based approach provides a likely alternative for such problems, while performing better whenever the problem is non-deterministic.

\appendix
\section{Additional Learning Curves} \label{section4}
Additional figures are included in this section.

\begin{figure}[H]%[t] %[h]
     \centering
     \begin{subfigure}{0.32\columnwidth}
         \centering
         \includegraphics[width=\columnwidth]{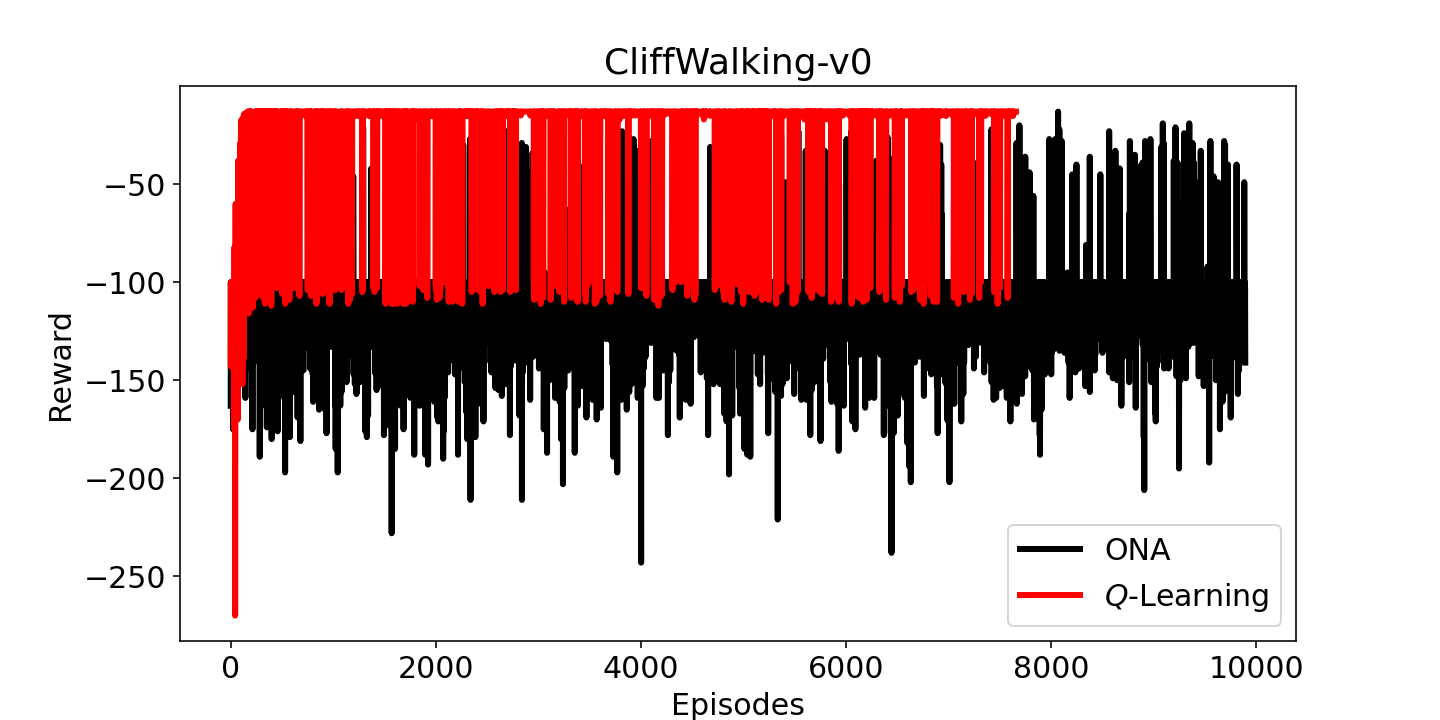}
         \caption{CliffWalking-v0}
         \label{Reward_vs_Episodes_CliffWalking-v0}
     \end{subfigure}
     \begin{subfigure}{0.32\columnwidth}
         \centering
         \includegraphics[width=\columnwidth]{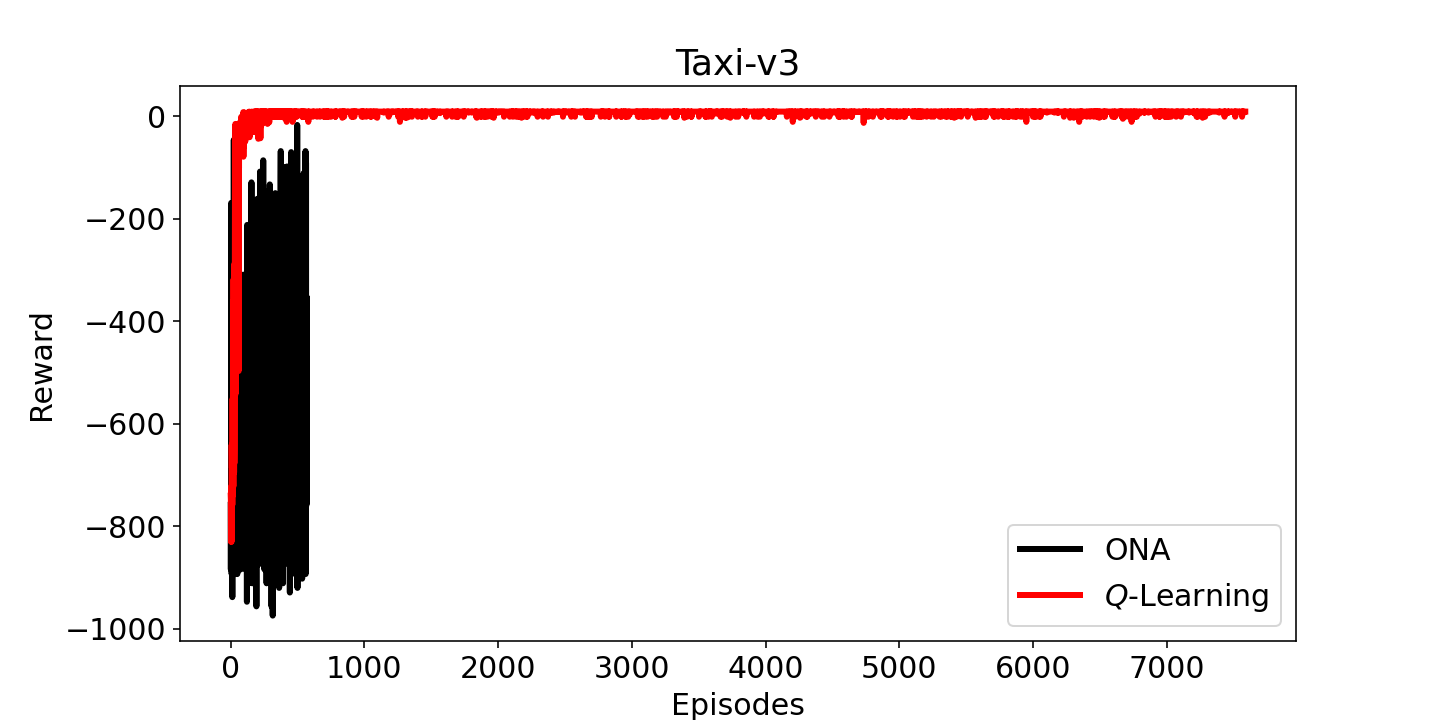}
         \caption{Taxi-v3}
         \label{Reward_vs_Episodes_Taxi-v3}
     \end{subfigure}
     \begin{subfigure}{0.32\columnwidth}
         \centering
         \includegraphics[width=\columnwidth]{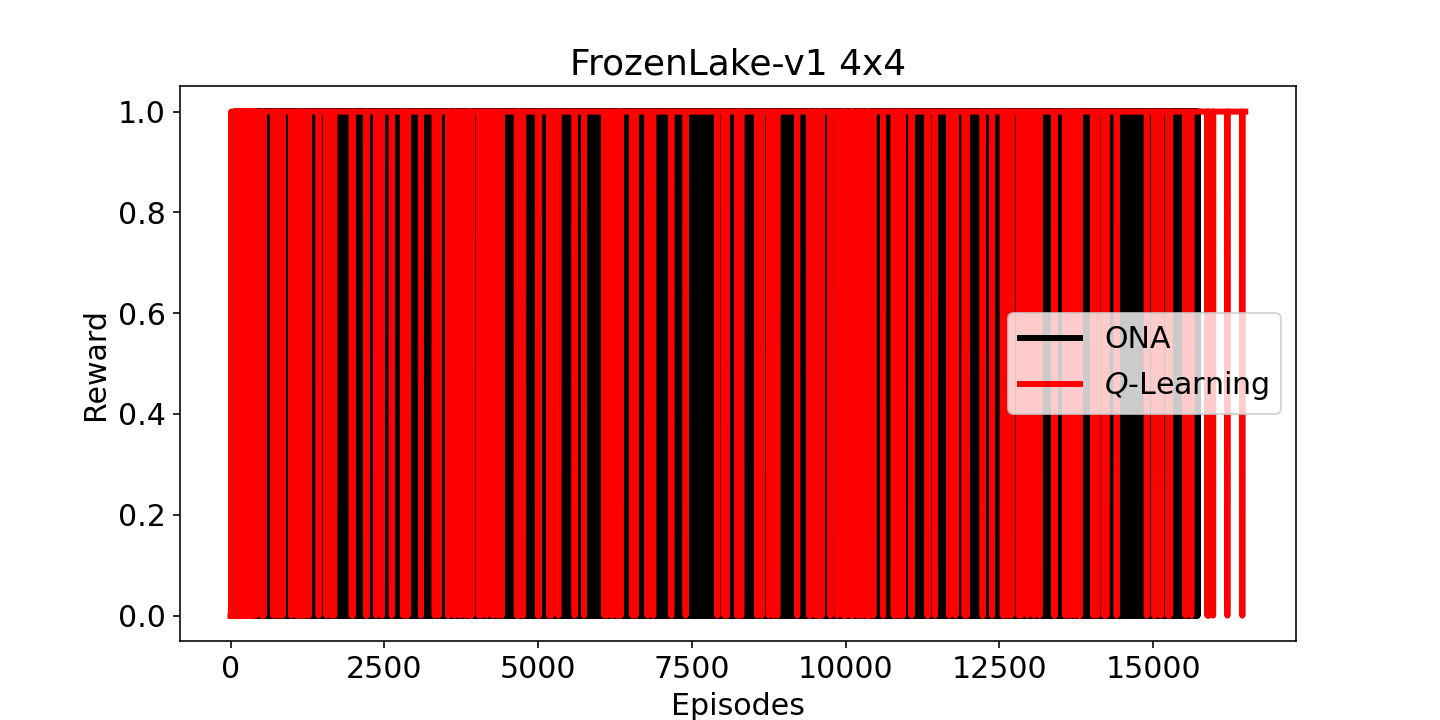}
         \caption{FrozenLake-v1 4x4}
         \label{Reward_vs_Episodes_FrozenLake-v1 4x4}
     \end{subfigure}
     \begin{subfigure}{0.32\columnwidth}
         \centering
         \includegraphics[width=\columnwidth]{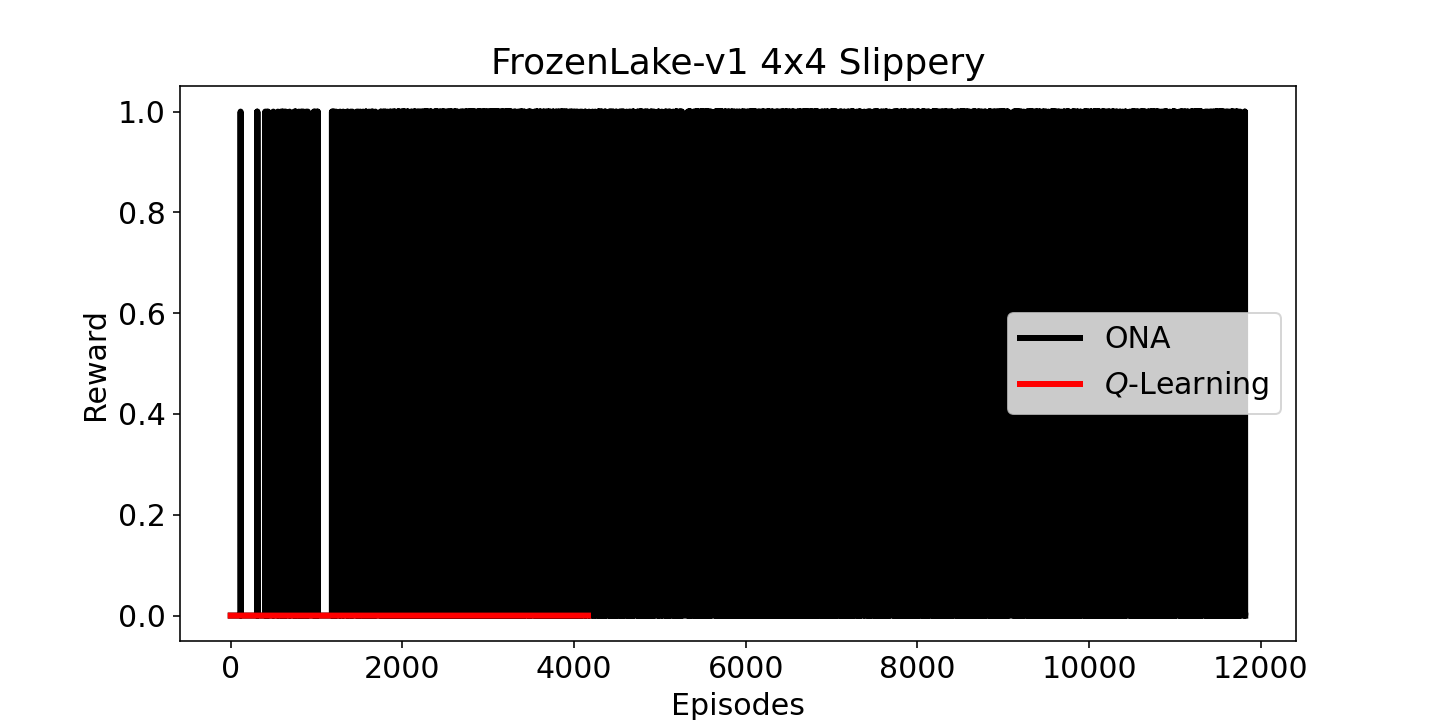}
         \caption{FrozenLake-v1 4x4 Slippery}
         \label{Reward_vs_Episodes_FrozenLake-v1_4x4_Slippery}
     \end{subfigure}
    \begin{subfigure}{0.32\columnwidth}
         \centering
         \includegraphics[width=\columnwidth]{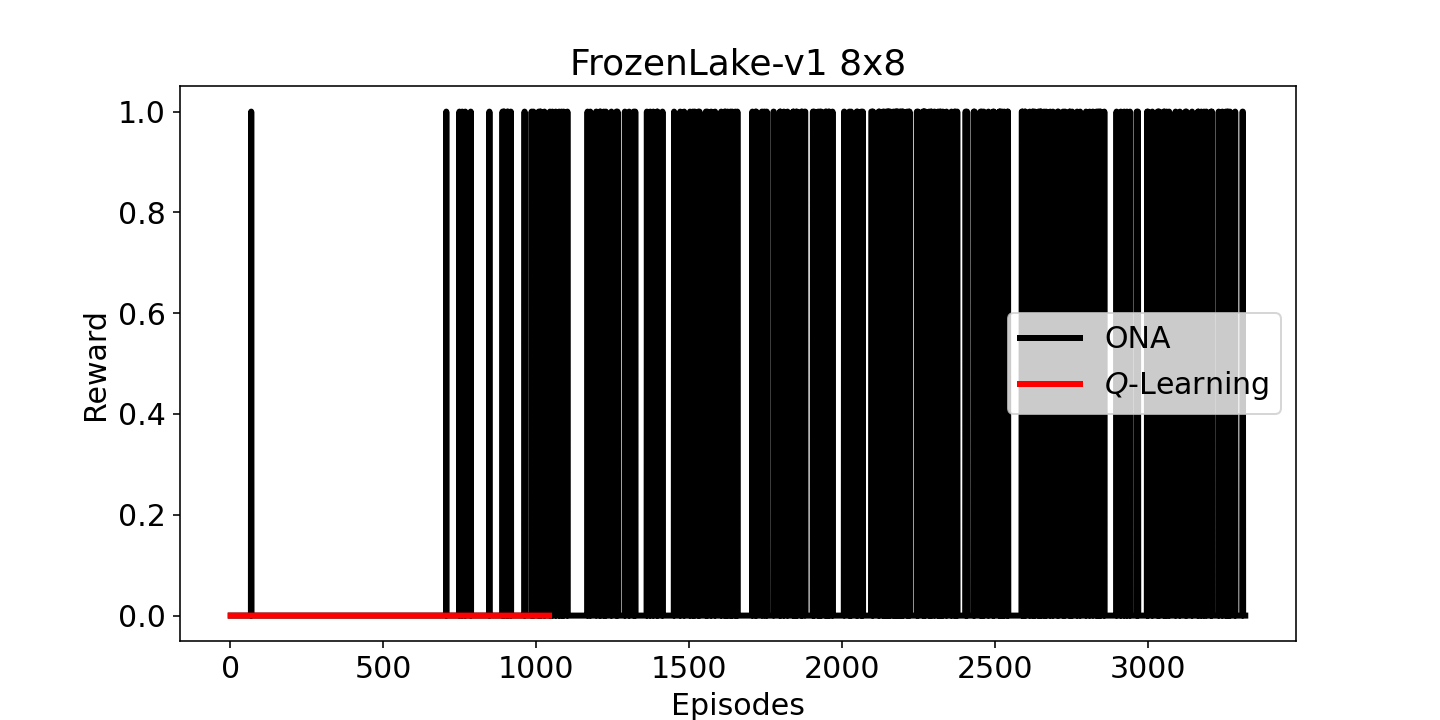}
         \caption{FrozenLake-v1 8x8}
         \label{Reward_vs_Episodes_FrozenLake-v1 8x8}
     \end{subfigure}
    \begin{subfigure}{0.32\columnwidth}
         \centering
         \includegraphics[width=\columnwidth]{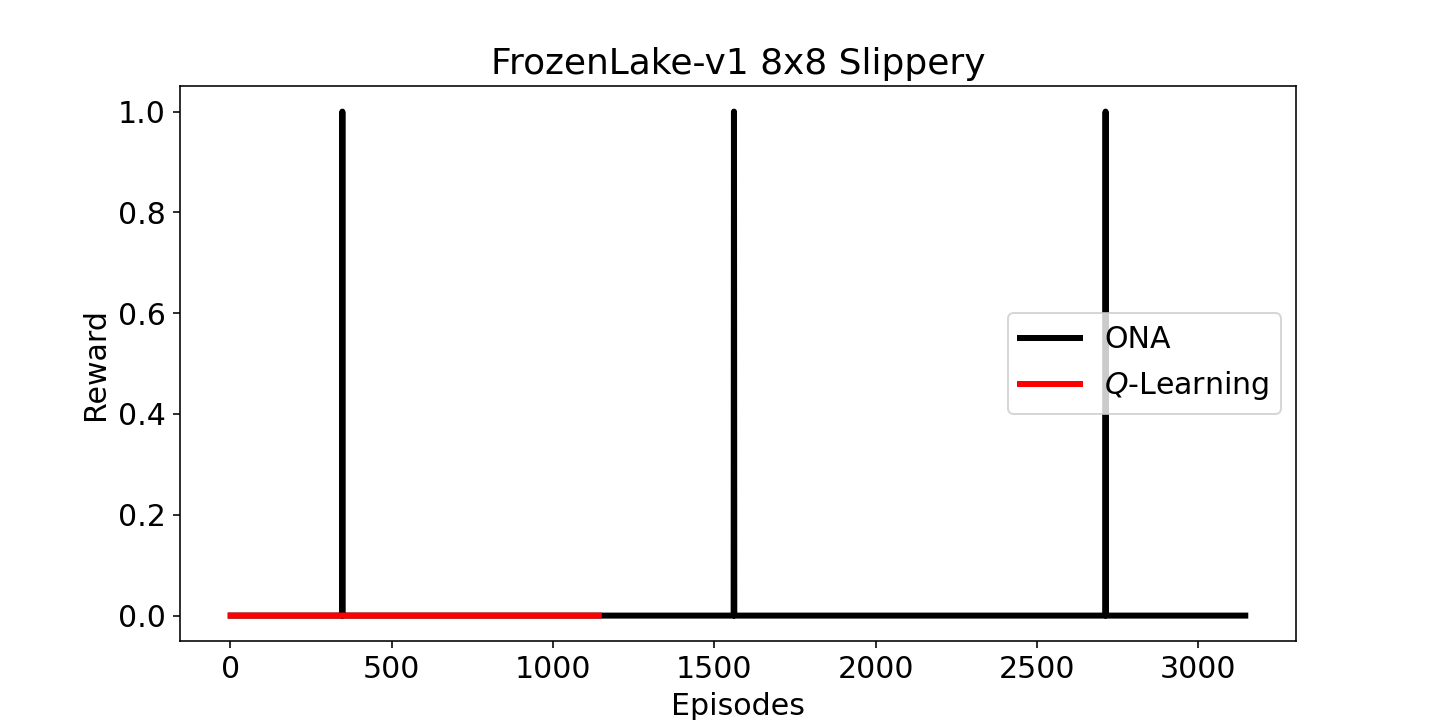}
         \caption{FrozenLake-v1 8x8 Slippery}
         \label{Reward_vs_Episodes_FrozenLake-v1_8x8_Slippery}
     \end{subfigure}
     \begin{subfigure}{0.32\columnwidth}
         \centering
         \includegraphics[width=\columnwidth]{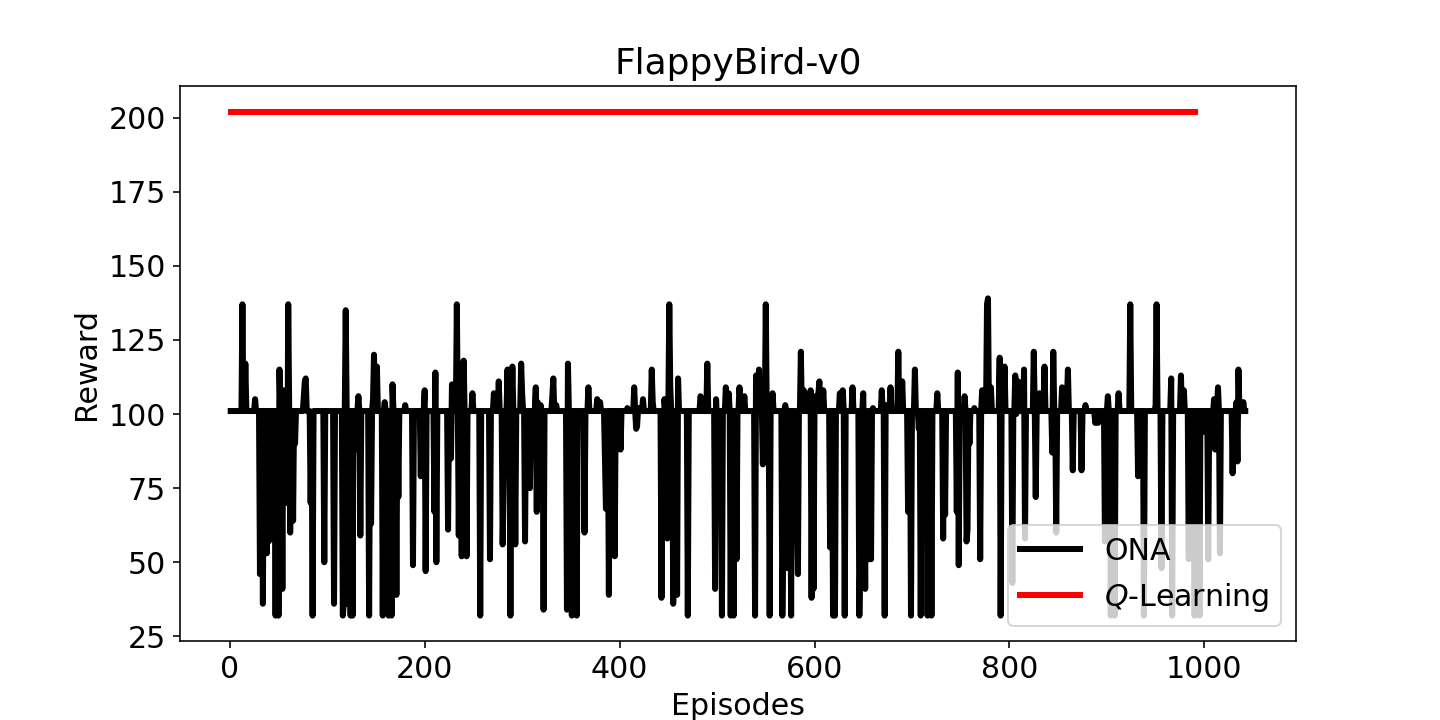}
         \caption{FlappyBird-v0}
         \label{Reward_vs_Episodes_FlappyBird-v0}
     \end{subfigure}
        \caption{Reward vs. Episodes. The reward is measured at time steps where the episode ends (by reaching the goal, truncating the episode length, falling into the hole, falling from the cliff, hitting the pipe.)}
        \label{Reward_vs_Episodes}
\end{figure}

\begin{figure}[H]%[t] %[h]
     \centering
     \begin{subfigure}{0.32\columnwidth}
         \centering
         \includegraphics[width=\columnwidth]{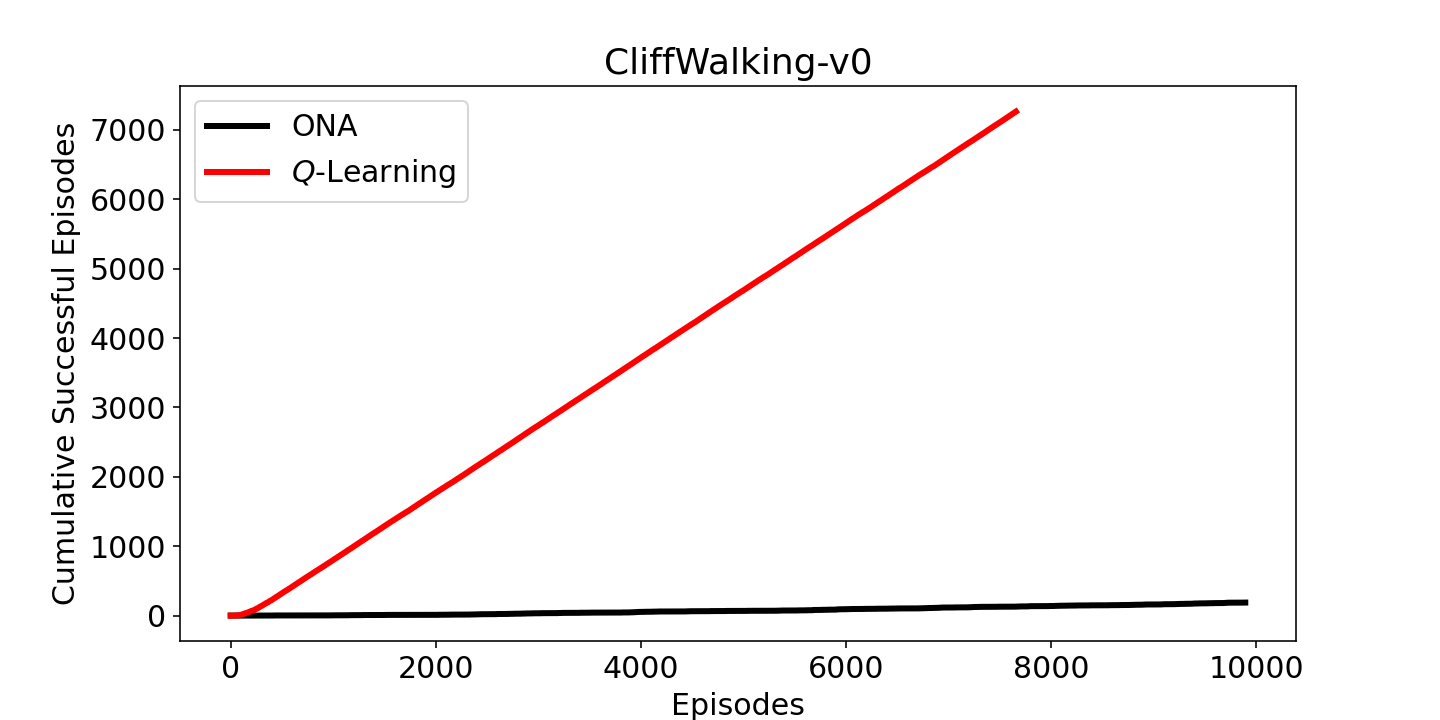}
         \caption{CliffWalking-v0}
         \label{Cumulative_Successful_Episodes_vs_Episodes_CliffWalking-v0}
     \end{subfigure}
     \begin{subfigure}{0.32\columnwidth}
         \centering
         \includegraphics[width=\columnwidth]{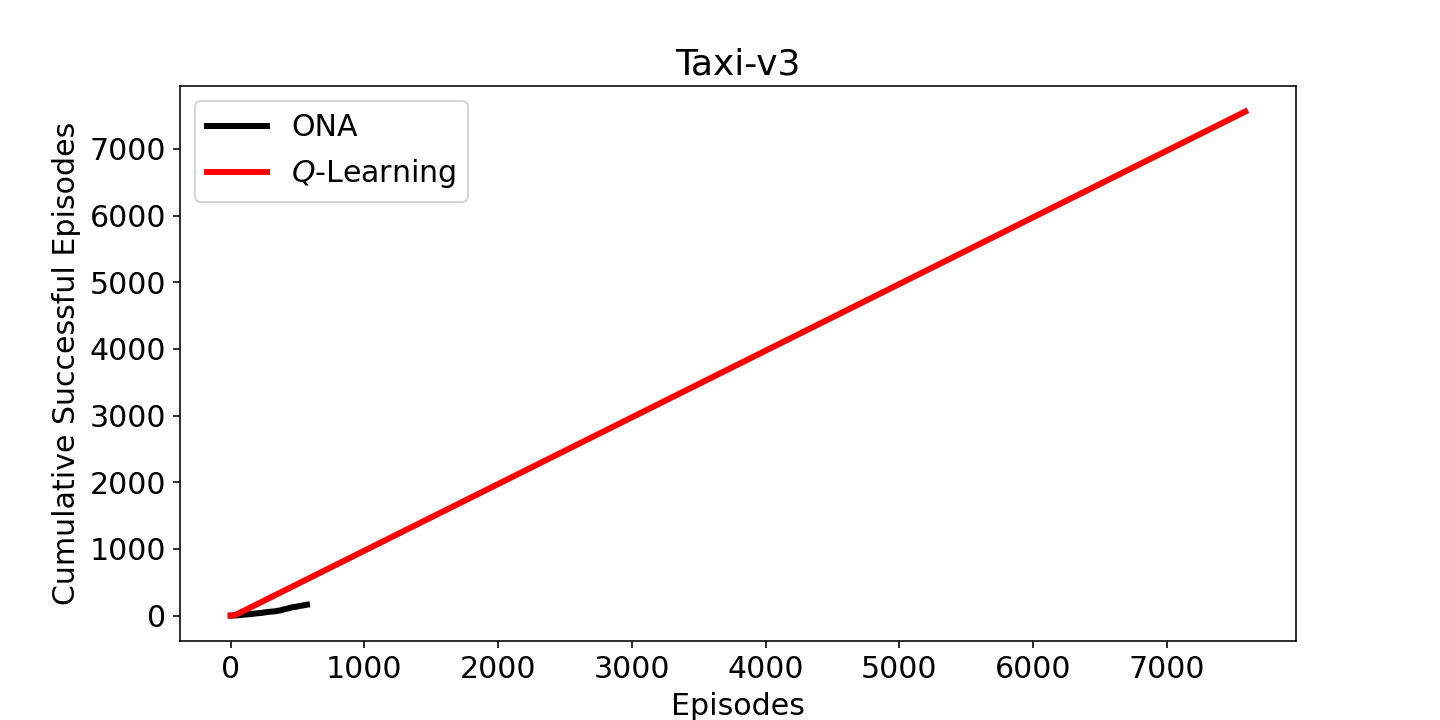}
         \caption{Taxi-v3}
         \label{Cumulative_Successful_Episodes_vs_Episodes_Taxi-v3}
     \end{subfigure}
     \begin{subfigure}{0.32\columnwidth}
         \centering
         \includegraphics[width=\columnwidth]{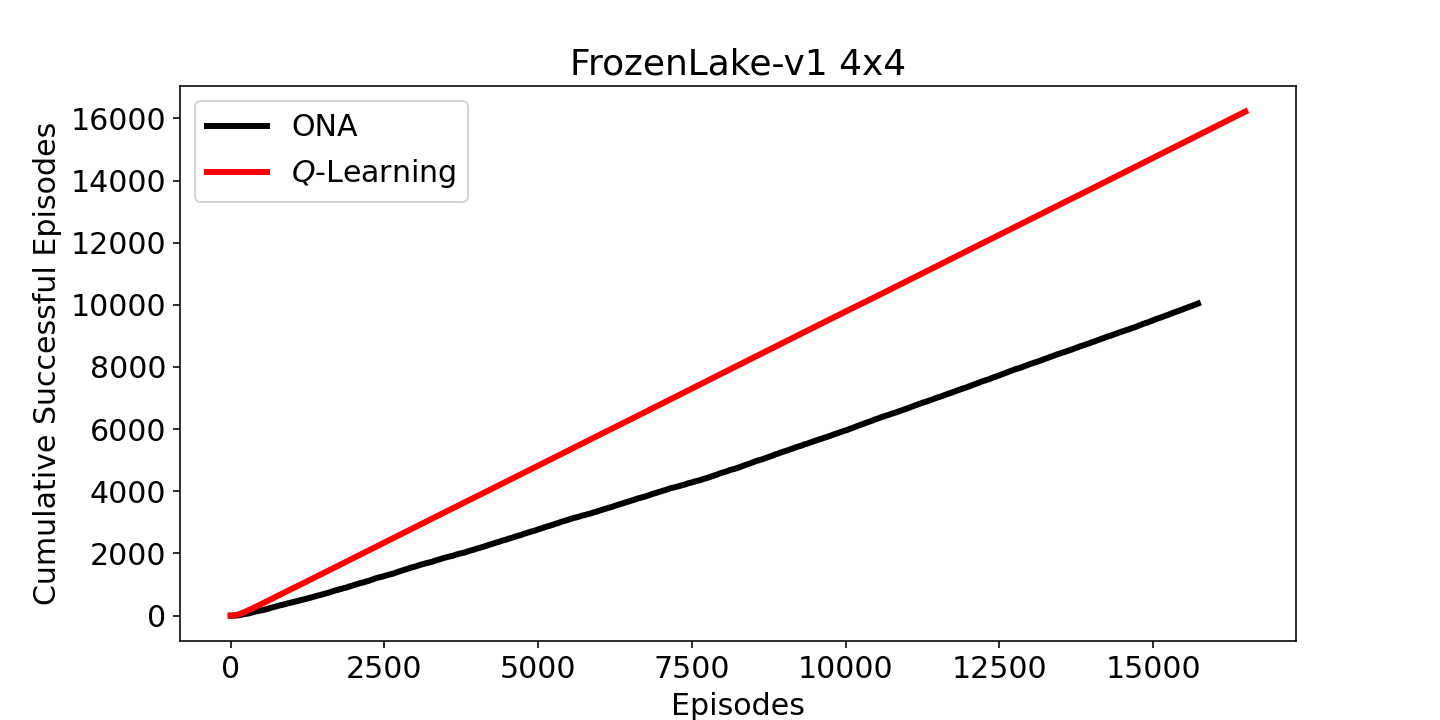}
         \caption{FrozenLake-v1 4x4}
         \label{Cumulative_Successful_Episodes_vs_Episodes_FrozenLake-v1 4x4}
     \end{subfigure}
     \begin{subfigure}{0.32\columnwidth}
         \centering
         \includegraphics[width=\columnwidth]{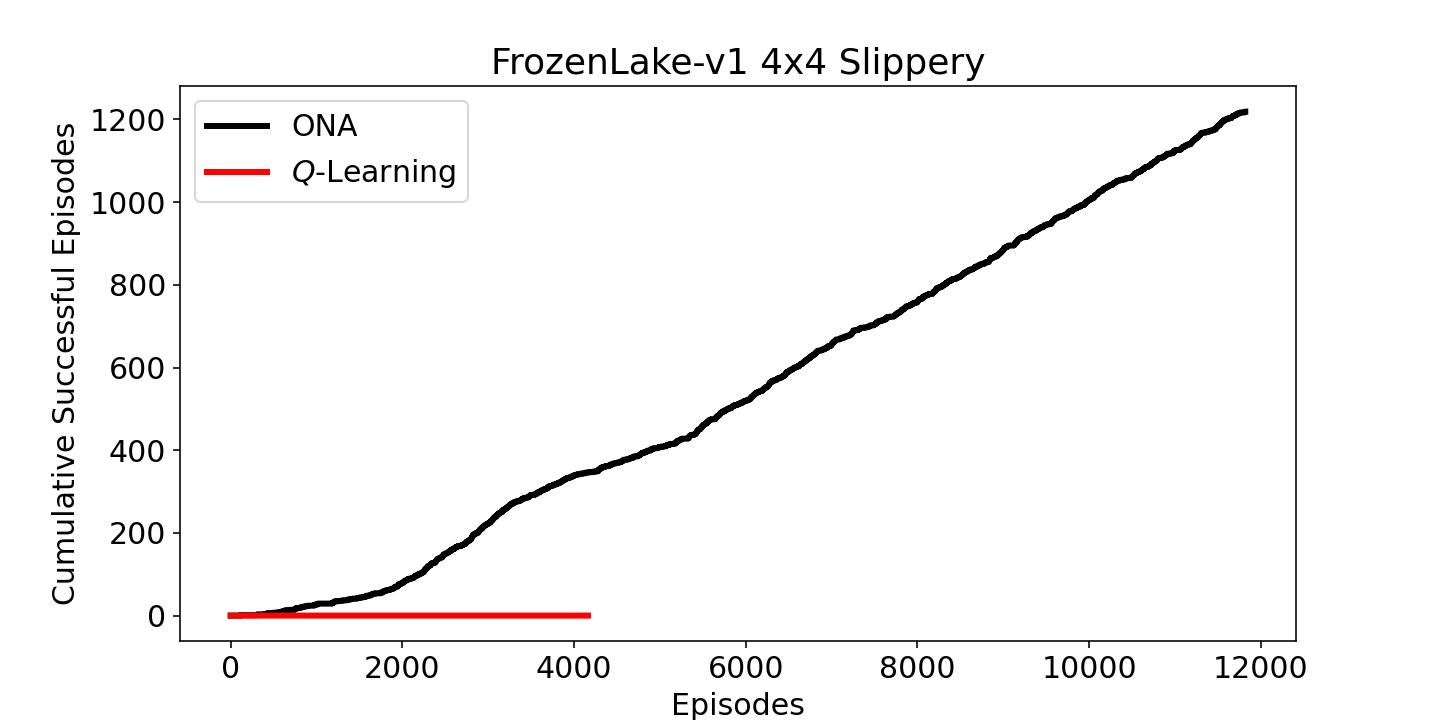}
         \caption{FrozenLake-v1 4x4 Slippery}
         \label{Cumulative_Successful_Episodes_vs_Episodes_FrozenLake-v1_4x4_Slippery}
     \end{subfigure}
    \begin{subfigure}{0.32\columnwidth}
         \centering
         \includegraphics[width=\columnwidth]{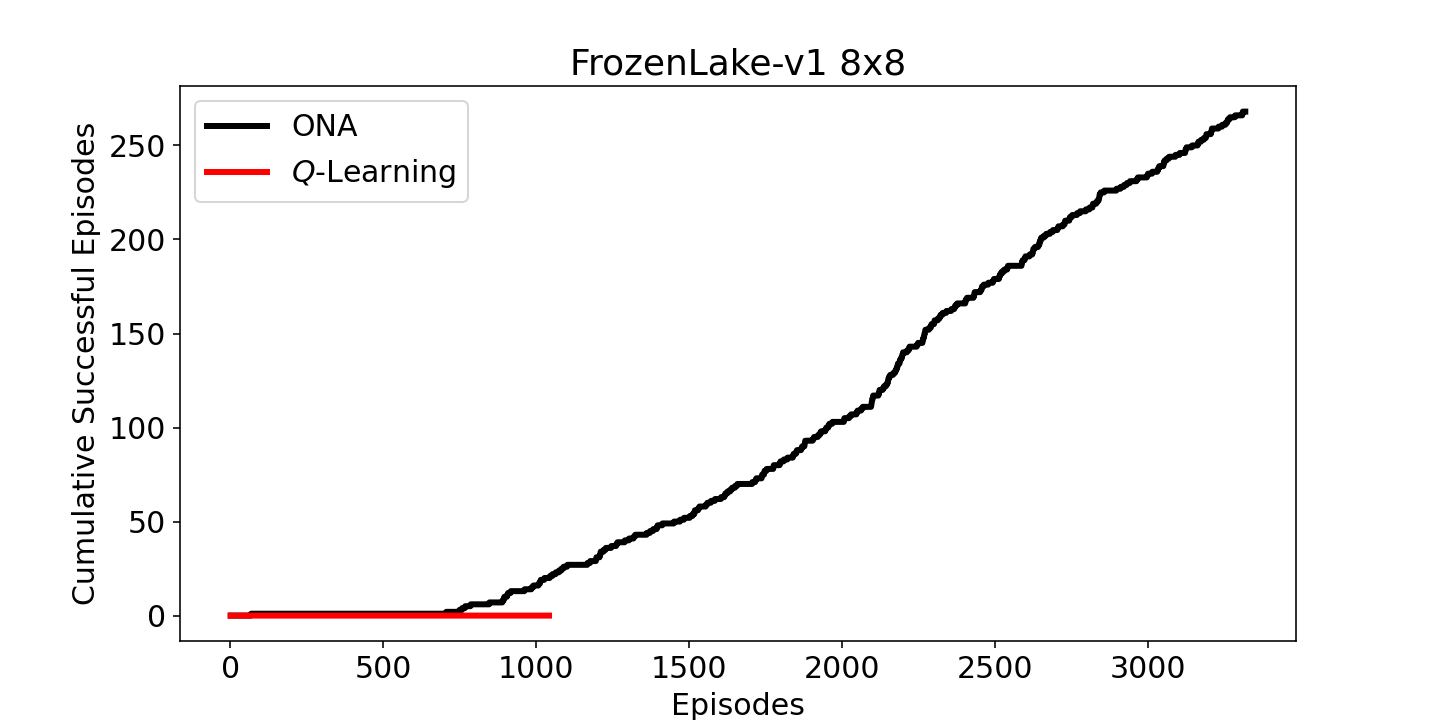}
         \caption{FrozenLake-v1 8x8}
         \label{Cumulative_Successful_Episodes_vs_Episodes_FrozenLake-v1 8x8}
     \end{subfigure}
    \begin{subfigure}{0.32\columnwidth}
         \centering
         \includegraphics[width=\columnwidth]{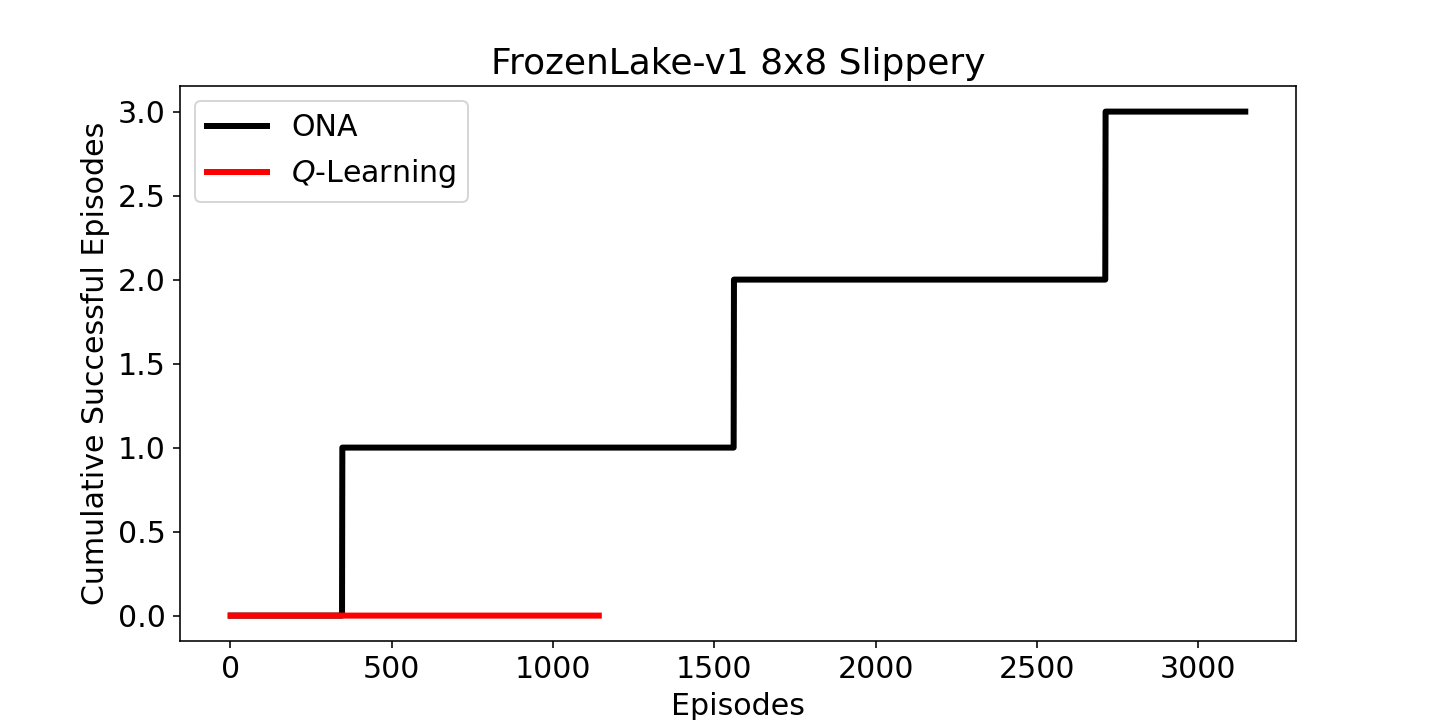}
         \caption{FrozenLake-v1 8x8 Slippery}
         \label{Cumulative_Successful_Episodes_vs_Episodes_FrozenLake-v1_8x8_Slippery}
     \end{subfigure}
        \caption{Cumulative Successful Episodes vs. Episodes.}
        \label{Cumulative_Successful_Episodes_vs_Episodes}
\end{figure}

\begin{figure}[H]%[t] %[h]
     \centering
     \begin{subfigure}{0.32\columnwidth}
         \centering
         \includegraphics[width=\columnwidth]{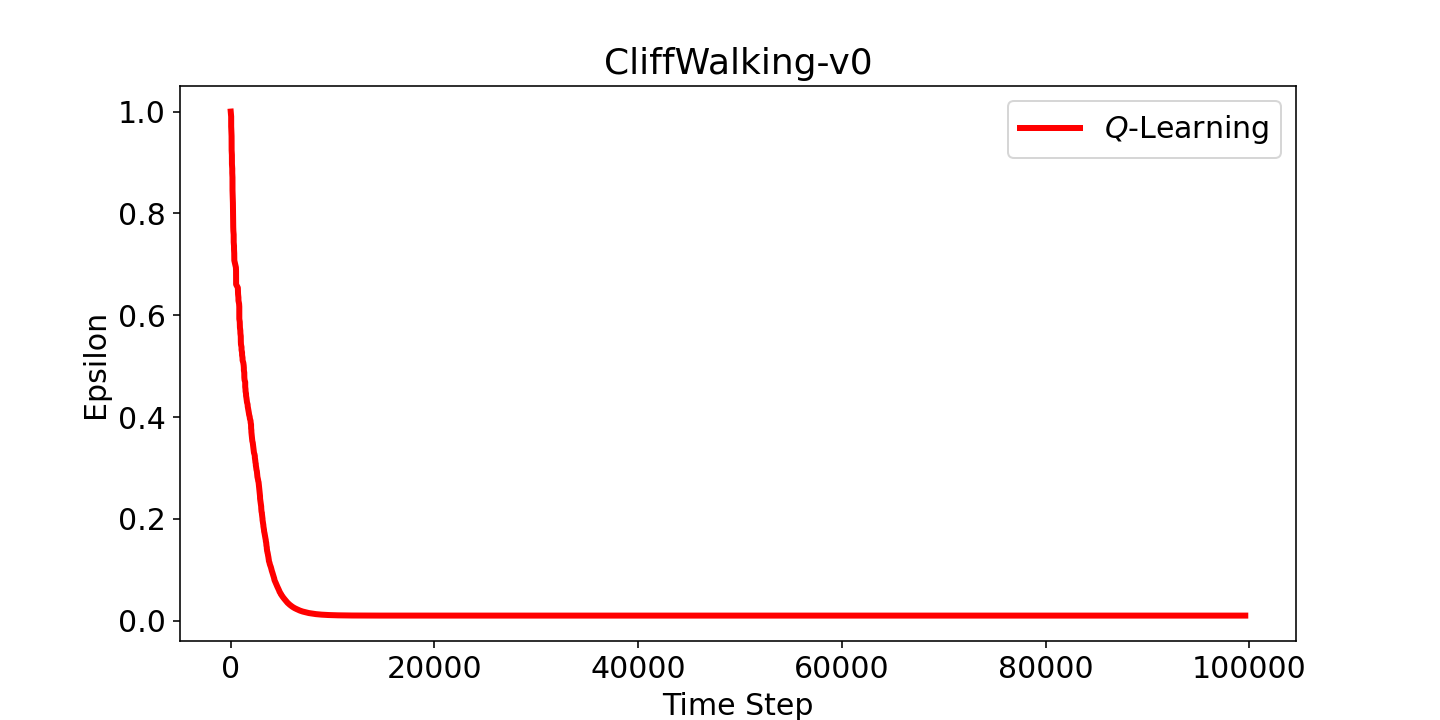}
         \caption{CliffWalking-v0}
         \label{Epsilon_vs_Time_Step_CliffWalking-v0}
     \end{subfigure}
     \begin{subfigure}{0.32\columnwidth}
         \centering
         \includegraphics[width=\columnwidth]{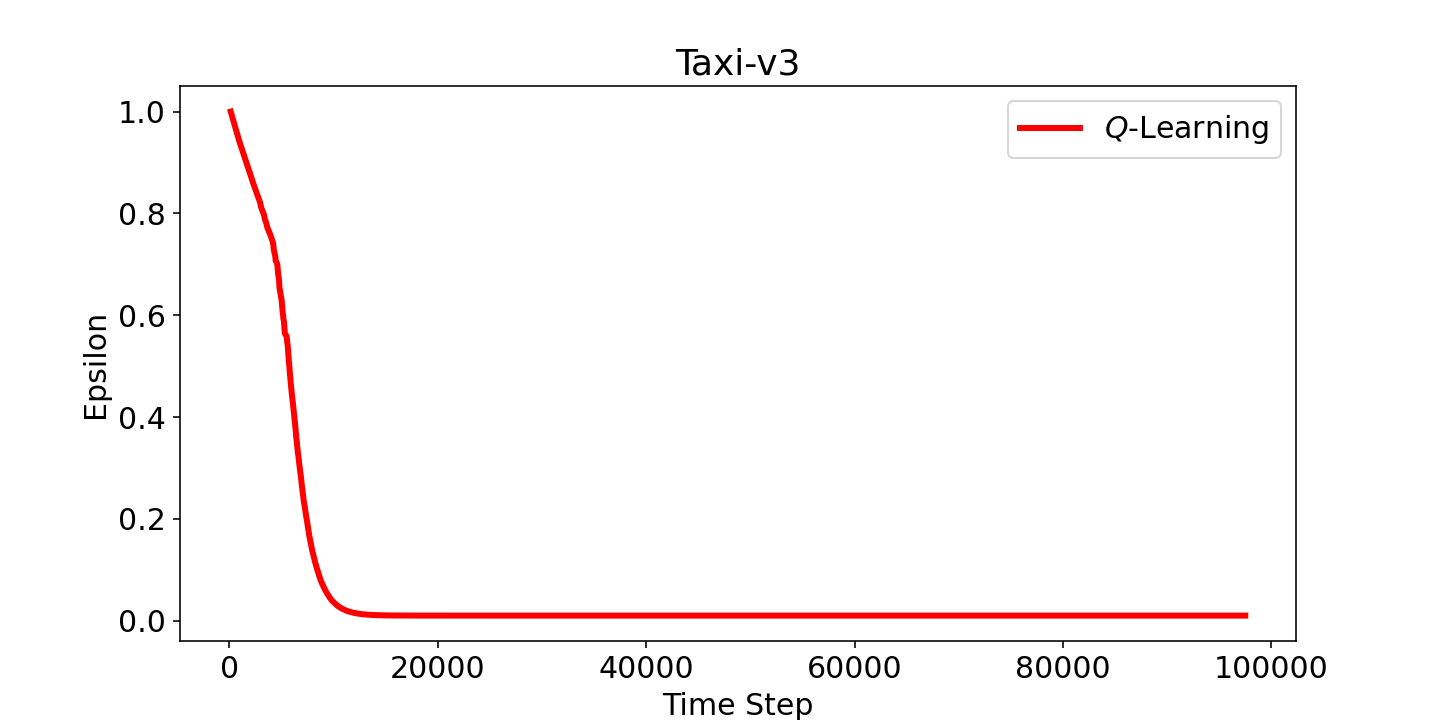}
         \caption{Taxi-v3}
         \label{Epsilon_vs_Time_Step_Taxi-v3}
     \end{subfigure}
     \begin{subfigure}{0.32\columnwidth}
         \centering
         \includegraphics[width=\columnwidth]{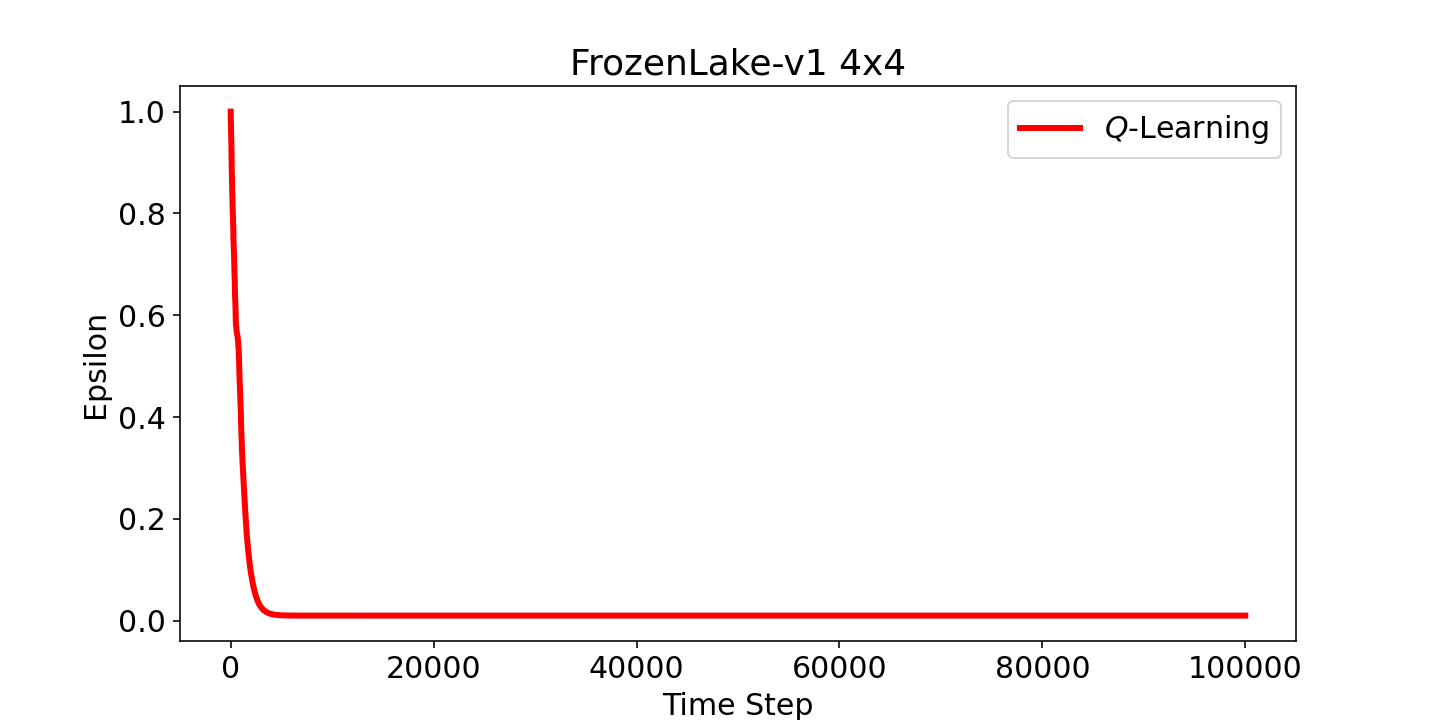}
         \caption{FrozenLake-v1 4x4}
         \label{Epsilon_vs_Time_Step_FrozenLake-v1 4x4}
     \end{subfigure}
     \begin{subfigure}{0.32\columnwidth}
         \centering
         \includegraphics[width=\columnwidth]{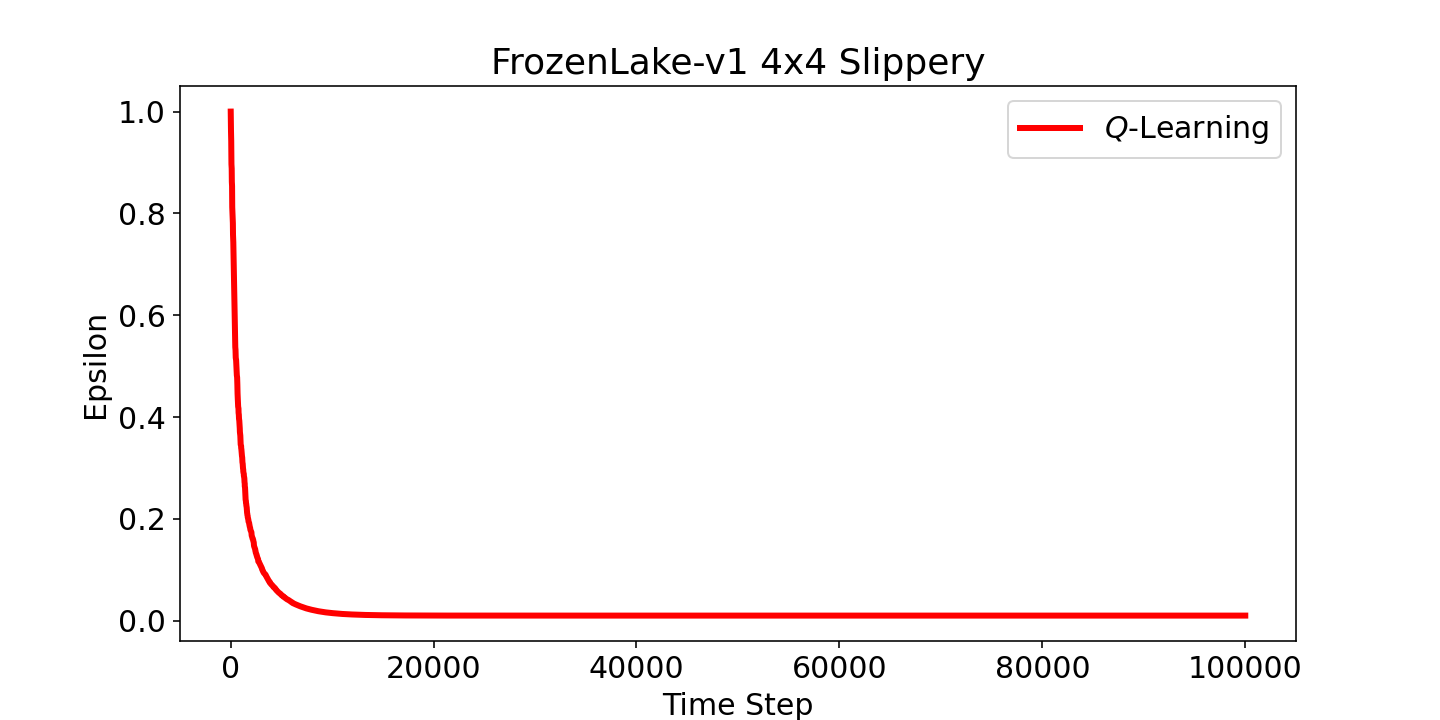}
         \caption{FrozenLake-v1 4x4 Slippery}
         \label{Epsilon_vs_Time_Step_FrozenLake-v1_4x4_Slippery}
     \end{subfigure}
    \begin{subfigure}{0.32\columnwidth}
         \centering
         \includegraphics[width=\columnwidth]{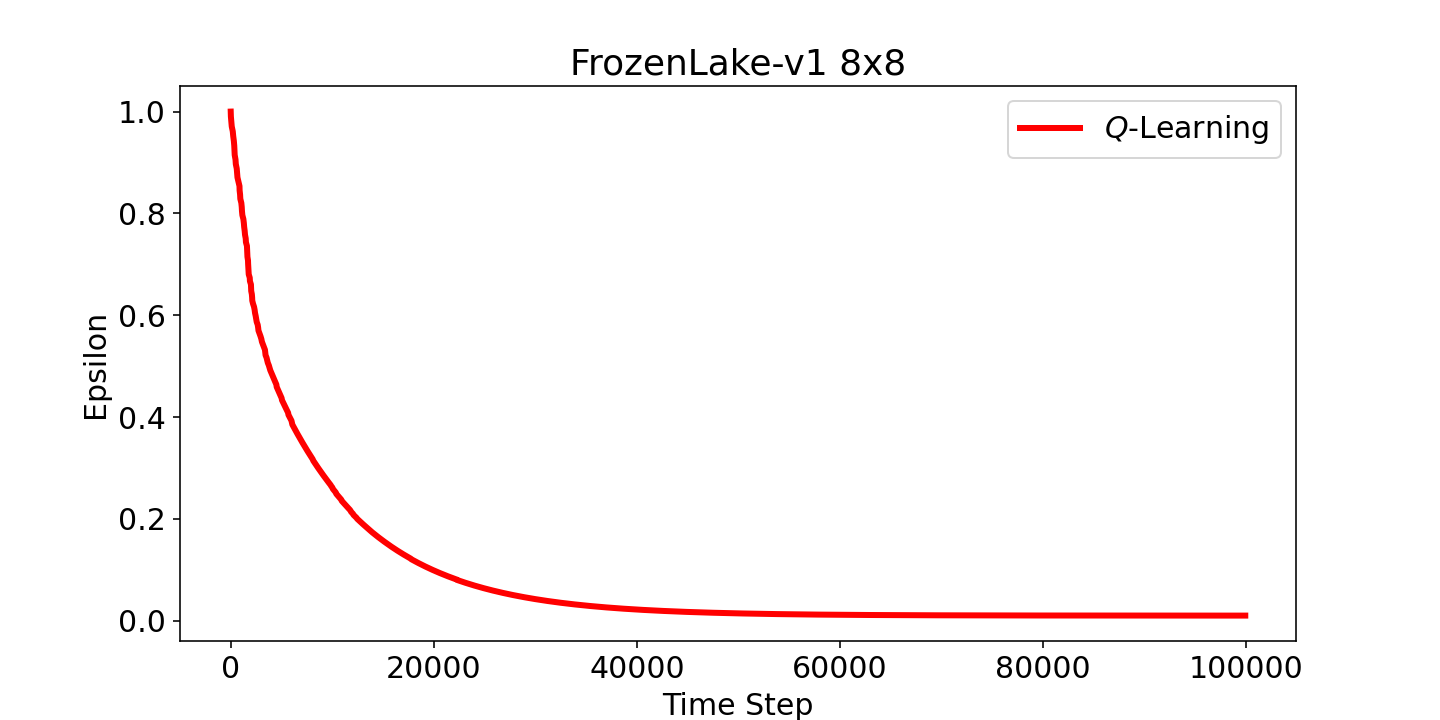}
         \caption{FrozenLake-v1 8x8}
         \label{Epsilon_vs_Time_Step_FrozenLake-v1 8x8}
     \end{subfigure}
    \begin{subfigure}{0.32\columnwidth}
         \centering
         \includegraphics[width=\columnwidth]{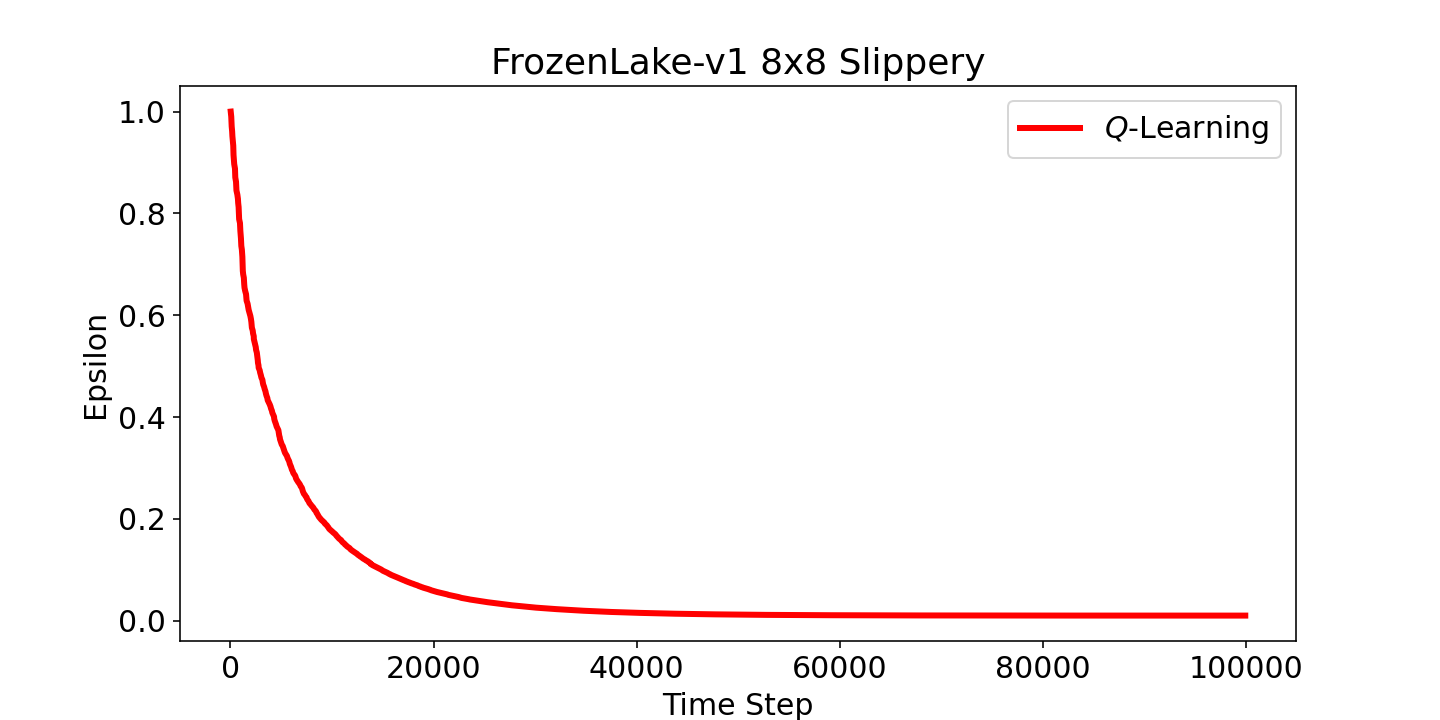}
         \caption{FrozenLake-v1 8x8 Slippery}
         \label{Epsilon_vs_Time_Step_FrozenLake-v1_8x8_Slippery}
     \end{subfigure}
     \begin{subfigure}{0.32\columnwidth}
         \centering
         \includegraphics[width=\columnwidth]{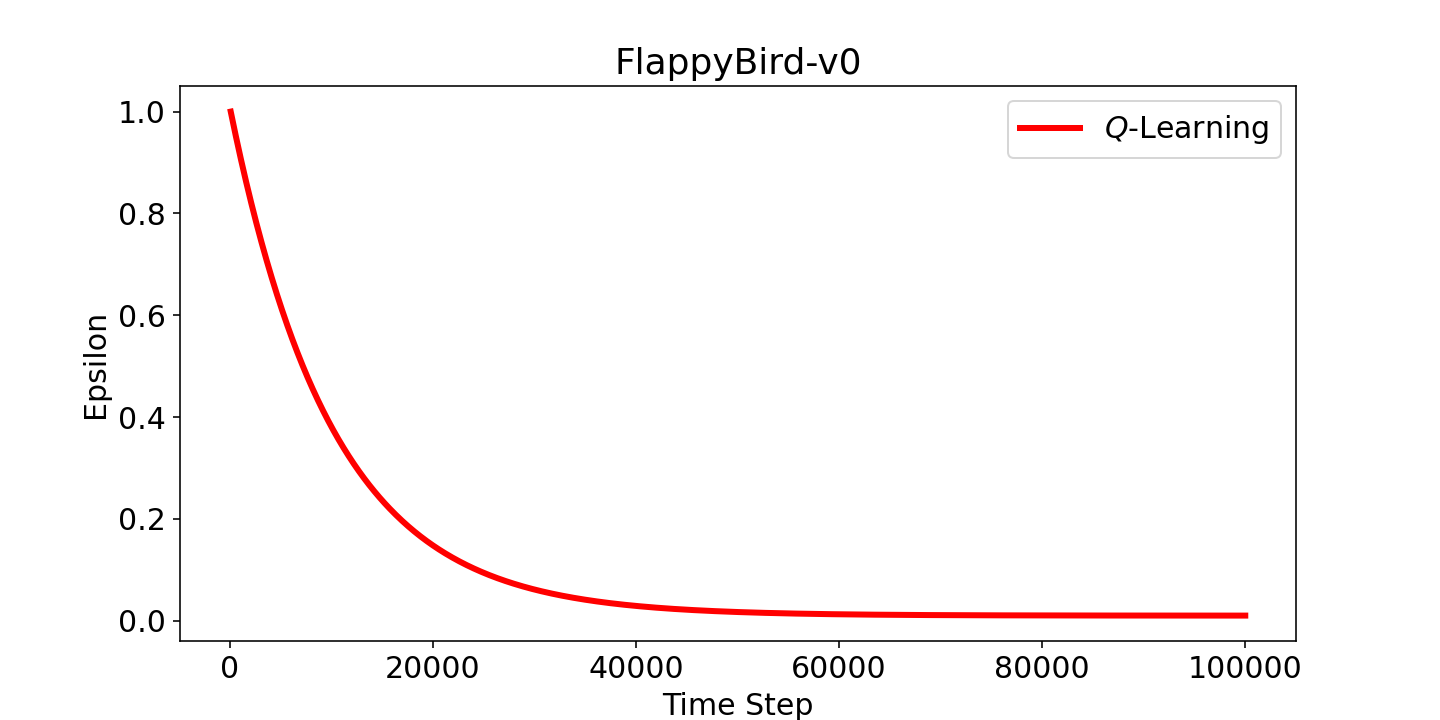}
         \caption{FlappyBird-v0}
         \label{Epsilon_vs_Time_Step_FlappyBird-v0}
     \end{subfigure}
        \caption{Decaying $\epsilon$ used for $Q$-Learning vs. Time steps. The $\epsilon$ value has been updated just before stating a new episode. }
        \label{Epsilon_vs_Time Step}
\end{figure}

\begin{figure}[H]%[t] %[h]
     \centering
     \begin{subfigure}{0.32\columnwidth}
         \centering
         \includegraphics[width=\columnwidth]{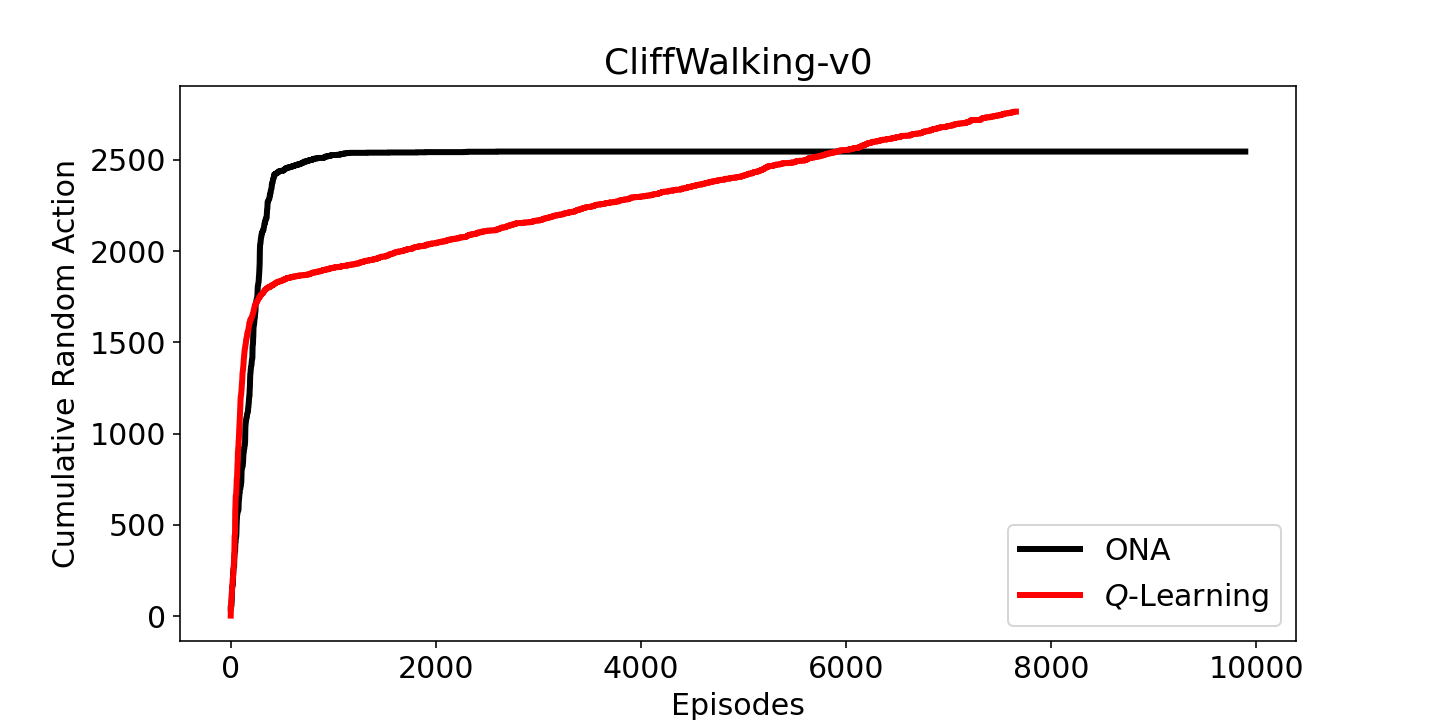}
         \caption{CliffWalking-v0}
         \label{Cumulative_Random_Action_vs_Episodes_CliffWalking-v0}
     \end{subfigure}
     \begin{subfigure}{0.32\columnwidth}
         \centering
         \includegraphics[width=\columnwidth]{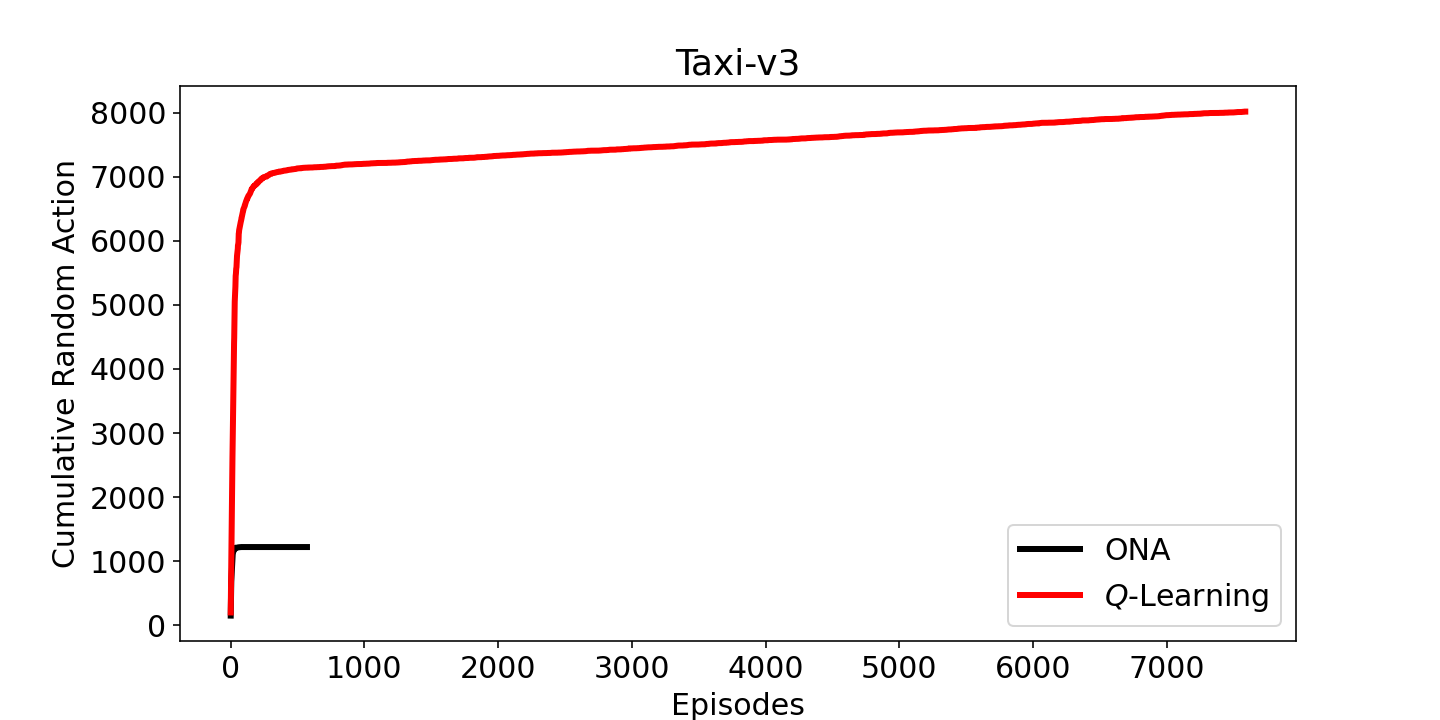}
         \caption{Taxi-v3}
         \label{Cumulative_Random_Action_vs_Episodes_Taxi-v3}
     \end{subfigure}
     \begin{subfigure}{0.32\columnwidth}
         \centering
         \includegraphics[width=\columnwidth]{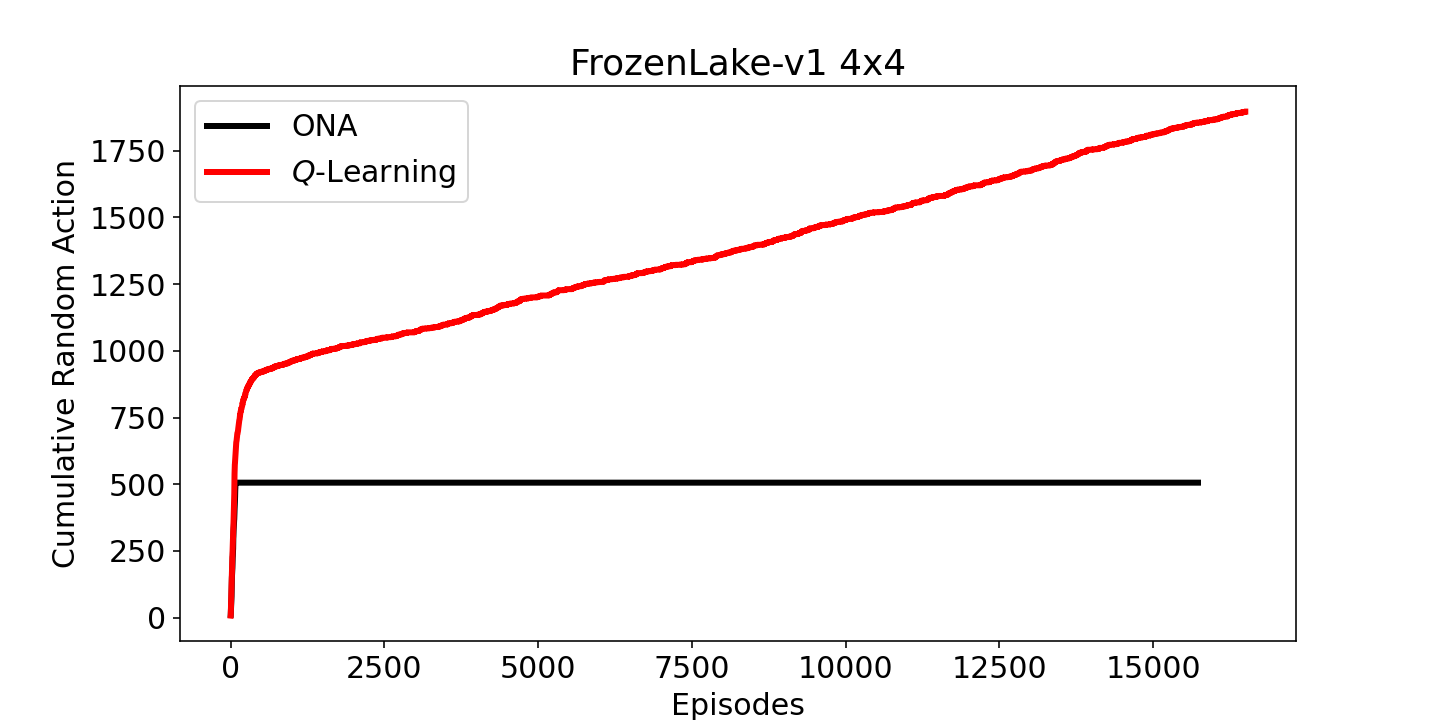}
         \caption{FrozenLake-v1 4x4}
         \label{Cumulative_Random_Action_vs_Episodes_FrozenLake-v1 4x4}
     \end{subfigure}
     \begin{subfigure}{0.32\columnwidth}
         \centering
         \includegraphics[width=\columnwidth]{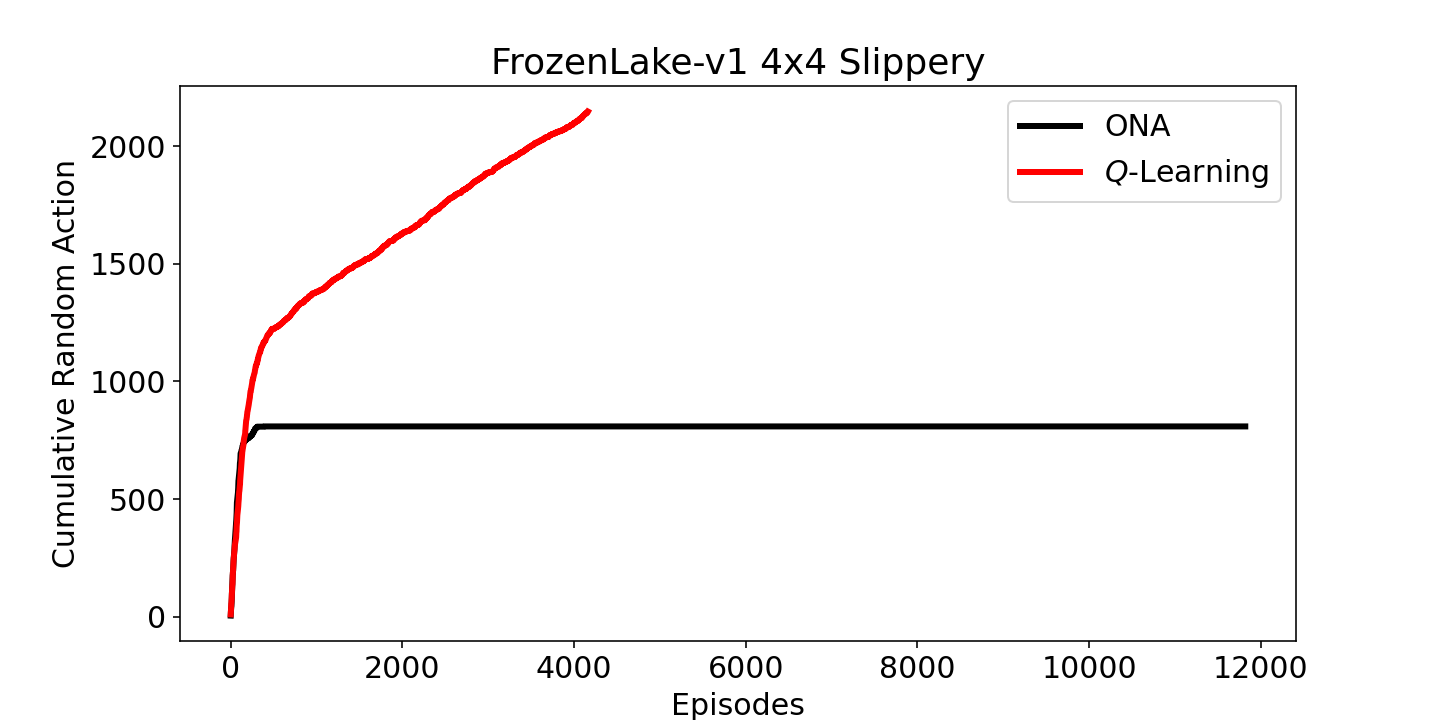}
         \caption{FrozenLake-v1 4x4 Slippery}
         \label{Cumulative_Random_Action_vs_Episodes_FrozenLake-v1_4x4_Slippery}
     \end{subfigure}
    \begin{subfigure}{0.32\columnwidth}
         \centering
         \includegraphics[width=\columnwidth]{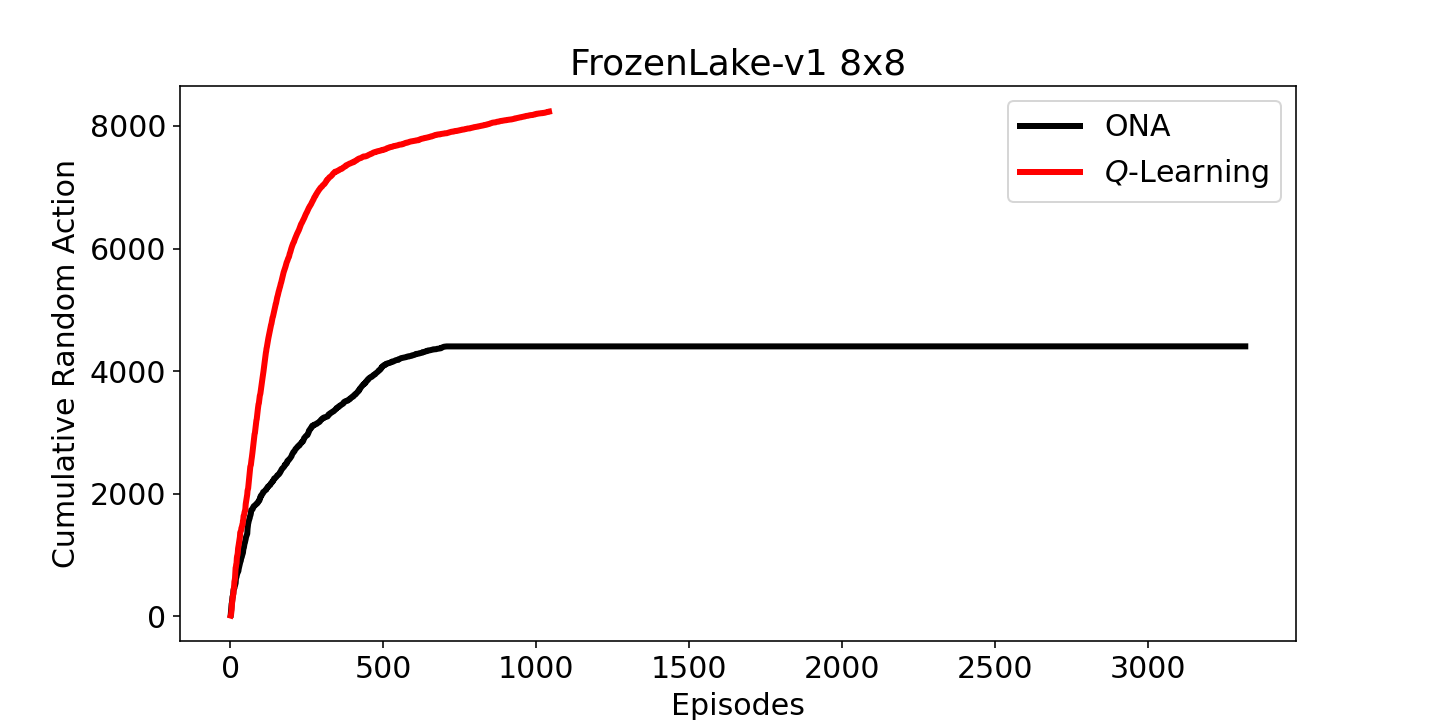}
         \caption{FrozenLake-v1 8x8}
         \label{Cumulative_Random_Action_vs_Episodes_FrozenLake-v1 8x8}
     \end{subfigure}
    \begin{subfigure}{0.32\columnwidth}
         \centering
         \includegraphics[width=\columnwidth]{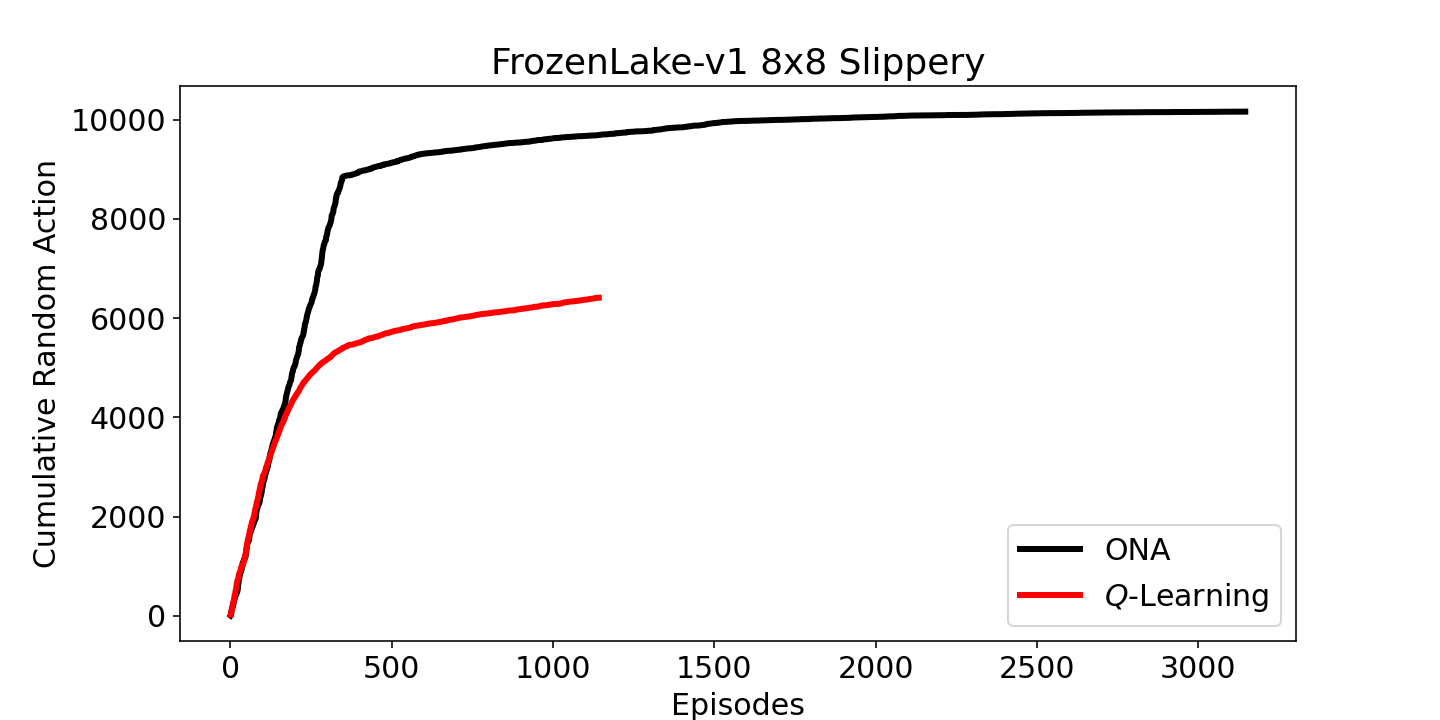}
         \caption{FrozenLake-v1 8x8 Slippery}
         \label{Cumulative_Random_Action_vs_Episodes_FrozenLake-v1_8x8_Slippery}
     \end{subfigure}
     \begin{subfigure}{0.32\columnwidth}
         \centering
         \includegraphics[width=\columnwidth]{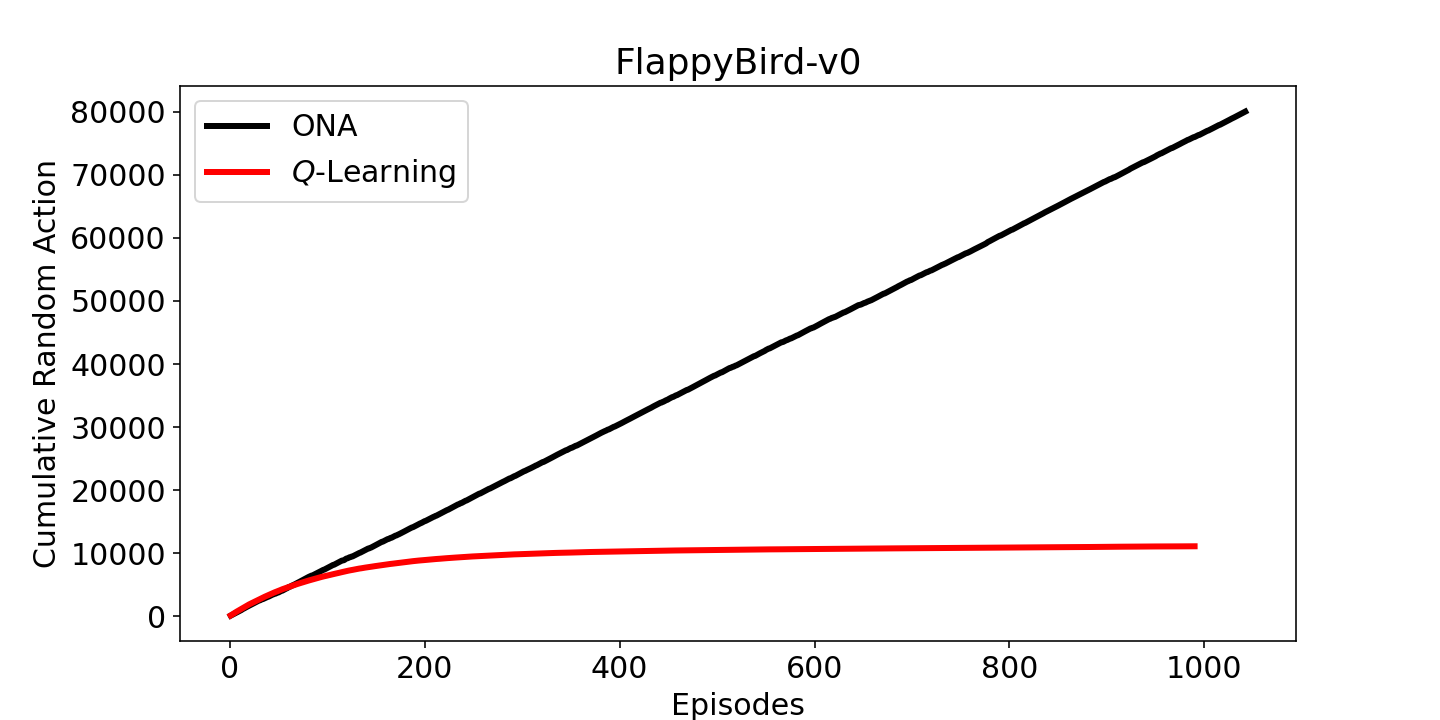}
         \caption{FlappyBird-v0}
         \label{Cumulative_Random_Action_vs_Episodes_FlappyBird-v0}
     \end{subfigure}
        \caption{Cumulative Random Action vs. Episodes.}
        \label{Cumulative_Random_Action_vs_Episodes}
\end{figure}

\begin{figure}[H]%[t] %[h]
     \centering
     \begin{subfigure}{0.32\columnwidth}
         \centering
         \includegraphics[width=\columnwidth]{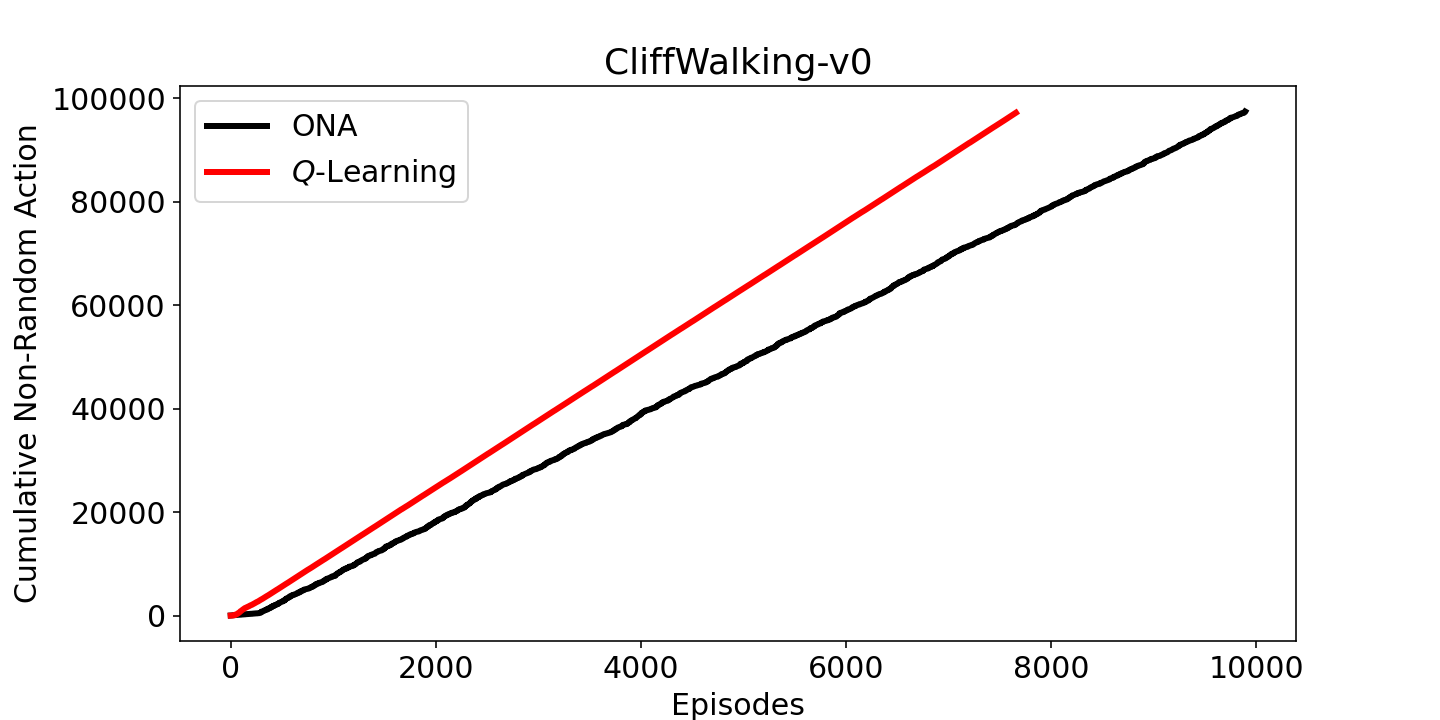}
         \caption{CliffWalking-v0}
         \label{Cumulative_Non-Random_Action_vs_Episodes_CliffWalking-v0}
     \end{subfigure}
     \begin{subfigure}{0.32\columnwidth}
         \centering
         \includegraphics[width=\columnwidth]{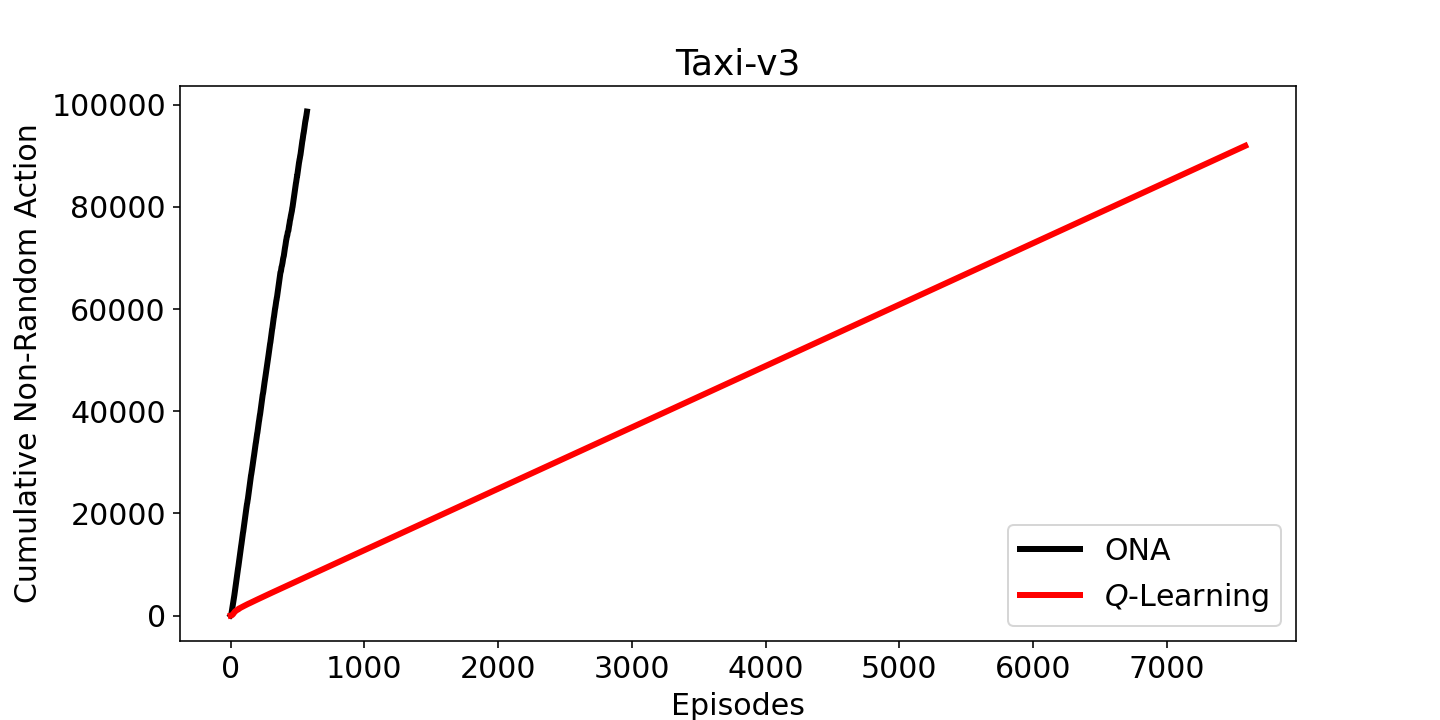}
         \caption{Taxi-v3}
         \label{Cumulative_Non-Random_Action_vs_Episodes_Taxi-v3}
     \end{subfigure}
     \begin{subfigure}{0.32\columnwidth}
         \centering
         \includegraphics[width=\columnwidth]{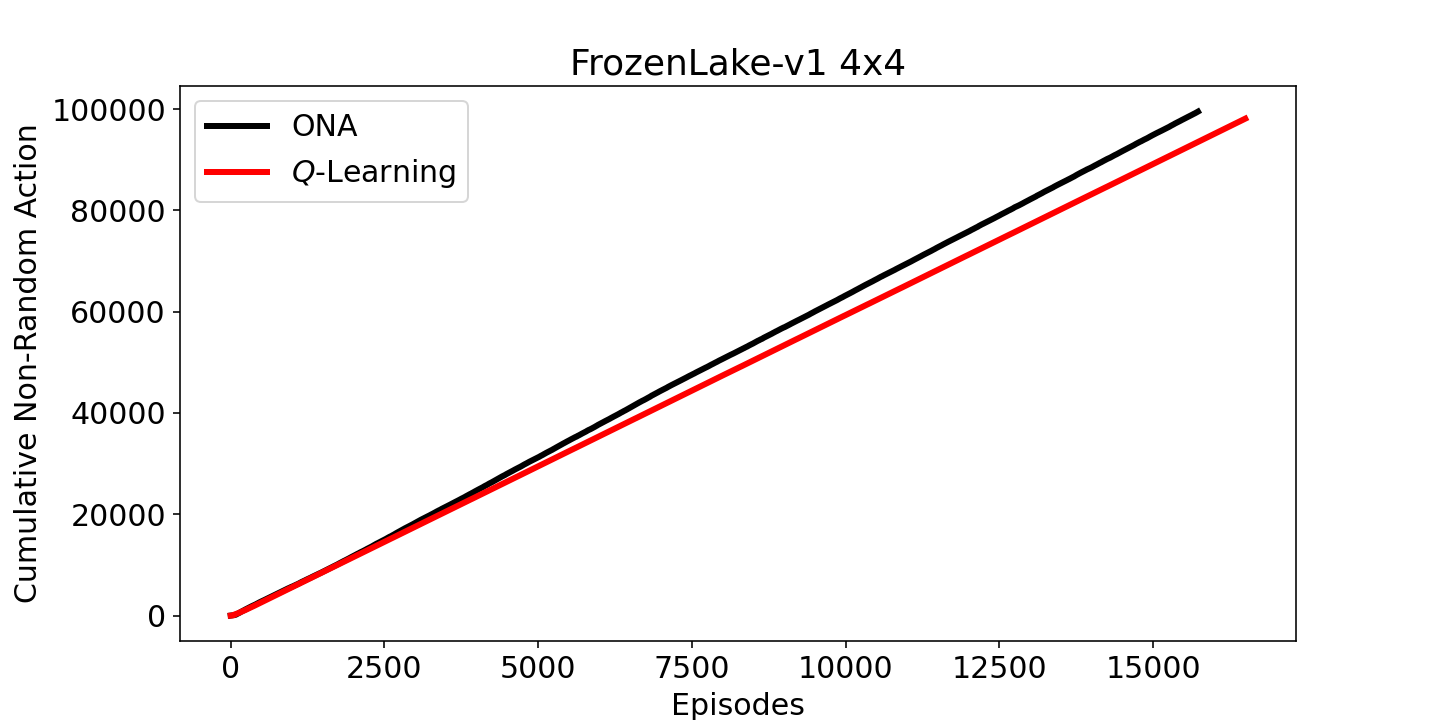}
         \caption{FrozenLake-v1 4x4}
         \label{Cumulative_Non-Random_Action_vs_Episodes_FrozenLake-v1 4x4}
     \end{subfigure}
     \begin{subfigure}{0.32\columnwidth}
         \centering
         \includegraphics[width=\columnwidth]{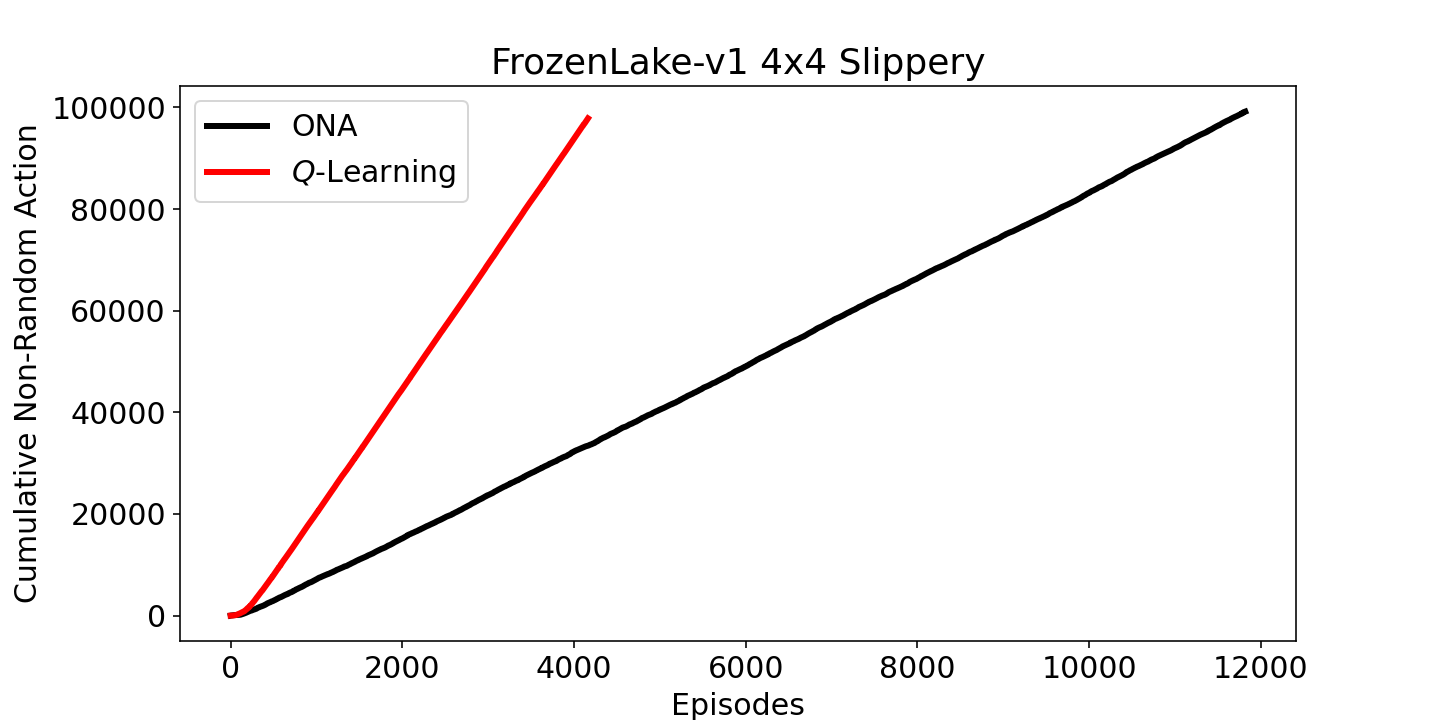}
         \caption{FrozenLake-v1 4x4 Slippery}
         \label{Cumulative_Non-Random_Action_vs_Episodes_FrozenLake-v1_4x4_Slippery}
     \end{subfigure}
    \begin{subfigure}{0.32\columnwidth}
         \centering
         \includegraphics[width=\columnwidth]{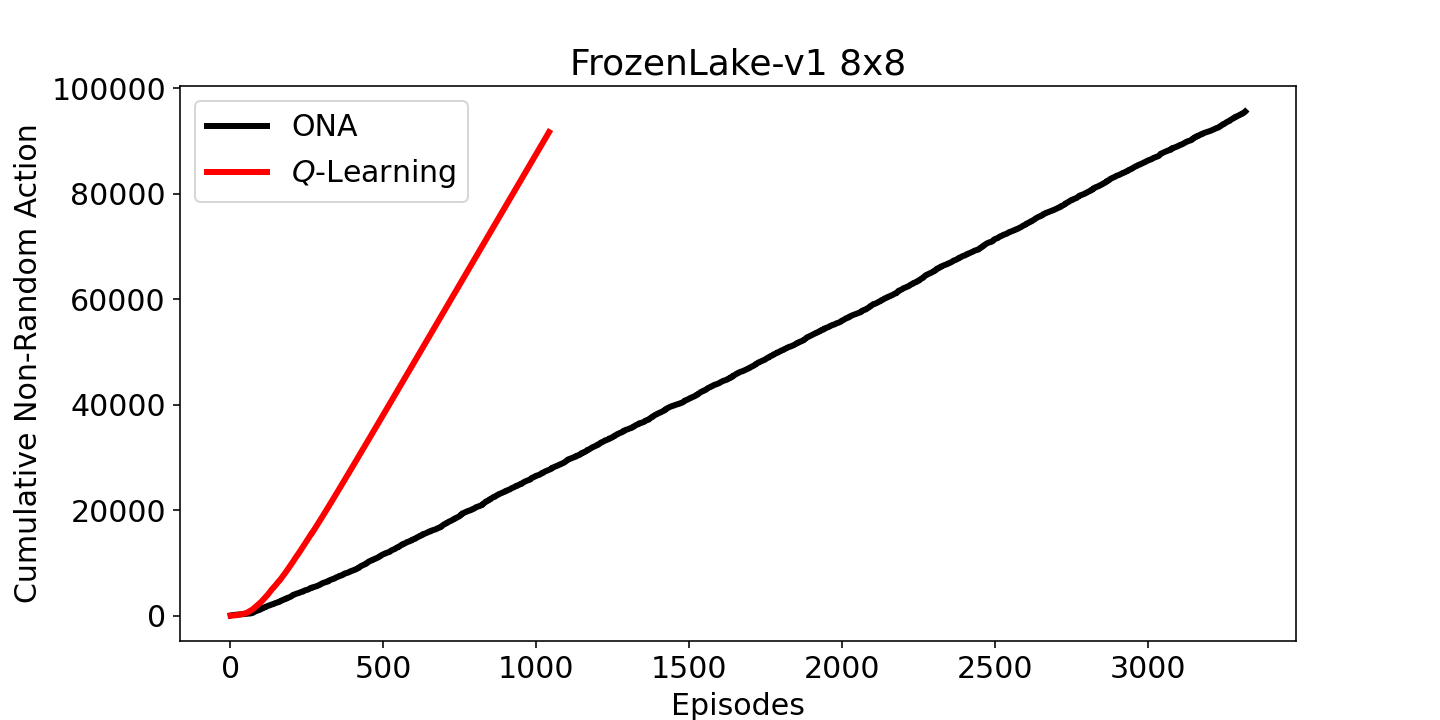}
         \caption{FrozenLake-v1 8x8}
         \label{Cumulative_Non-Random_Action_vs_Episodes_FrozenLake-v1 8x8}
     \end{subfigure}
    \begin{subfigure}{0.32\columnwidth}
         \centering
         \includegraphics[width=\columnwidth]{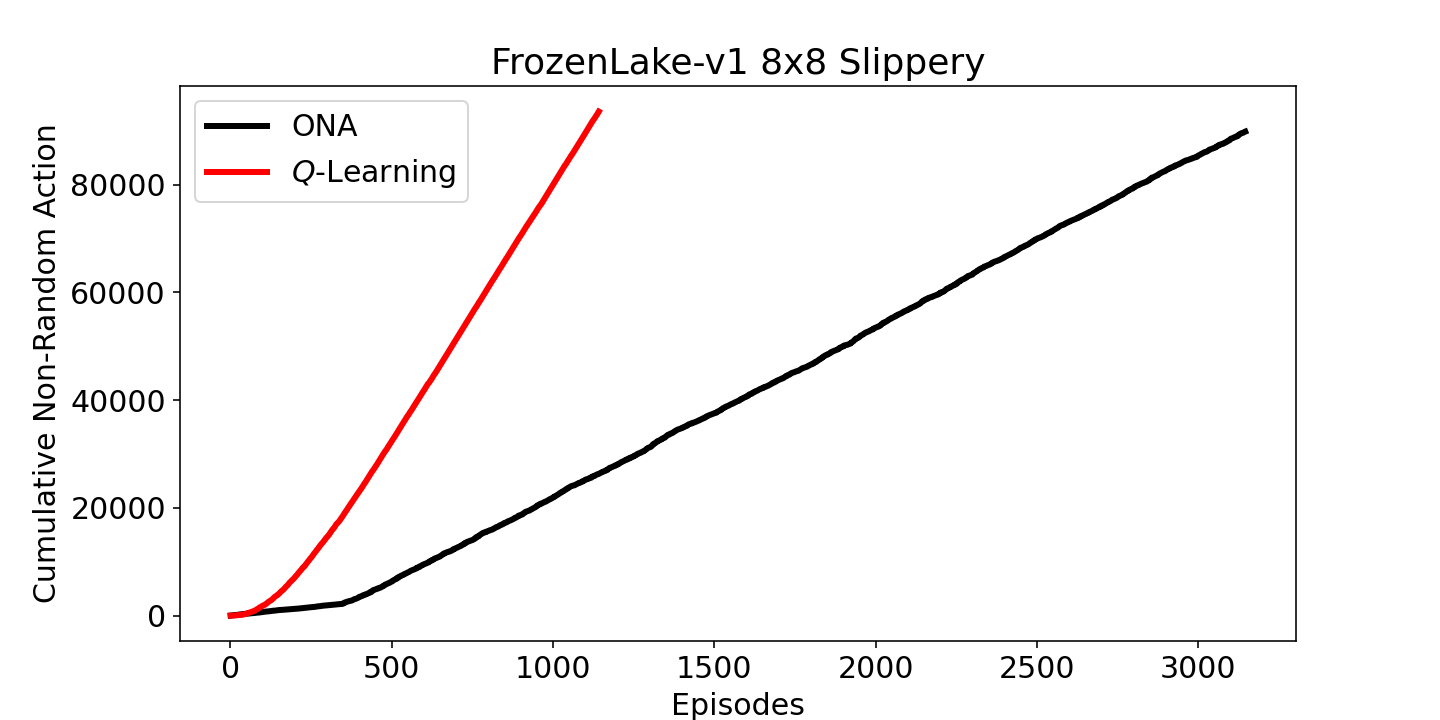}
         \caption{FrozenLake-v1 8x8 Slippery}
         \label{Cumulative_Non-Random_Action_vs_Episodes_FrozenLake-v1_8x8_Slippery}
     \end{subfigure}
     \begin{subfigure}{0.32\columnwidth}
         \centering
         \includegraphics[width=\columnwidth]{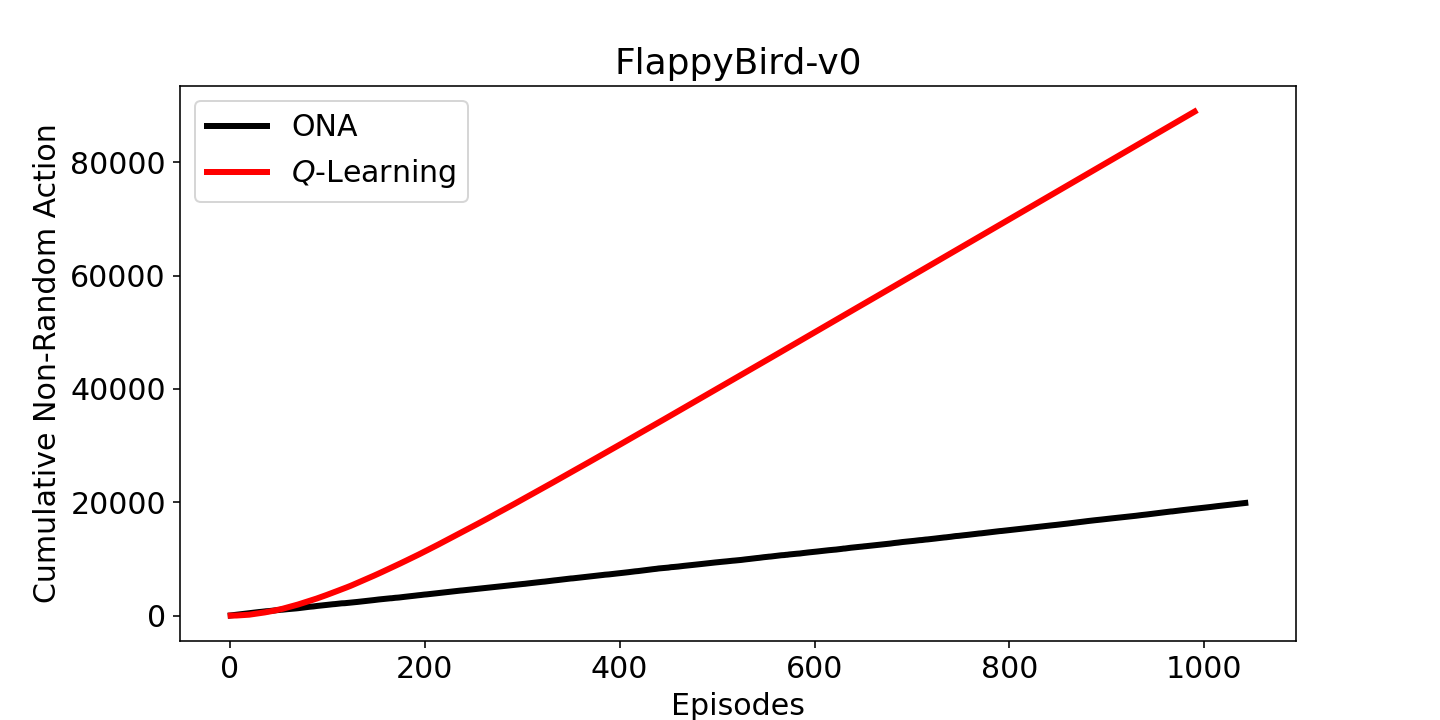}
         \caption{FlappyBird-v0}
         \label{Cumulative_Non-Random_Action_vs_Episodes_FlappyBird-v0}
     \end{subfigure}
        \caption{Cumulative Non-Random Action vs. Episodes.}
        \label{Cumulative_Non-Random_Action_vs_Episodes}
\end{figure}

\begin{figure}[H]%[t] %[h]
     \centering
     \begin{subfigure}{0.32\columnwidth}
         \centering
         \includegraphics[width=\columnwidth]{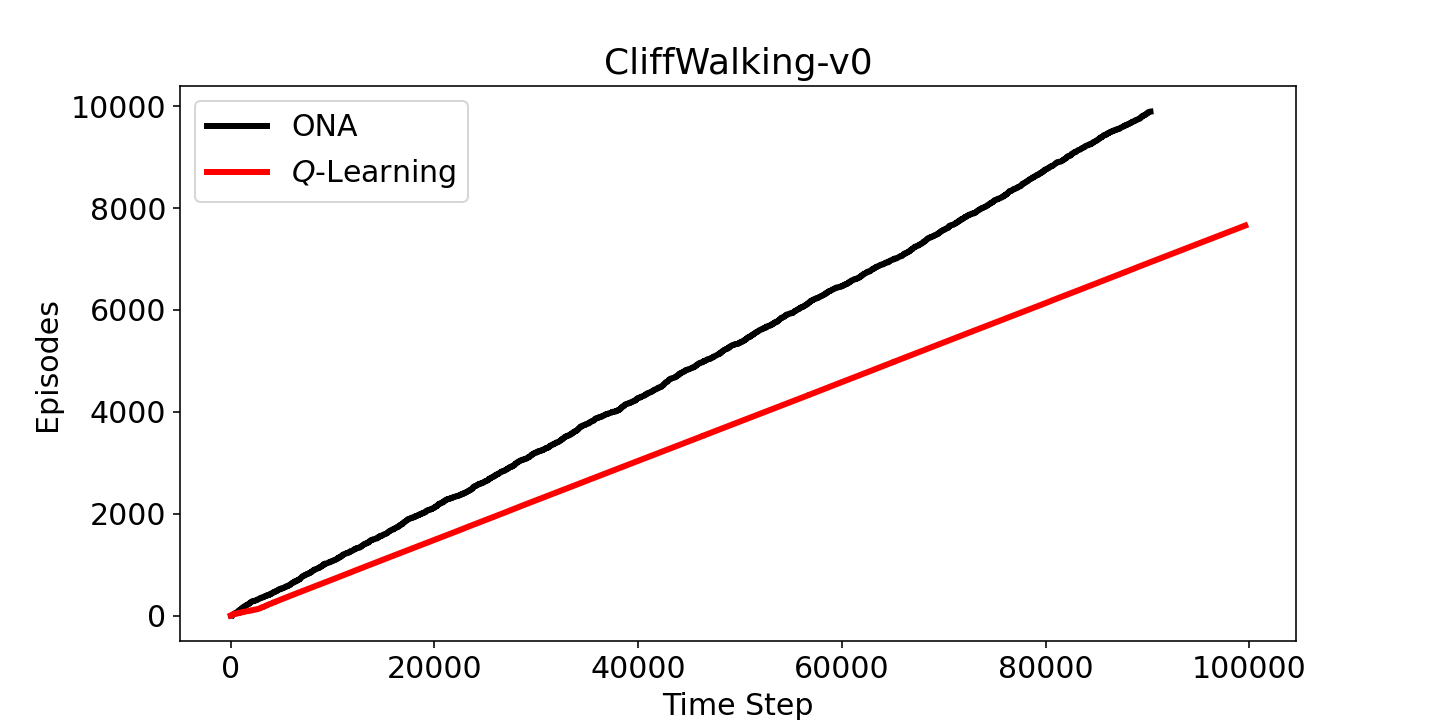}
         \caption{CliffWalking-v0}
         \label{Episodes_vs_Time_Step_CliffWalking-v0}
     \end{subfigure}
     \begin{subfigure}{0.32\columnwidth}
         \centering
         \includegraphics[width=\columnwidth]{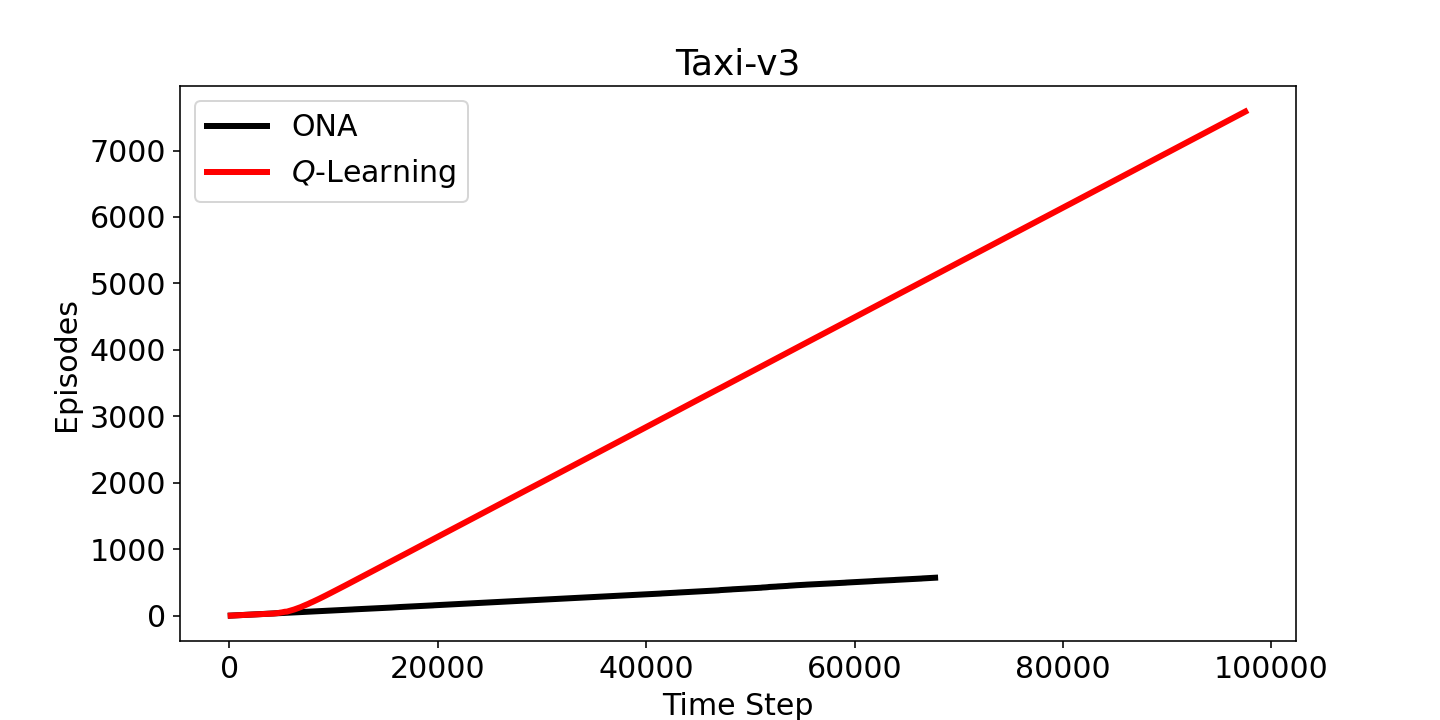}
         \caption{Taxi-v3}
         \label{Episodes_vs_Time_Step_Taxi-v3}
     \end{subfigure}
     \begin{subfigure}{0.32\columnwidth}
         \centering
         \includegraphics[width=\columnwidth]{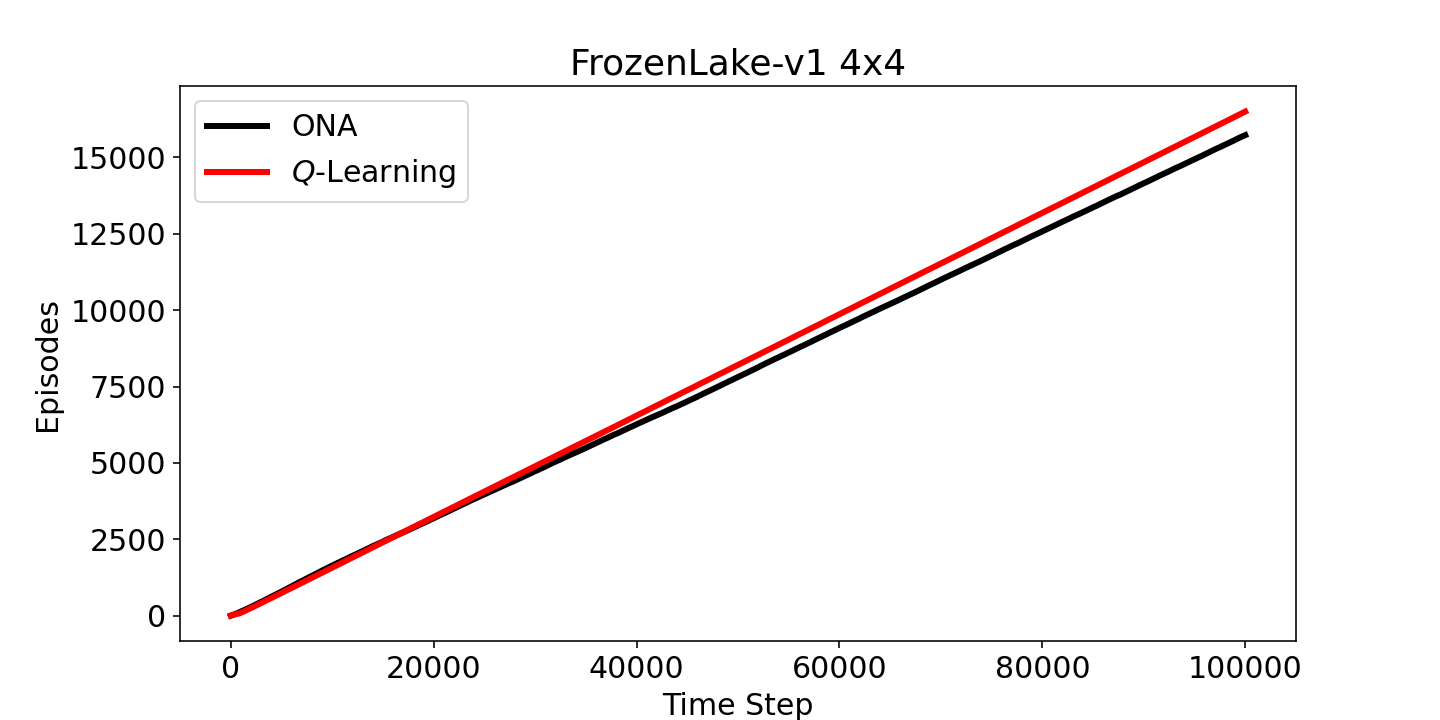}
         \caption{FrozenLake-v1 4x4}
         \label{Episodes_vs_Time_Step_FrozenLake-v1 4x4}
     \end{subfigure}
     \begin{subfigure}{0.32\columnwidth}
         \centering
         \includegraphics[width=\columnwidth]{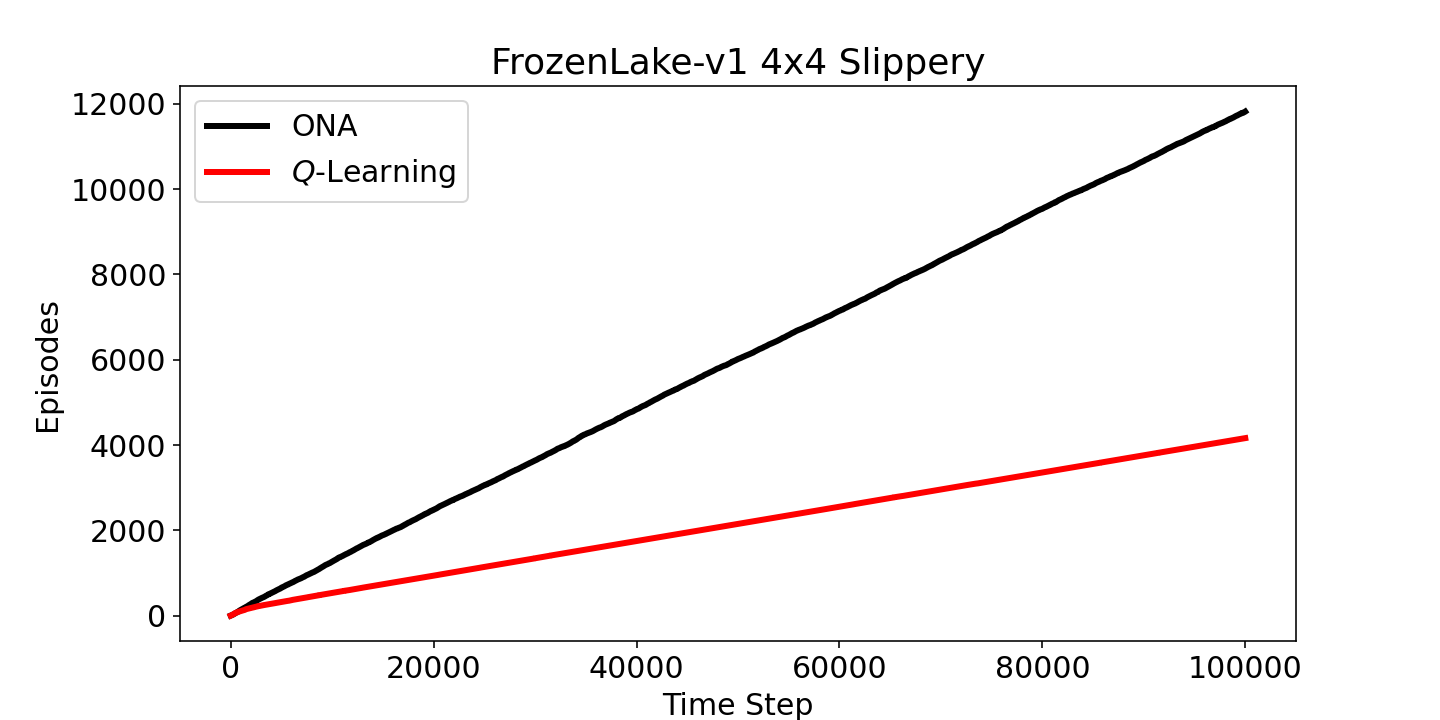}
         \caption{FrozenLake-v1 4x4 Slippery}
         \label{Episodes_vs_Time_Step_FrozenLake-v1_4x4_Slippery}
     \end{subfigure}
    \begin{subfigure}{0.32\columnwidth}
         \centering
         \includegraphics[width=\columnwidth]{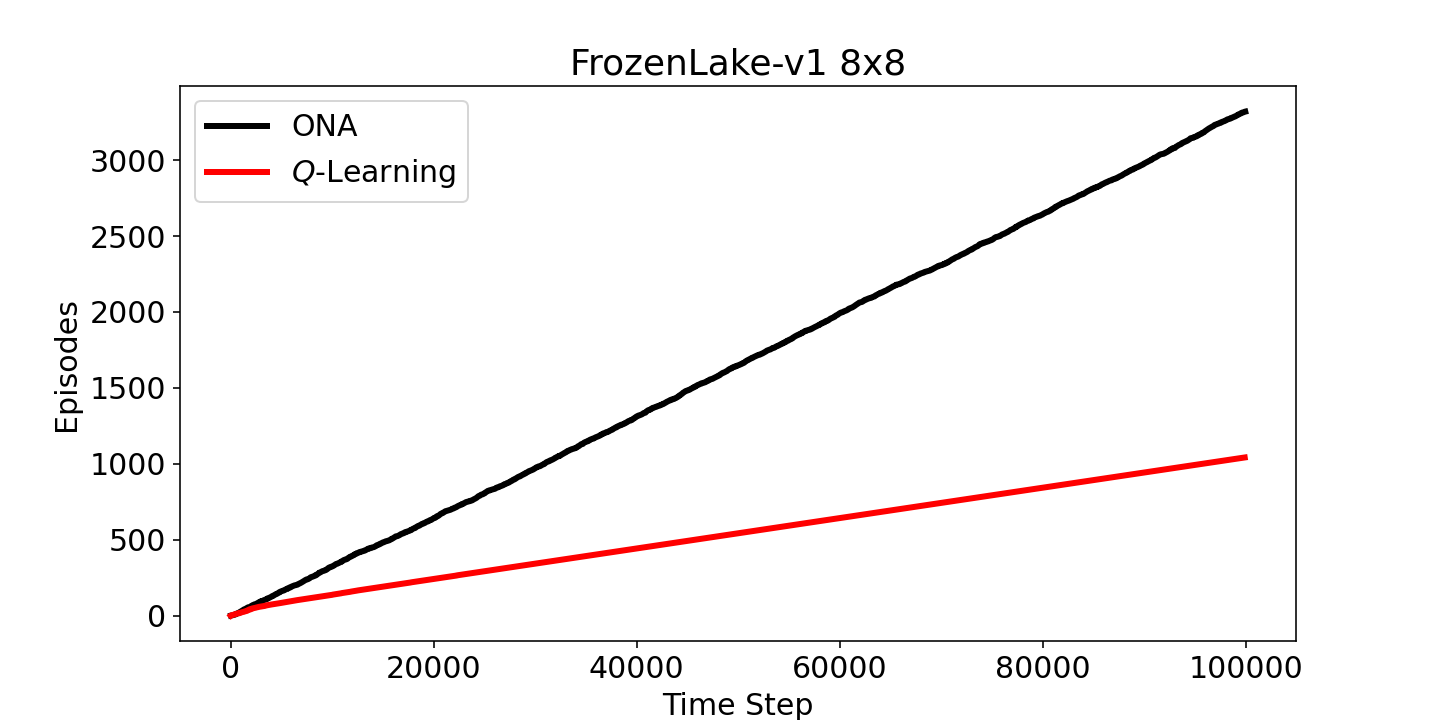}
         \caption{FrozenLake-v1 8x8}
         \label{Episodes_vs_Time_Step_FrozenLake-v1 8x8}
     \end{subfigure}
    \begin{subfigure}{0.32\columnwidth}
         \centering
         \includegraphics[width=\columnwidth]{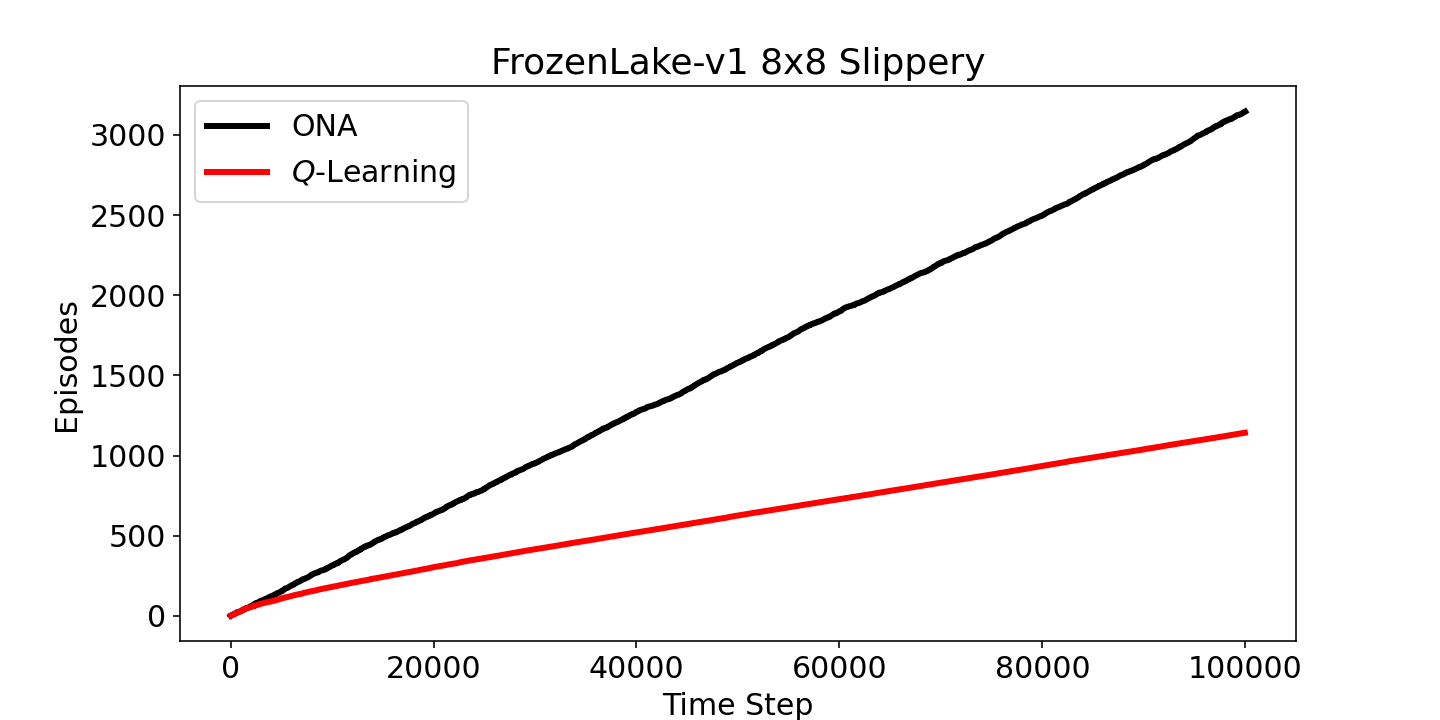}
         \caption{FrozenLake-v1 8x8 Slippery}
         \label{Episodes_vs_Time_Step_FrozenLake-v1_8x8_Slippery}
     \end{subfigure}
     \begin{subfigure}{0.32\columnwidth}
         \centering
         \includegraphics[width=\columnwidth]{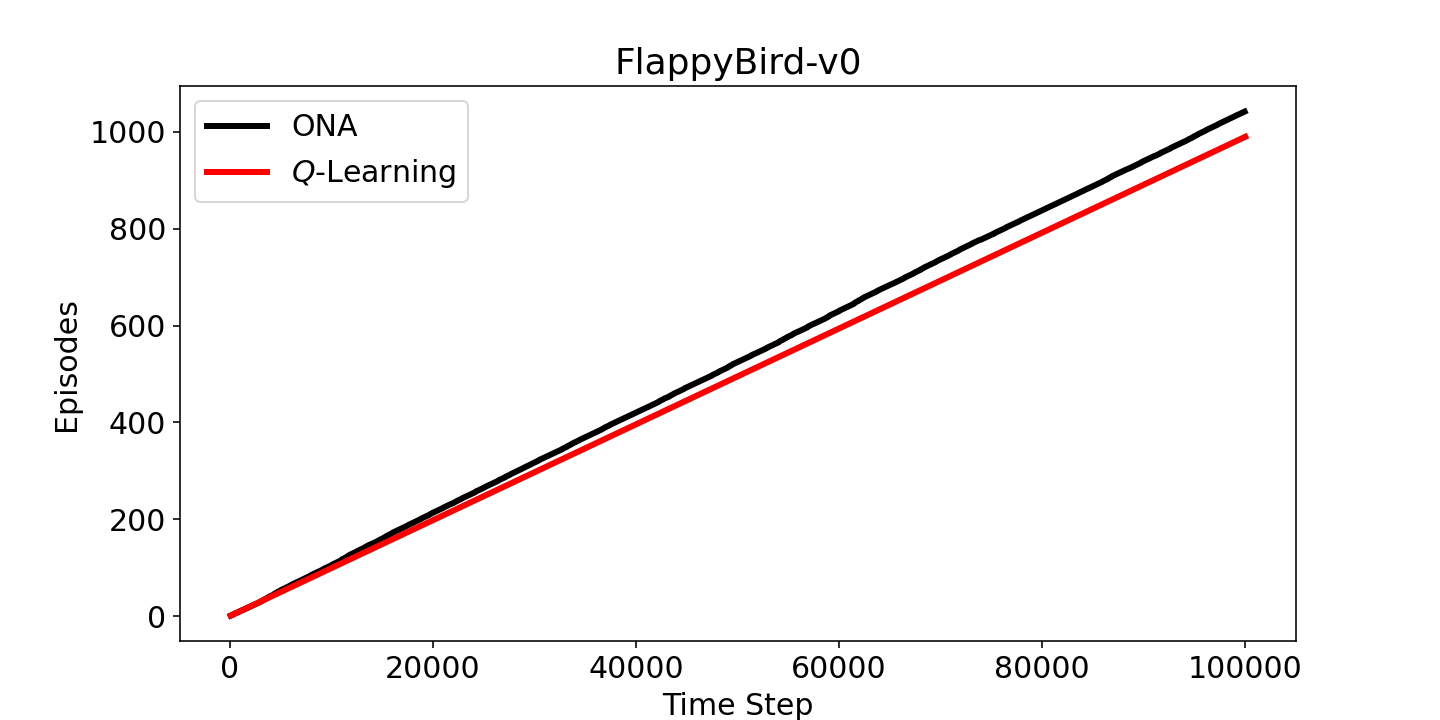}
         \caption{FlappyBird-v0}
         \label{Episodes_vs_Time_Step_FlappyBird-v0}
     \end{subfigure}
        \caption{Episodes vs. Time steps.}
        \label{Episodes_vs_Time Step}
\end{figure}

\begin{figure}[h]%[t] %[h]
     \centering
     \begin{subfigure}{0.32\columnwidth}
         \centering
         \includegraphics[width=\columnwidth]{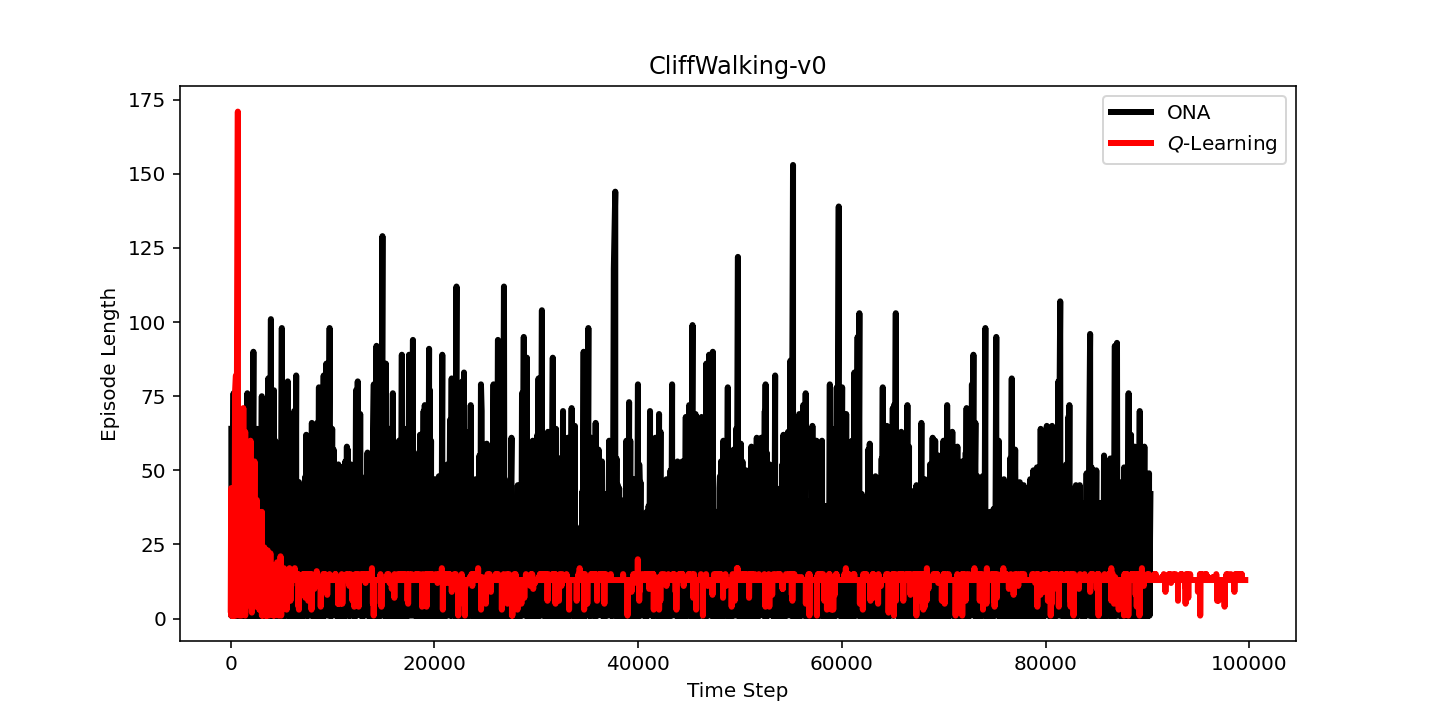}
         \caption{CliffWalking-v0}
         \label{Episode_Length_vs_Time_Step_CliffWalking-v0}
     \end{subfigure}
     \begin{subfigure}{0.32\columnwidth}
         \centering
         \includegraphics[width=\columnwidth]{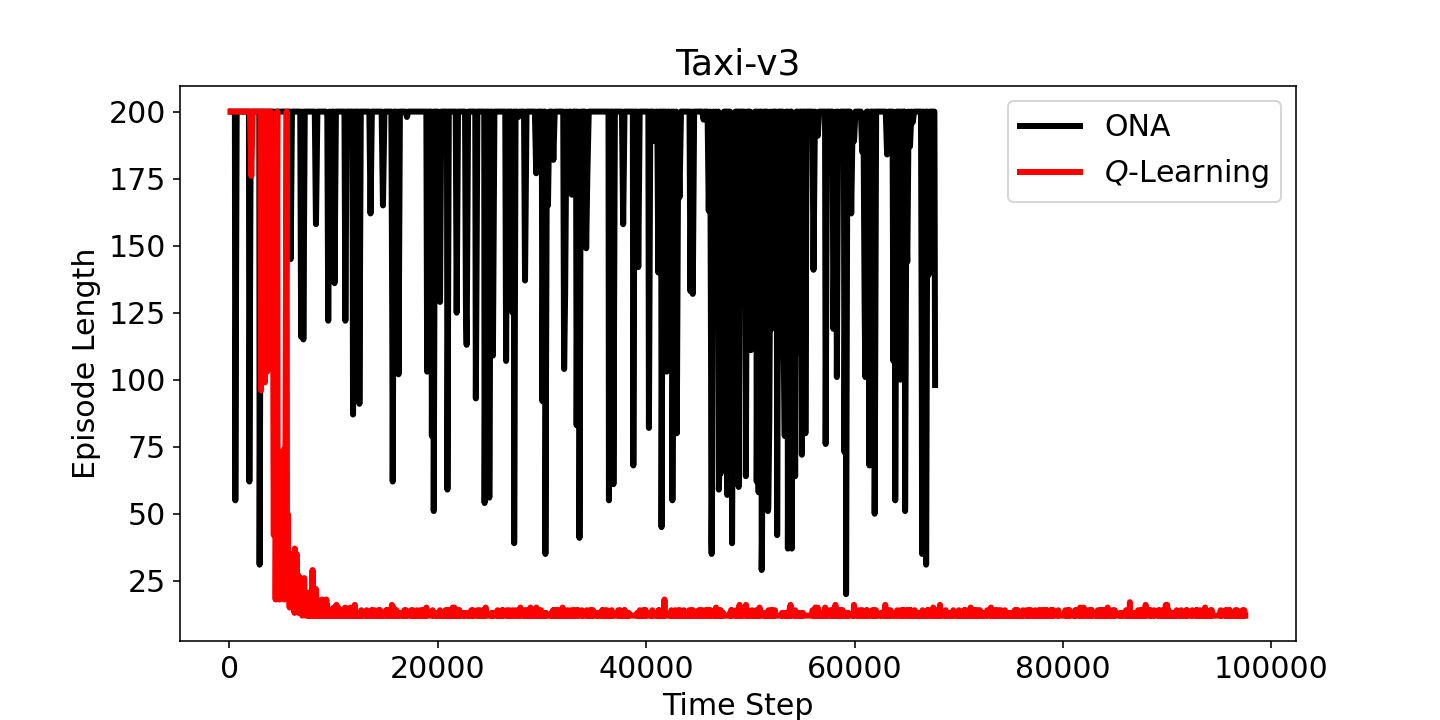}
         \caption{Taxi-v3}
         \label{Episode_Length_vs_Time_Step_Taxi-v3}
     \end{subfigure}
     \begin{subfigure}{0.32\columnwidth}
         \centering
         \includegraphics[width=\columnwidth]{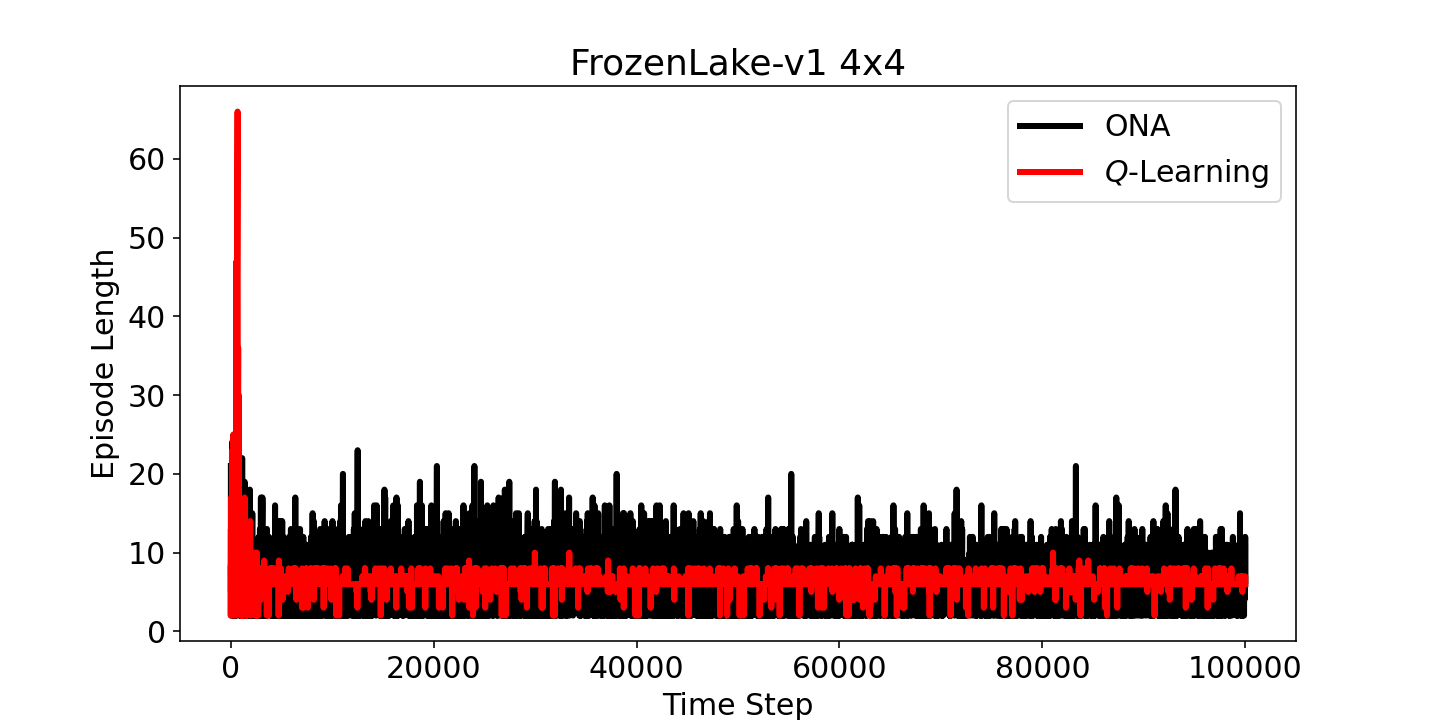}
         \caption{FrozenLake-v1 4x4}
         \label{Episode_Length_vs_Time_Step_FrozenLake-v1 4x4}
     \end{subfigure}
     \begin{subfigure}{0.32\columnwidth}
         \centering
         \includegraphics[width=\columnwidth]{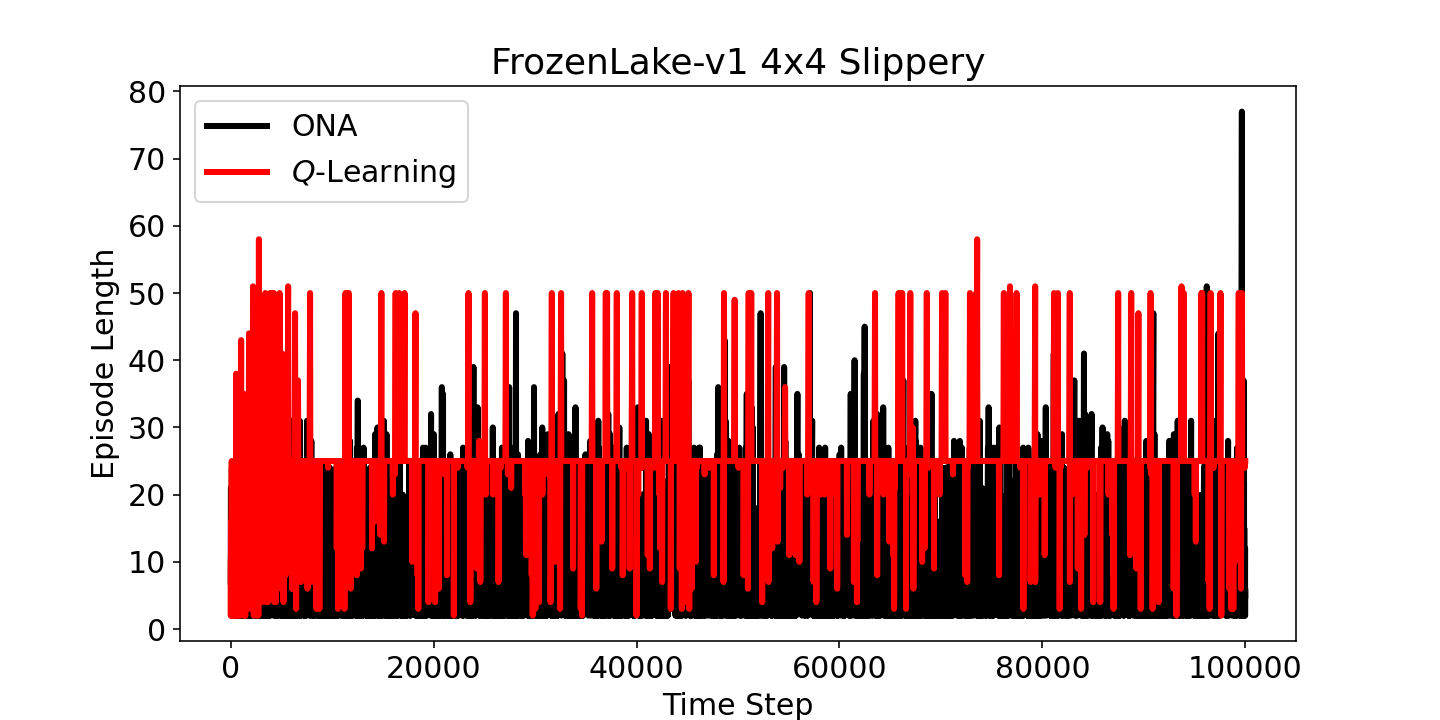}
         \caption{FrozenLake-v1 4x4 Slippery}
         \label{Episode_Length_vs_Time_Step_FrozenLake-v1_4x4_Slippery}
     \end{subfigure}
    \begin{subfigure}{0.32\columnwidth}
         \centering
         \includegraphics[width=\columnwidth]{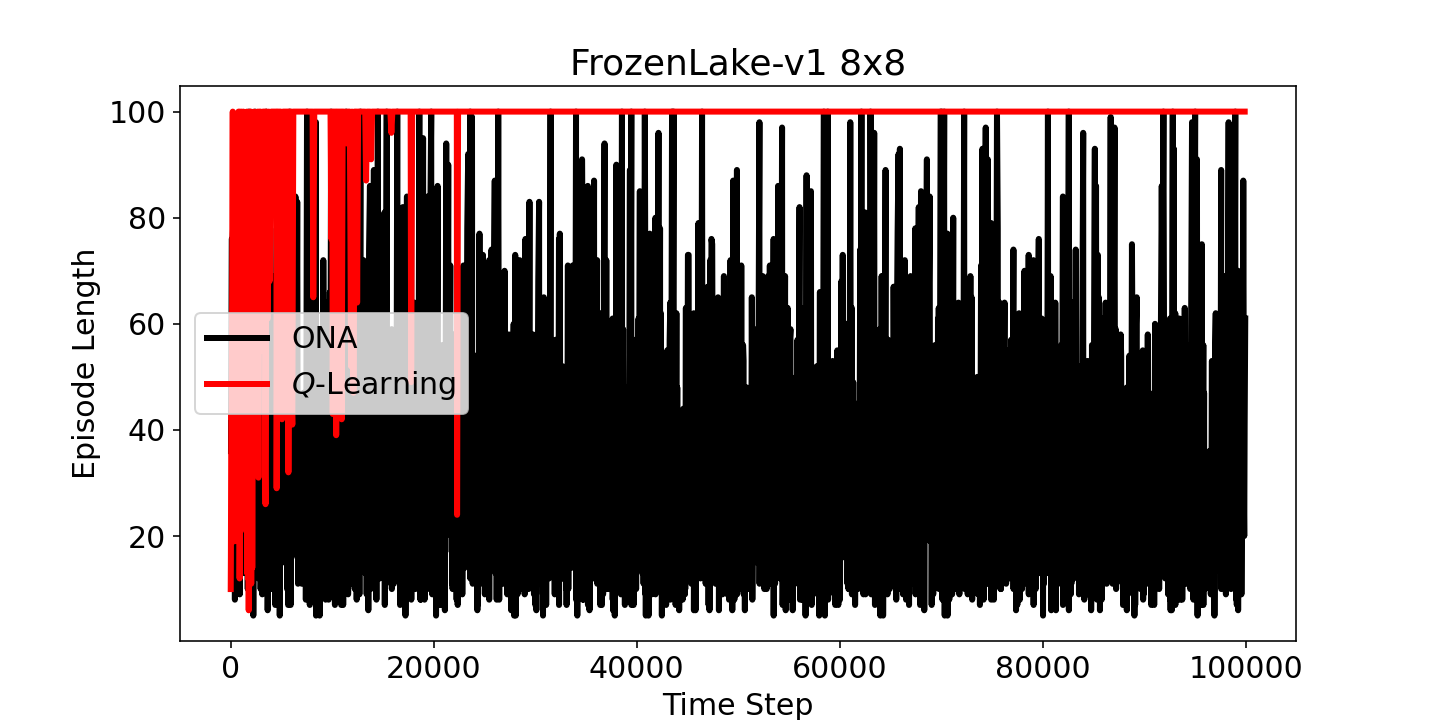}
         \caption{FrozenLake-v1 8x8}
         \label{Episode_Length_vs_Time_Step_FrozenLake-v1 8x8}
     \end{subfigure}
    \begin{subfigure}{0.32\columnwidth}
         \centering
         \includegraphics[width=\columnwidth]{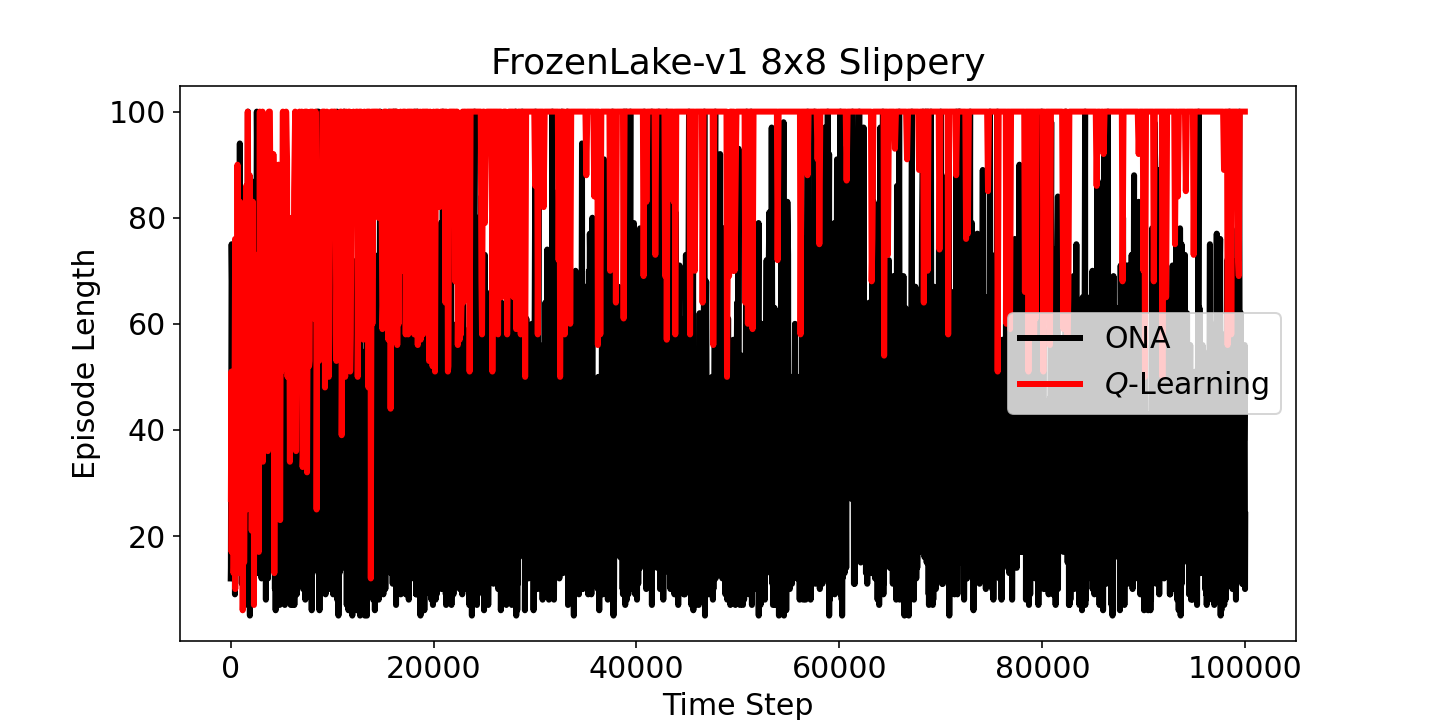}
         \caption{FrozenLake-v1 8x8 Slippery}
         \label{Episode_Length_vs_Time_Step_FrozenLake-v1_8x8_Slippery}
     \end{subfigure}
     \begin{subfigure}{0.32\columnwidth}
         \centering
         \includegraphics[width=\columnwidth]{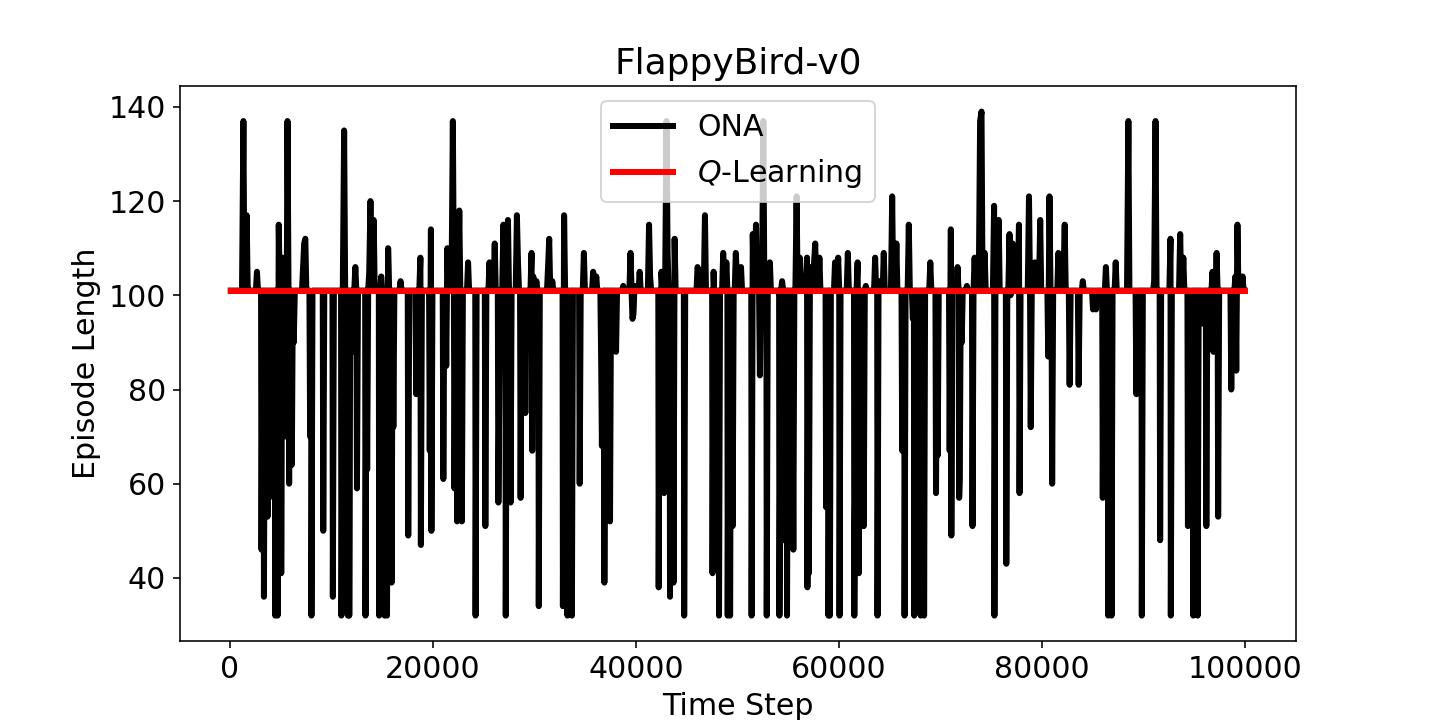}
         \caption{FlappyBird-v0}
         \label{Episode_Length_vs_Time_Step_FlappyBird-v0}
     \end{subfigure}
        \caption{Episode Length vs. Time steps.}
        \label{Episode_Length_vs_Time Step}
\end{figure}

\begin{figure}[H]%[t] %[h]
     \centering
     \begin{subfigure}{0.32\columnwidth}
         \centering
         \includegraphics[width=\columnwidth]{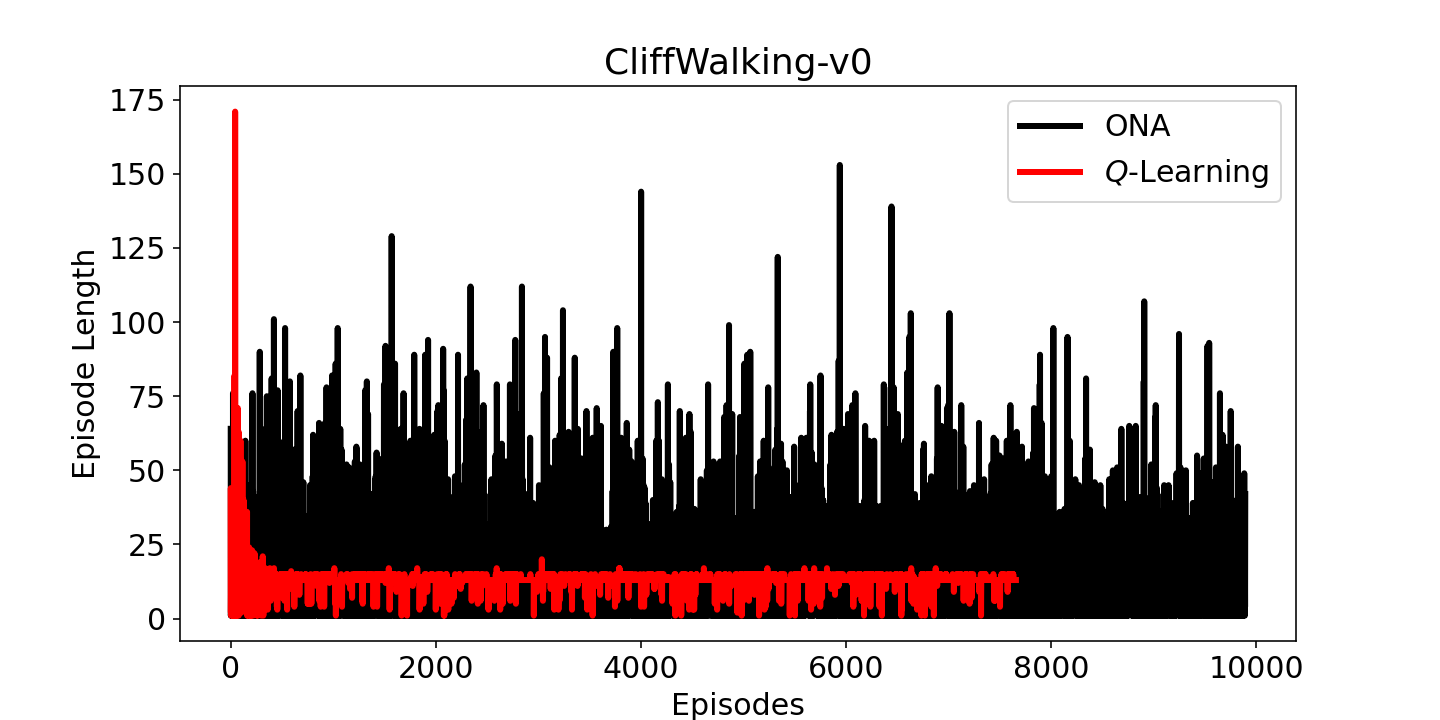}
         \caption{CliffWalking-v0}
         \label{Episode_Length_vs_Episodes_CliffWalking-v0}
     \end{subfigure}
     \begin{subfigure}{0.32\columnwidth}
         \centering
         \includegraphics[width=\columnwidth]{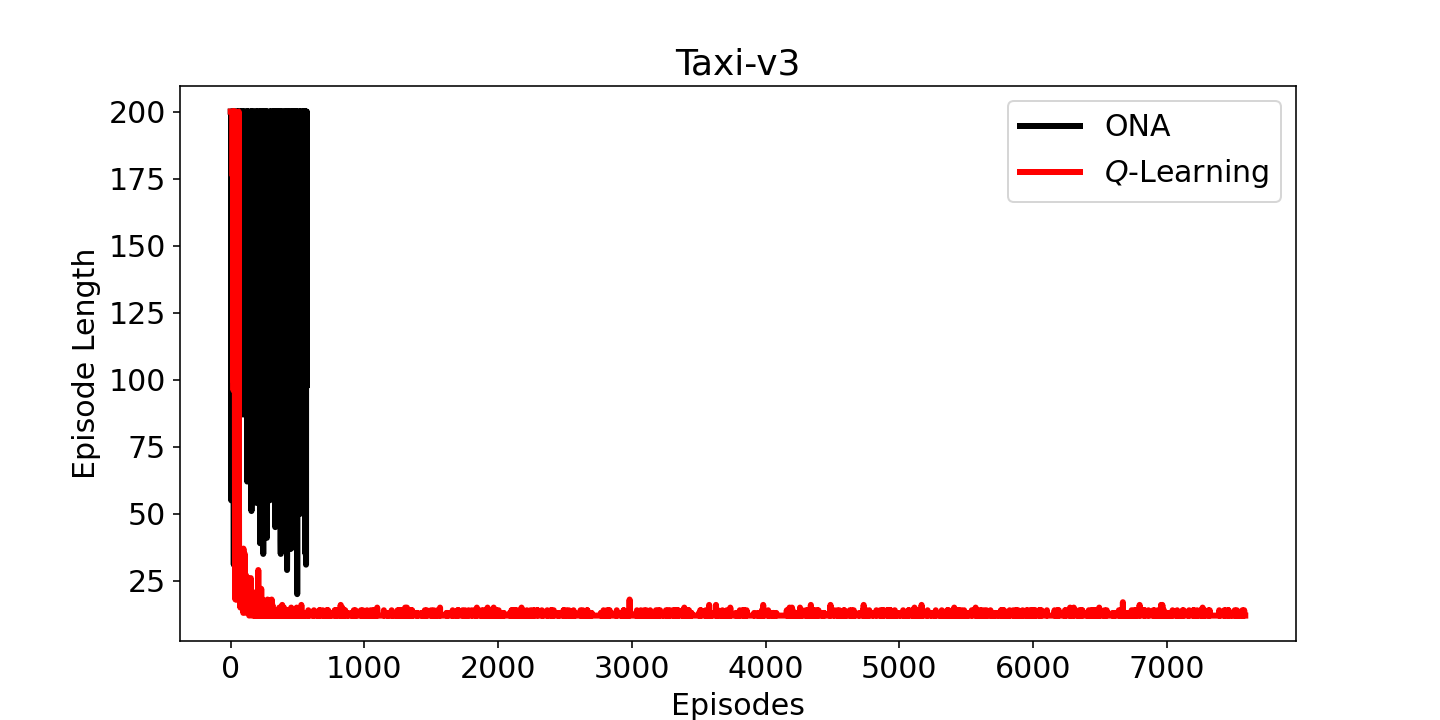}
         \caption{Taxi-v3}
         \label{Episode_Length_vs_Episodes_Taxi-v3}
     \end{subfigure}
     \begin{subfigure}{0.32\columnwidth}
         \centering
         \includegraphics[width=\columnwidth]{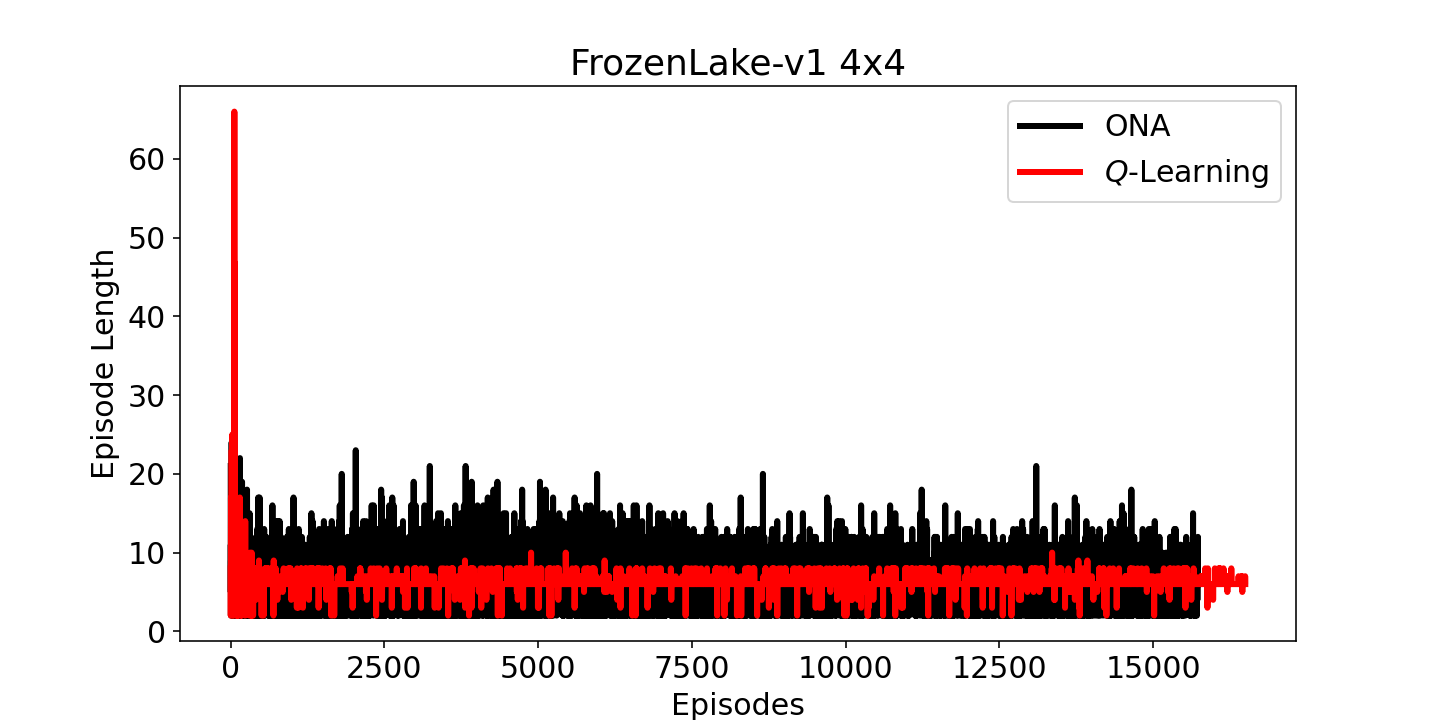}
         \caption{FrozenLake-v1 4x4}
         \label{Episode_Length_vs_Episodes_FrozenLake-v1 4x4}
     \end{subfigure}
     \begin{subfigure}{0.32\columnwidth}
         \centering
         \includegraphics[width=\columnwidth]{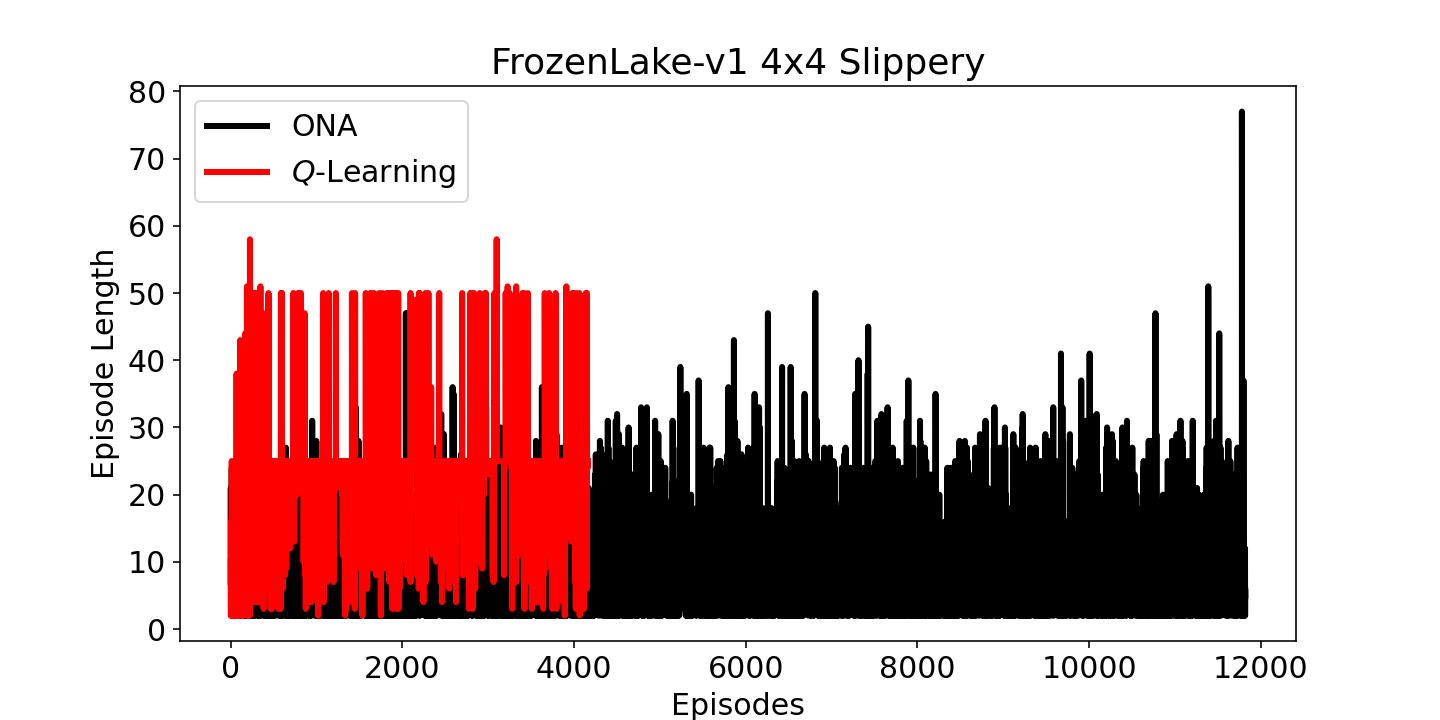}
         \caption{FrozenLake-v1 4x4 Slippery}
         \label{Episode_Length_vs_Episodes_FrozenLake-v1_4x4_Slippery}
     \end{subfigure}
    \begin{subfigure}{0.32\columnwidth}
         \centering
         \includegraphics[width=\columnwidth]{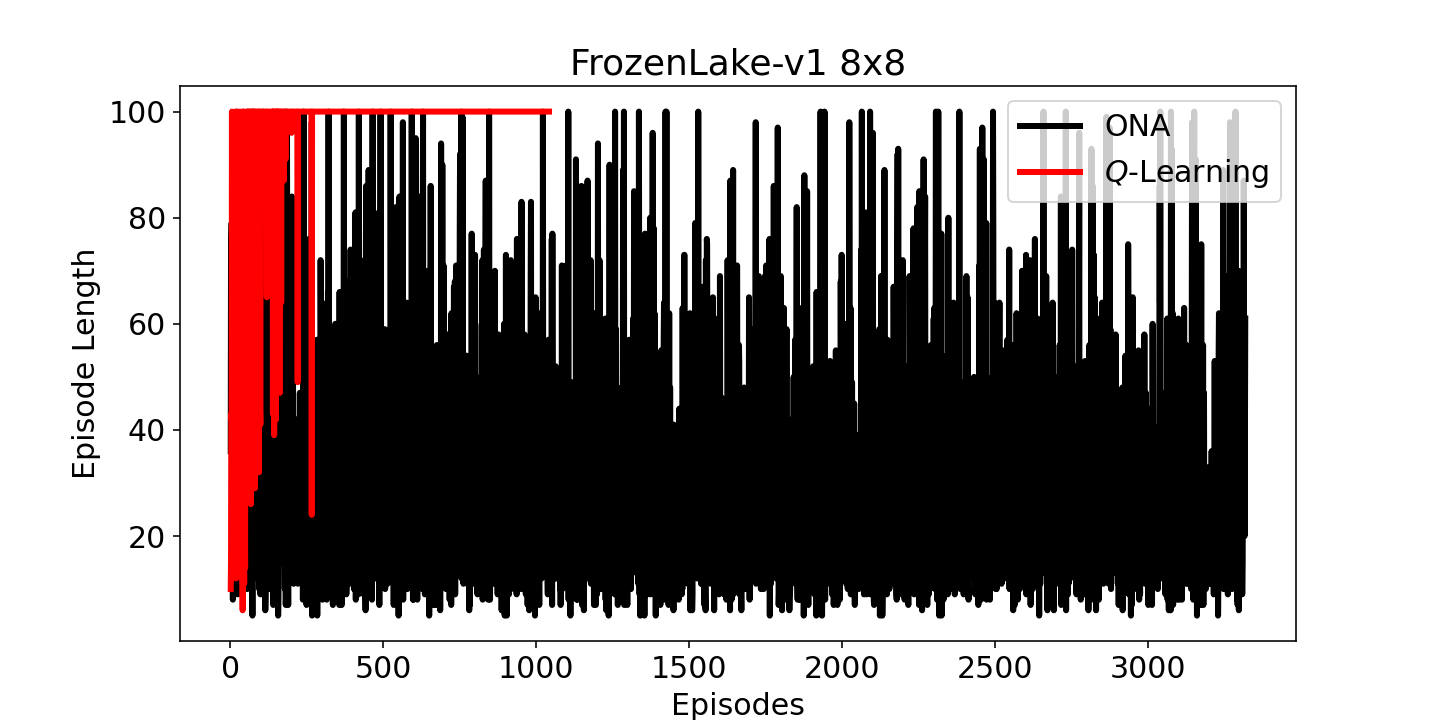}
         \caption{FrozenLake-v1 8x8}
         \label{Episode_Length_vs_Episodes_FrozenLake-v1 8x8}
     \end{subfigure}
    \begin{subfigure}{0.32\columnwidth}
         \centering
         \includegraphics[width=\columnwidth]{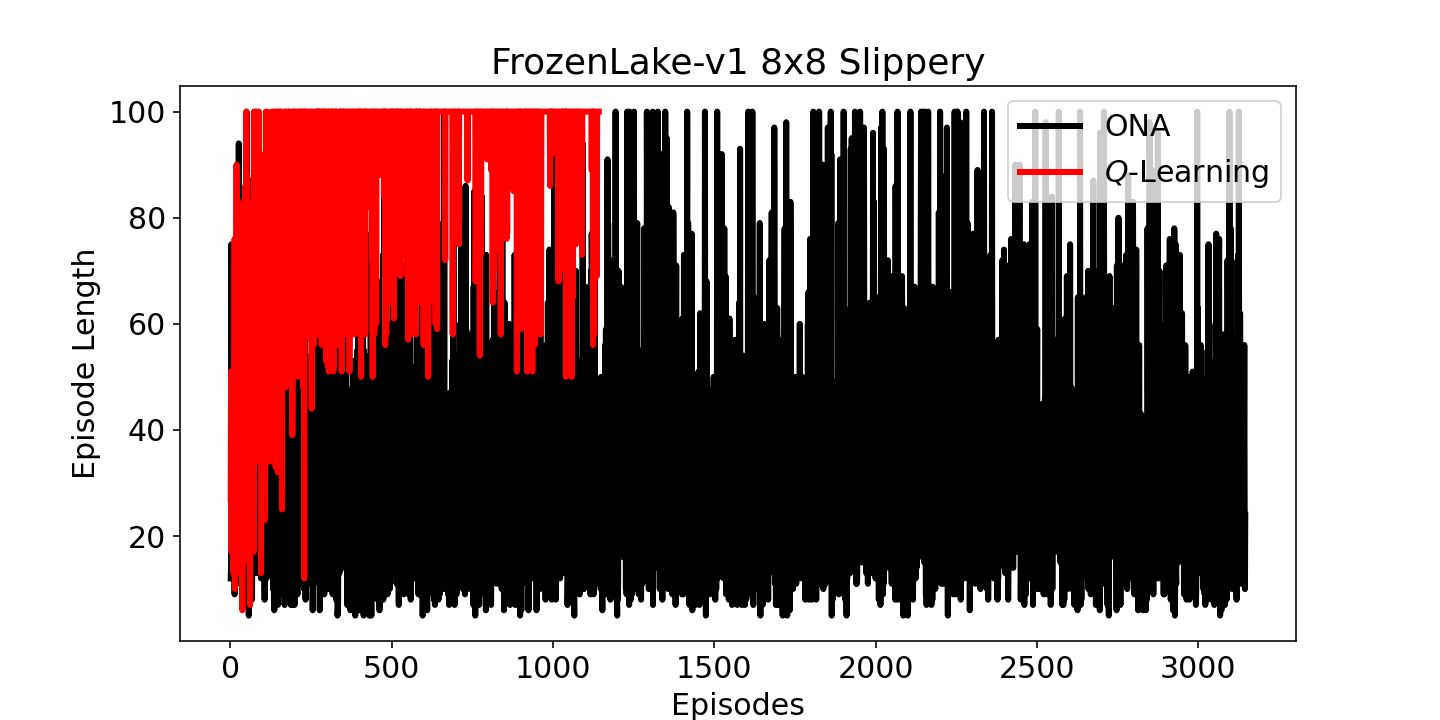}
         \caption{FrozenLake-v1 8x8 Slippery}
         \label{Episode_Length_vs_Episodes_FrozenLake-v1_8x8_Slippery}
     \end{subfigure}
     \begin{subfigure}{0.32\columnwidth}
         \centering
         \includegraphics[width=\columnwidth]{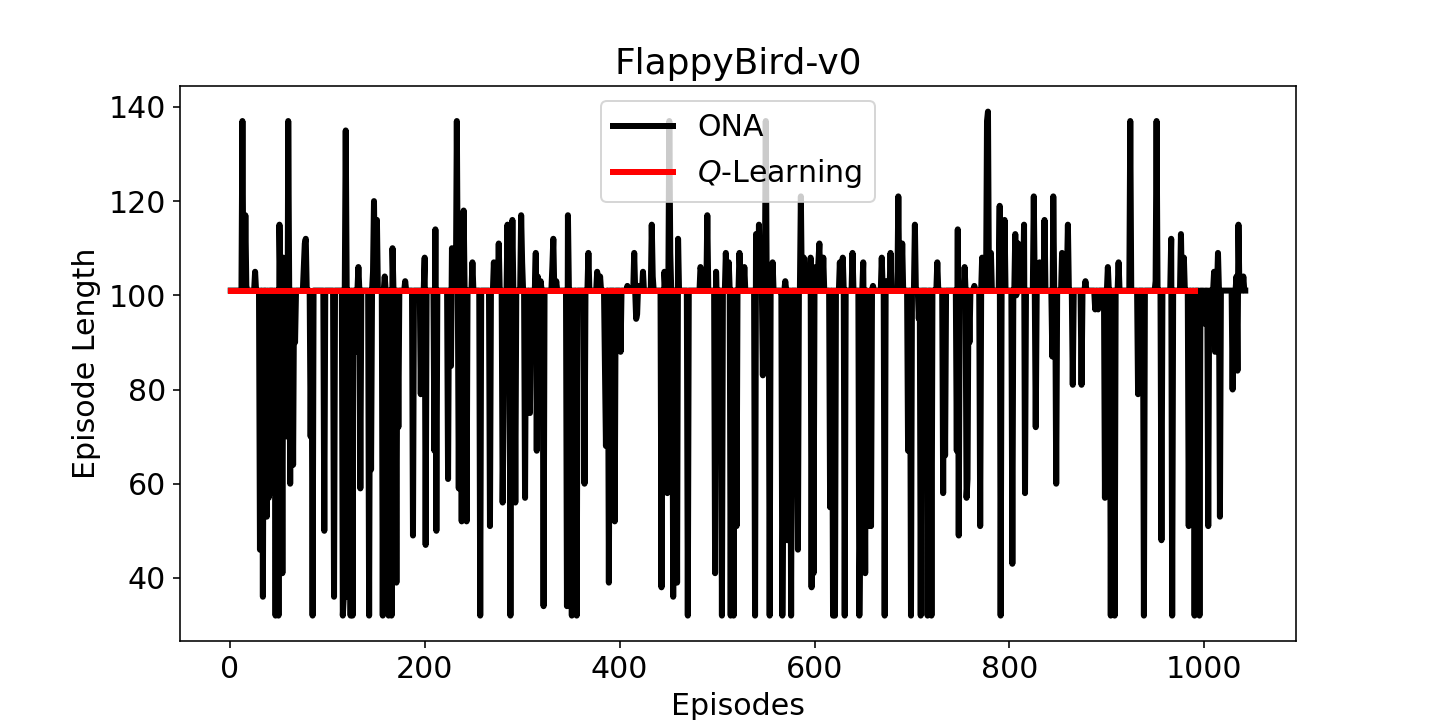}
         \caption{FlappyBird-v0}
         \label{Episode_Length_vs_Episodes_FlappyBird-v0}
     \end{subfigure}
        \caption{Episode Length vs. Episodes.}
        \label{Episode_Length_vs_Episodes}
\end{figure}

\begin{figure}[H]%[t] %[h]
     \centering
     \begin{subfigure}{0.32\columnwidth}
         \centering
         \includegraphics[width=\columnwidth]{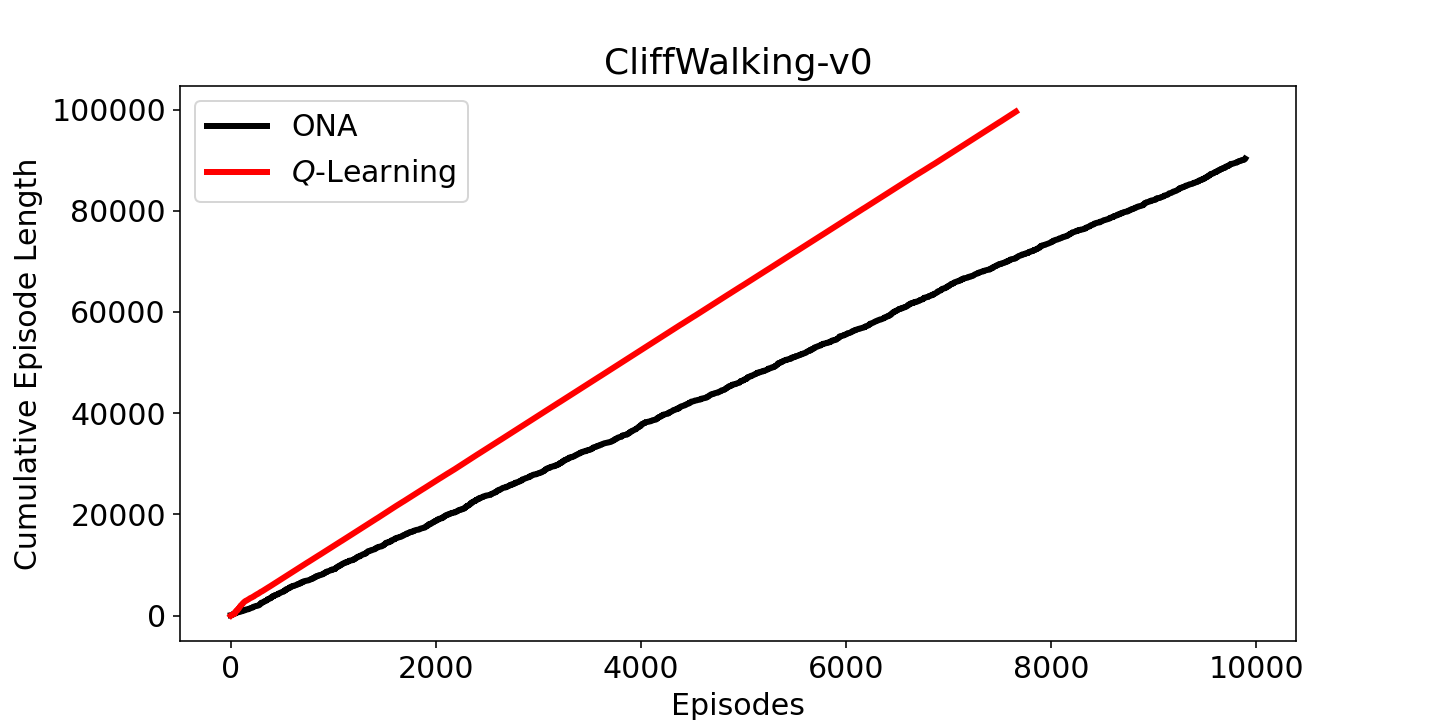}
         \caption{CliffWalking-v0}
         \label{Cumulative_Episode_Length_vs_Episodes_CliffWalking-v0}
     \end{subfigure}
     \begin{subfigure}{0.32\columnwidth}
         \centering
         \includegraphics[width=\columnwidth]{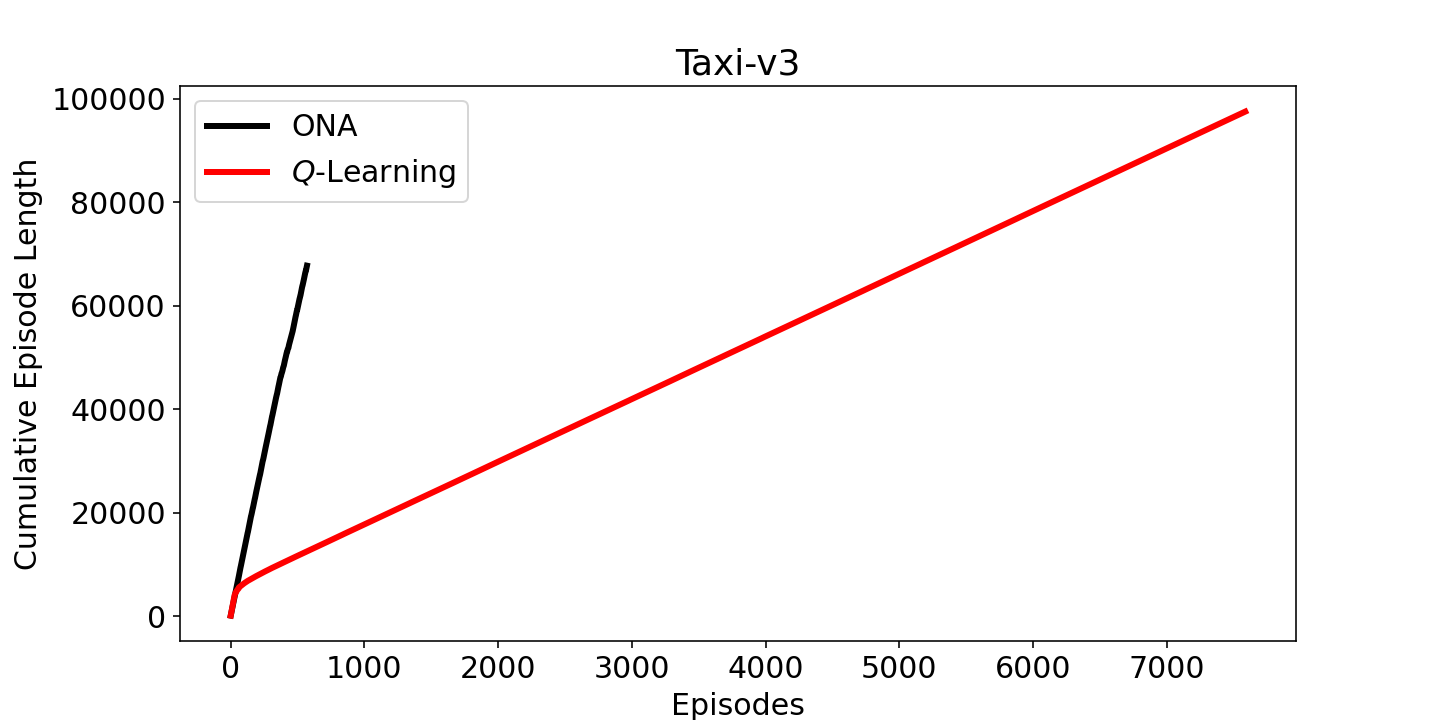}
         \caption{Taxi-v3}
         \label{Cumulative_Episode_Length_vs_Episodes_Taxi-v3}
     \end{subfigure}
     \begin{subfigure}{0.32\columnwidth}
         \centering
         \includegraphics[width=\columnwidth]{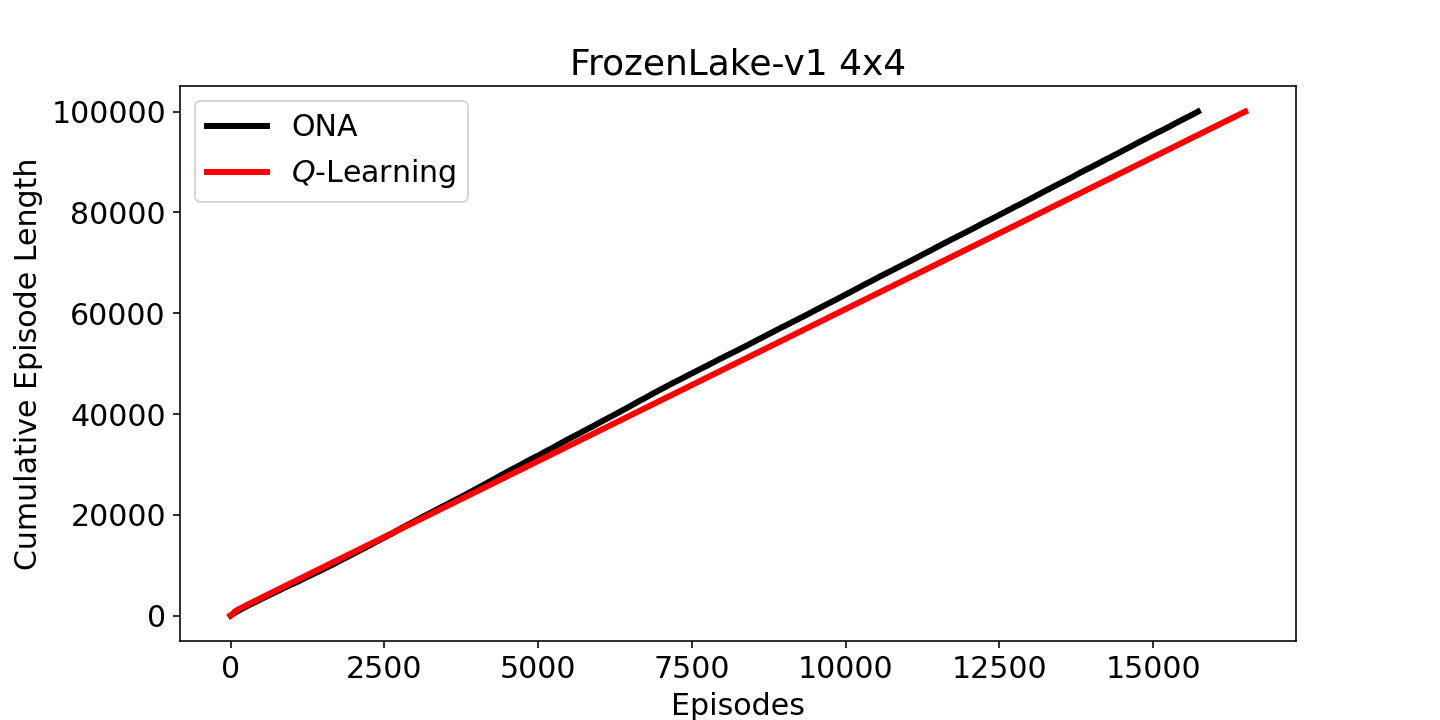}
         \caption{FrozenLake-v1 4x4}
         \label{Cumulative_Episode_Length_vs_Episodes_FrozenLake-v1 4x4}
     \end{subfigure}
     \begin{subfigure}{0.32\columnwidth}
         \centering
         \includegraphics[width=\columnwidth]{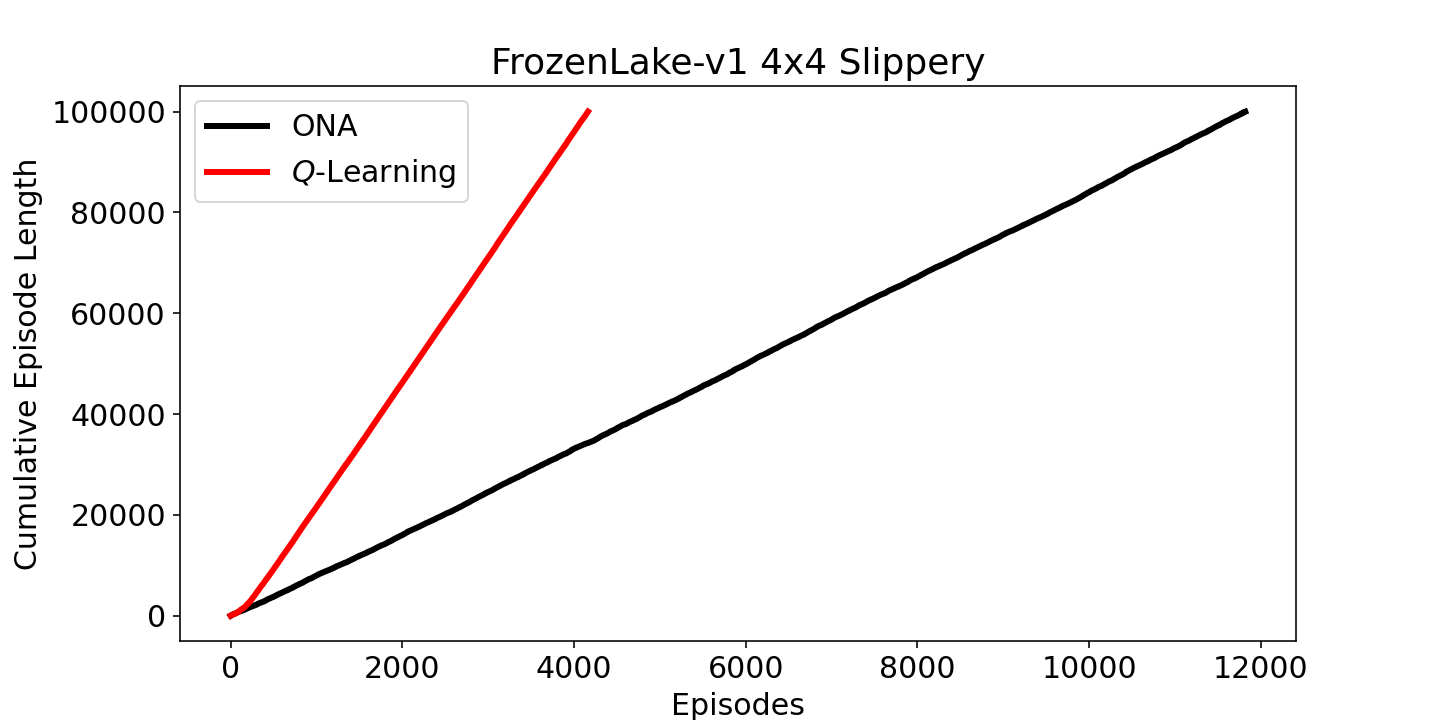}
         \caption{FrozenLake-v1 4x4 Slippery}
         \label{Cumulative_Episode_Length_vs_Episodes_FrozenLake-v1_4x4_Slippery}
     \end{subfigure}
    \begin{subfigure}{0.32\columnwidth}
         \centering
         \includegraphics[width=\columnwidth]{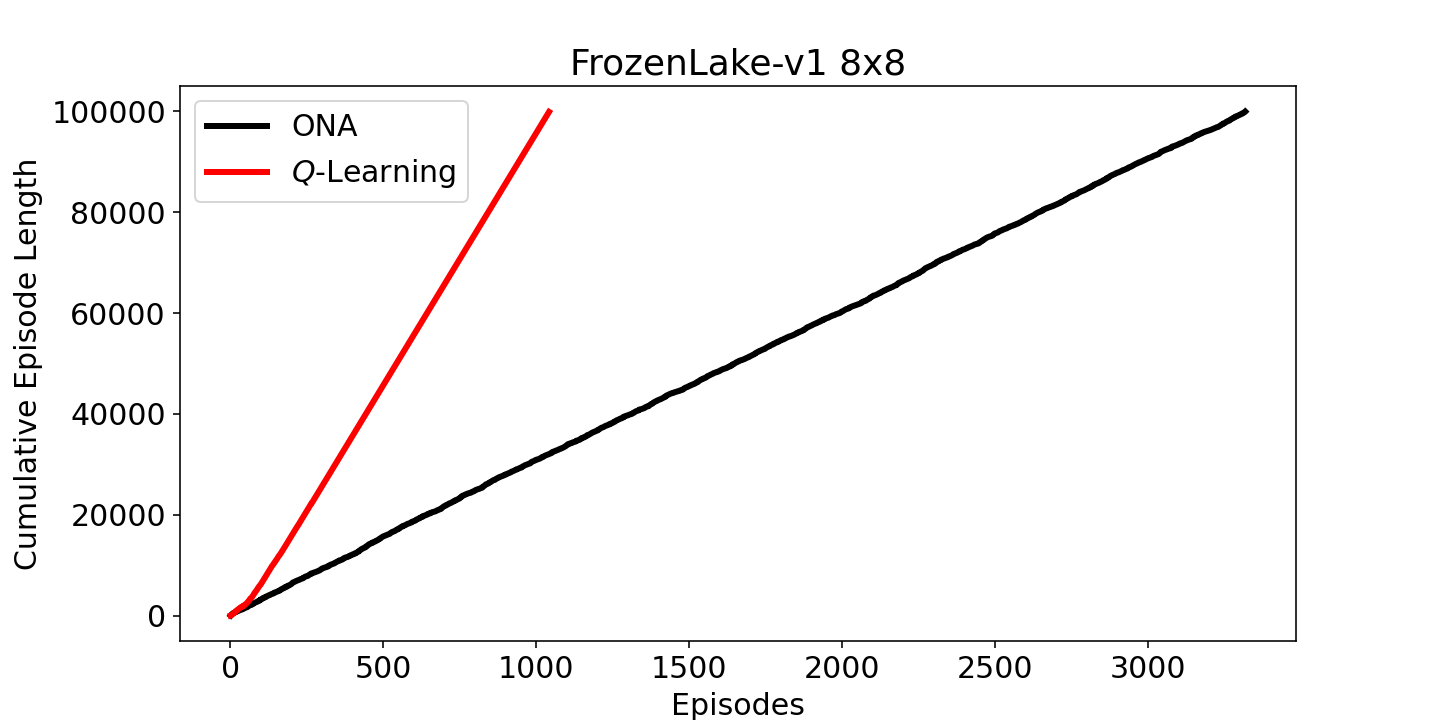}
         \caption{FrozenLake-v1 8x8}
         \label{Cumulative_Episode_Length_vs_Episodes_FrozenLake-v1 8x8}
     \end{subfigure}
    \begin{subfigure}{0.32\columnwidth}
         \centering
         \includegraphics[width=\columnwidth]{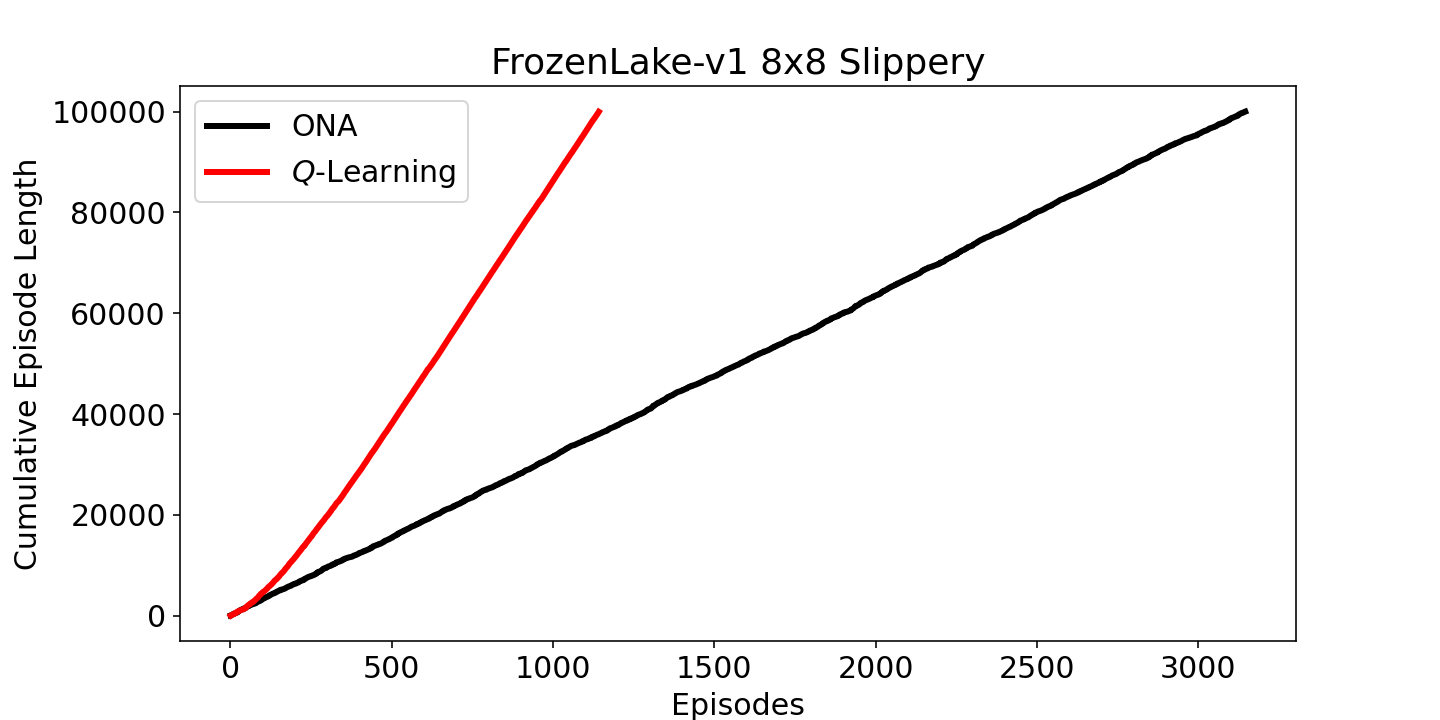}
         \caption{FrozenLake-v1 8x8 Slippery}
         \label{Cumulative_Episode_Length_vs_Episodes_FrozenLake-v1_8x8_Slippery}
     \end{subfigure}
     \begin{subfigure}{0.32\columnwidth}
         \centering
         \includegraphics[width=\columnwidth]{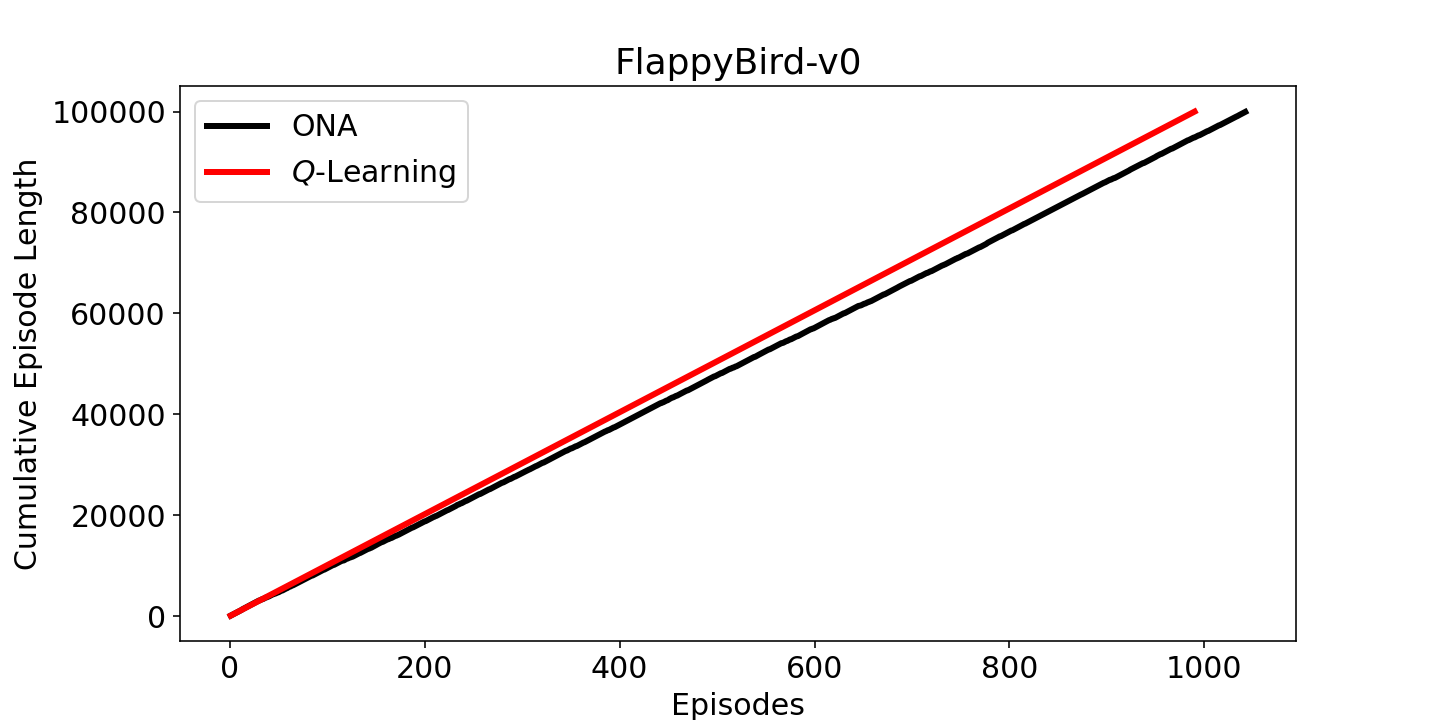}
         \caption{FlappyBird-v0}
         \label{Cumulative_Episode_Length_vs_Episodes_FlappyBird-v0}
     \end{subfigure}
        \caption{Cumulative Episode Length vs. Episodes.}
        \label{Cumulative_Episode_Length_vs_Episodes}
\end{figure}

\bibliographystyle{IEEEtran}
\bibliography{./AGI-Project.bib}
\end{document}